\documentclass[3p]{elsarticle}
\usepackage{blindtext}
\usepackage{graphicx}
\usepackage{amsmath}
\usepackage{bm}
\usepackage{hyperref}
\usepackage{color}
\usepackage[normalem]{ulem}

\usepackage[]{caption}  % font=footnotesize,labelfont=bf
\usepackage[]{subcaption}  % font=footnotesize,labelfont=bf

\usepackage{nomencl}
\makenomenclature
\renewcommand{\nomname}{List of abbreviations used in this paper}

\newcommand{\mx}{m(\bm{x})}
\newcommand{\kx}{k(\bm{x}, \bm{x}')}

\makeatletter
\def\ps@pprintTitle{%
   \let\@oddhead\@empty
   \let\@evenhead\@empty
   \let\@oddfoot\@empty
   \let\@evenfoot\@oddfoot
}
\makeatother

\begin{document}

\nomenclature{ESN}{echo state network}
\nomenclature{GP}{Gaussian process}
\nomenclature{RC}{reservoir computing}
\nomenclature{HO}{hyper-parameter optimization}
\nomenclature{NDN}{nonlinear delay node}
\nomenclature{NMSE}{normalized mean squared error}
\nomenclature{NRMSE}{normalized root mean squared error}
\nomenclature{MG}{Mackey-Glass}
\nomenclature{SGD}{stochastic gradient descent}
\nomenclature{NARMA}{nonlinear auto-regressive moving average}
\nomenclature{SNR}{signal-to-noise ratio}
\nomenclature{i.i.d}{independent identically distributed}
\nomenclature{SER}{symbol error rate}
\nomenclature{OSI}{one-step integration}
\nomenclature{MSI}{multi-step integration}
\nomenclature{MDP}{modular toolkit for data processing}
\nomenclature{PSO}{particle swarm optimization}
\nomenclature{GA}{genetic algorithm}

\title{Bayesian optimization of hyper-parameters in reservoir computing}

\author[uh]{Jan Yperman\corref{cor1}}
\ead{jan\_yperman@uhasselt.be}
\author[uh]{Thijs Becker}
\ead{thijs.becker@uhasselt.be}

\address[uh]{Hasselt University, B-3590 Diepenbeek, Belgium}

\cortext[cor1]{Corresponding author}

\begin{abstract}
We describe a method for searching the optimal hyper-parameters in reservoir computing, which consists of a Gaussian process with Bayesian optimization. It provides an alternative to other frequently used optimization methods such as grid, random, or manual search.
In addition to a set of optimal hyper-parameters, the method also provides a probability distribution of the cost function as a function of the hyper-parameters. We apply this method to two types of reservoirs: nonlinear delay nodes and echo state networks. It shows excellent performance on all considered benchmarks, either matching or significantly surpassing results found in the literature.
% We find that some values for hyper-parameters that have become standard in the research community, are in fact suboptimal for most of the problems we considered. 
In general, the algorithm achieves optimal results in fewer iterations when compared to other optimization methods. We have optimized up to six hyper-parameters simultaneously, which would have been infeasible using, e.g., grid search. Due to its automated nature, this method significantly reduces the need for expert knowledge when optimizing the hyper-parameters in reservoir computing. Existing software libraries for Bayesian optimization, such as Spearmint, make the implementation of the algorithm straightforward. A fork of the Spearmint framework along with a tutorial on how to use it in practice % on how to apply this to a reservoir computing problem
is available at \url{https://bitbucket.org/uhasseltmachinelearning/spearmint/}
\end{abstract}


\maketitle

\printnomenclature

\section{Introduction}

Error backpropagation combined with stochastic gradient descent (SGD) is a simple and highly successful training method for feedforward neural networks \cite{lecun2012efficient,lecun2015deep}. In contrast, training recurrent neural networks with this method poses considerable difficulties \cite{hochreiter2001gradient}. Different approaches such as, e.g., more complicated training schemes \cite{arjovsky2016unitary} or different architectures \cite{hochreiter1997long, schuster1997bidirectional} have been explored to tackle this problem, and have been quite successful. A popular method skips the training stage of the recurrent network completely. Only the connections of the network to the output are trained. As a result, the training stage is computationally fast and straightforward to implement. This training paradigm goes by the name of reservoir computing (RC) \cite{lukovsevivcius2012practical}. Despite its simplicity compared to other techniques, it achieves state-of-the-art results on several machine learning tasks \cite{lukovsevivcius2012practical}. However, on large datasets and more challenging benchmarks, RC techniques fall short of e.g.~LSTMs \cite{hochreiter1997long}, which have become feasible in recent years. Therefore, RC techniques are now mainly being developed in the form of hardware implementations, using e.g.~photonics \cite{fischer2016photonic, duport2016fully, antonik2016towards, katumba2017multiple} or electronics \cite{soriano2015delay}, which could be computationally faster compared to digital ones.

Although the training stage is dramatically simplified, one still needs to set several hyper-parameters that determine the general properties of the network, such as its size and spectral radius. Hyper-parameter optimization (HO) requires an experienced researcher, i.e., acquiring optimal results still necessitates expert input \cite{jaeger2007optimization}. It is, therefore, of interest to automate the search for hyper-parameters in reservoir computing. A straightforward HO method is grid or random search, which is suitable for finding the optimum when considering only a few hyper-parameters. If there are many hyper-parameters such an approach is not viable, because the volume of the hyper-parameter space grows exponentially with the number of hyper-parameters. A step by step plan for manual HO for RC is provided in \cite{lukovsevivcius2012practical}. As noted in \cite{lukovsevivcius2012practical}, automated HO is a common topic in machine learning, and these methods are applicable to RC. A few methods have been proposed for automated HO, including Particle Swarm Optimization (PSO) \cite{rabin2013sensitivity, sergio2012pso, basterrech2014experimental}, various forms of Genetic Algorithms (GAs) \cite{ferreira2009genetic, ferreira2013approach, rigamonti2016echo}, and stochastic gradient descent applied to the hyper-parameters \cite{jaeger2007optimization}.
% To our knowledge, the only attempt at an automated HO scheme for RC is the application of stochastic gradient descent to the hyper-parameters themselves \cite{jaeger2007optimization}.

In this paper, we show that a Gaussian process with Bayesian optimization is able to achieve an optimal choice for RC hyper-parameters in an automated way. The method is applied to two types of reservoirs: dynamical nonlinear delay nodes (NDNs) and echo state networks (ESNs). For both the NDN and ESN, our results either match the considered benchmarks, or surpass them by one or several orders of magnitude. The implementation of Gaussian processes with Bayesian optimization is non-trivial. We therefore use the Spearmint library \cite{NIPS2012_4522}, which was developed in the context of HO in machine learning. HO with Bayesian optimization matches or surpasses human performance on several machine learning tasks and for different algorithms, such as deep learning and support vector machines \cite{NIPS2012_4522}. However, it seems to fail completely on other tasks \cite{Goodfellow-et-al-2016-Book}. Because we achieve good results for all considered types of reservoirs and benchmarks, our work indicates that its performance for HO in RC is robust.

The paper is organized as follows. In Section \ref{sec:RC}, we describe the ESN and NDN formalisms. The theory of Bayesian optimization of Gaussian processes is briefly sketched in Section \ref{sec:GP}. In Section \ref{sec:tasks}, we describe the benchmark tasks. Details of the implementation are discussed in Section \ref{sec:implement}.
In Section \ref{sec:NDNres}, we present and discuss the results. In Section \ref{sec:spdem}, we discuss the behaviour of the Bayesian search process. We end with a conclusion and outlook to possible future work in Section \ref{sec:conclusion}.

\section{Reservoir Computing}\label{sec:RC}

\subsection{Echo state networks}

Echo state networks were introduced independently by Herbert Jaeger \cite{jaeger2001echo} and Wolfgang Maass (under the name \textit{Liquid State Machines} \cite{maass2002real}). The idea is to use a recurrent neural network which is fixed, i.e., its connection weights and any biases are not trained. One can drive this network using an input, which will change the state of the network (i.e., the value at each of the nodes). The output is then constructed using a linear combination of all, or a subset of, the node values. The weights of this linear combination are most commonly obtained using linear regression.
The values of the nodes are updated according to the rule:
\begin{equation}
    \bm{x}\left(n + 1\right) = \sigma\left(\bm{W x}(n) + \bm{W}^{\textit{in}}\bm{u}(n + 1) + \bm{b}\right),
    \label{eq:esnstate}
\end{equation}
where $\bm{x}(n)$ is the $N$-dimensional state vector of the network at time $n$, $\bm{u}(n)$ is the $K$-dimensional input vector at time $n$, $\bm{W}$ is the $N \times N$ reservoir weight matrix, $\bm{W}^{\textit{in}}$ is the $N \times K$ input weight matrix, $\bm{b}$ is a constant bias term and $\sigma(.)$ is a sigmoid function (in our case we use the $\tanh$ function). 

The reservoir weight matrix is rescaled as $\bm{W'} = \rho \bm{W} / \left| \lambda_{\textit{max}} \right|$, where $ \left| \lambda_{\textit{max}} \right|$ is the spectral radius of the network (i.e., the largest eigenvalue of $\bm{W}$) and $\rho$ is a scaling parameter (effectively the spectral radius of the rescaled network). This rescaling operation is a popular choice in the reservoir computing literature. It, however, does not guarantee the echo state property \cite{scardapane2017randomness}. See, e.g., \cite{manjunath2013echo,wainrib2016local} for rigorous discussions on sufficient conditions for the echo state property.

The output is given by:
\begin{equation}
\bm{y}(n) = \bm{U}^{\textit{out}} \bm{x}(n),
\end{equation}
where the output weights matrix $\bm{U}^{\textit{out}}$ is obtained using, e.g., linear regression. 
We performed linear regression using ridge regression (a.k.a.~Tikhonov regularization), which is a modified version of the linear regression equations \cite{lukovsevivcius2012practical}:
\begin{equation}
\bm{U}^{\textit{out}} = \left(\bm{X}^T \bm{X} + \lambda^2 \bm{I}\right)^{-1} \bm{X}^T \bm{y},
\label{eq:ridge}
\end{equation}
where $\bm{I}$ is the identity matrix, $\lambda$ is the regularization parameter, and $\bm{X}$ is the matrix of all reservoir states.
% \textcolor{red}{Ok om gewoon $\bm{x}$ te gebruiken? Kan eventueel ook subset van reservoir toestanden zijn. \cite{rodan2011minimum} doet dit ook.}
Performance is measured with the normalized mean squared error (NMSE):
\begin{equation}
    \text{NMSE} = \frac{\langle\lVert\hat{y}(t) - y(t)\rVert^2\rangle}{\langle\lVert y(t)-\langle y(t) \rangle\rVert^2\rangle},
\end{equation}
where $\hat{y}(t)$ is the predicted value and $y(t)$ is the target value. The denominator is simply the variance of the test set under consideration. The root mean squared version (NRMSE) is:
\begin{equation}
    \text{NRMSE} = \sqrt{ \frac{\langle\lVert\hat{y}(t) - y(t)\rVert^2\rangle}{\langle\lVert y(t)-\langle y(t) \rangle\rVert^2\rangle} }.
\end{equation}

\subsection{Nonlinear delay nodes}

The concept of a nonlinear delay node as a reservoir was introduced in \cite{appeltant2011information}. It can be implemented in hardware with optical and electronic components \cite{larger2012photonic,paquot2012optoelectronic,brunner2013parallel,soriano2015delay,siliconRC,fischer2016photonic, duport2016fully, antonik2016towards, katumba2017multiple,soriano2015delay}. We use two different types of NDNs.

\subsubsection{Mackey-Glass delay differential equation}\label{sec:MG}

Reservoir states can be computed by solving a delay differential equation.
In this work, we consider the Mackey-Glass (MG) differential equation with delay time $\tau$:
\begin{equation}\label{eq:MGdynamic}
\dot{X}(t) = - X(t) + \eta \frac{X(t-\tau)+\gamma J(t)}{1 + [X(t-\tau)+\gamma J(t)]^p},
\end{equation}
where $\gamma$, $\eta$, and $p$ are adjustable parameters, $X(t)$ is the state of the system, and $J(t)$ is the input.
We note that other delay differential equations can be used equally well.

\begin{figure}
\centering
\includegraphics[width=0.75\columnwidth]{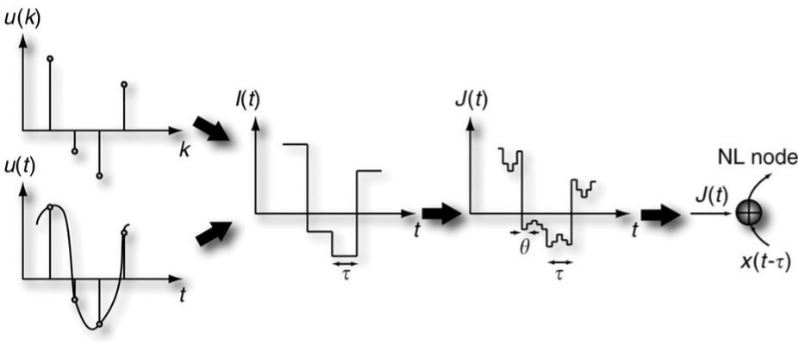}
\caption{Illustration of the preprocessing of the input signal. The discrete (or discretized) input signal $u(k)$ is transformed to $J(t)$ by multiplying it with a random mask. Note that in the figure the mask has more values than $-1$ or $1$. $J(t)$ is the input to the nonlinear delay node. (The step that transforms the discrete signal to a stepwise continuous signal, from $u(k)$ to $I(t)$, is not explicitly mentioned in the main text.) 
Reprinted by permission from Macmillan Publishers Ltd: Nat.~Comm.~\cite{appeltant2011information}, copyright (2011).}
\label{fig:ndn_schematic}
\end{figure}

The reservoir states are calculated as follows. Consider a discrete (or discretized) one-dimensional signal $u(k)$, with $k$ integer. We want to map each scalar value $u(k)$ to a reservoir state of dimension $N$.
This reservoir state is obtained by solving the MG differential equation \eqref{eq:MGdynamic}.
Solving over a time interval $\tau$ enables us to compute one reservoir state. Suppose we start at time $t=0$ and end at $t=\tau$. The input $J(t)$ is calculated by multiplying $u(k)$ with a mask $M$. For a reservoir of dimension $N$, the mask is a set of $N$ values, which we take to be either -1 or 1: $M$ is a random element from the set $\{-1, 1\}^N$. By multiplying the mask with the signal, $J(t) = M \times u(k)$, one finds $N$ values, equal to $\pm u(k)$, which are constant over a time $\tau / N$
\begin{align*}
[J(0), J(\tau / N)] &= M(0) u(k) \\
]J(\tau / N), J(2 \tau / N)] &= M(1) u(k) \\
\ldots \\
]J((N-1) \tau / N), J(\tau)] &= M(N) u(k).
\end{align*}
This preprocessing sequence of the signal is illustrated in Figure \ref{fig:ndn_schematic}.

\begin{figure}
\centering
\includegraphics[width=0.75\columnwidth]{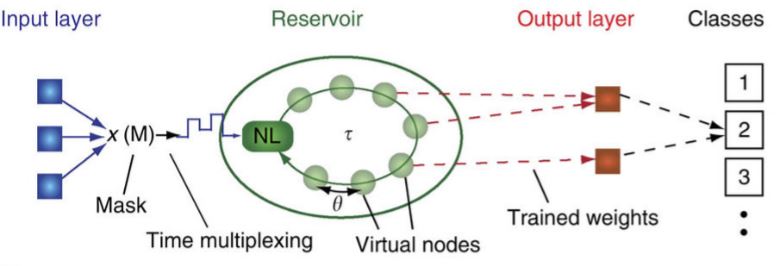}
\caption{Schematic representation of a nonlinear delay node.
The input signal is first multiplied by a mask (see Figure \ref{fig:ndn_schematic}). The nonlinear delay node solves the MG equation. Values of the nodes are the solution to the MG equation, taken at time intervals $\theta$. The node values are connected to the output layer by trained weights. 
In this figure the output performs a classification, but the output could also, e.g., map to a scalar number. Note that this figure illustrates the more general case of higher dimensional input and output. In our case, input and output are scalar values. Reprinted by permission from Macmillan Publishers Ltd: Nat.~Comm.~\cite{appeltant2011information}, copyright (2011).}
\label{fig:ndn_drawing}
\end{figure}

Given the input $J(t)$, the MG equation can be solved. The reservoir state $x_i$, $i \in \{1, \ldots, N \}$, is equal to the value of $X(t)$ at time $t = i \tau / N$. The following reservoir state is then calculated by solving equation \eqref{eq:MGdynamic} from $t=\tau$ to $t=2\tau$ with input $M \times u(k+1)$, and so on. The times at which the solution to the MG equation is read can be interpreted as nodes from a network. They are referred to as ``virtual nodes''. Training of the connections between the reservoir state and the output is done the same as with ESNs.
The whole NDN system is illustrated in Figure \ref{fig:ndn_drawing}.

It is clear that the reservoir states depend on the previous state, via $X(t-\tau)$, and the input, via $J(t)$. The delay-dynamical equation therefore functions as a recurrent network. It is important that the values of the input mask $M$ alternate. Otherwise, the differential equation relaxes to its stationary value, and the dynamics cannot perform a projection to a rich set of reservoir states, cf.~the discussion in \cite{appeltant2011information}.

The parameters in the MG dynamics equation \eqref{eq:MGdynamic} have specific functions. The input scaling $\gamma$ determines how strongly the dynamics is influenced by new input values $J(t)$. $\eta$ determines the influence of both $J(t)$ and $X(t-\tau)$. The parameter $p$ determines the nonlinearity of the dynamics. If other nonlinear delay differential equations are used, the parameters in those equations will have similar roles.

\subsubsection{Sine}\label{sec:Sine}

First proposed in \cite{paquot2012optoelectronic}, this version of the nonlinear delay node uses an architecture similar to that of \cite{appeltant2011information} (reviewed in Section \ref{sec:MG}), but with a simpler dynamics. The update rule for the reservoir states is given by:
\begin{align}
    x_i(n) = \begin{cases}
    \sin\left(\alpha x_{i-k}\left(n-1\right) + \beta m_i u(n) + \phi\right) & k \leq i < N \\
    \sin\left(\alpha x_{N + i - k}\left(n-2\right) + \beta m_i u(n) + \phi\right) & 0 \leq i < k,
    \end{cases}
\end{align}
where $\alpha$, $\beta$ and $\phi$ are adjustable parameters, $x_i(n)$ represents the state of the $i$th virtual node at discrete timestep $n$, $N$ is the total number of virtual nodes, $m_i$ represents the $i$th value of the mask (chosen at random from $\{-1,1\}$ as before), $u(n)$ represents the input at discrete timestep $n$ and $k \in \{1,\dots,N-1\}$. In our implementation we set $k = \left \lfloor N  / 3 \right \rfloor$, where $\lfloor . \rfloor$ denotes the floor function. Further details can be found in \cite{paquot2012optoelectronic}.

%%%%%%%%%%%%%%%%%%%%%
%% GAUSSIAN PROCESSES
%%%%%%%%%%%%%%%%%%%%%

\section{Gaussian Processes}\label{sec:GP}

When fitting a function, several approaches are possible. One option is to consider a class of functions, for example all linear functions. Another approach is to specify a probability distribution over all possible functions, where more likely functions have higher probabilities. %, cf.~Figure \ref{fig:gp_example}.
The latter approach can be achieved with Gaussian processes (GPs). A GP is equivalent to Bayesian linear regression with an infinite number of basis functions \cite{williams2006gaussian}. It is therefore possible to fit a large class of functions exactly. A function $f(\bm{x})$ fitted by a GP probability distribution is denoted by
\begin{equation}
f(\bm{x}) \sim \mathcal{GP} (m(\bm{x}), k(\bm{x}, \bm{x}')).
\end{equation}
The vector $\bm{x}$ are the hyper-parameters of the reservoir and $f(\bm{x})$ is the error function, e.g., $f(\bm{x}) = \text{NMSE}(\{\gamma, \eta, p\})$ for the Mackey-Glass NDN. Note that the symbol $\bm{x}$ is also used for the reservoir states. Which one is meant should be clear from the context.
Gaussian processes are able to describe probability distributions over complicated functions, which is what we need for the optimization algorithm.
For an introduction to Gaussian processes, we refer to \cite{williams2006gaussian}. Several applications are discussed in \cite{brochu2010tutorial}.

A GP is completely specified by its mean function $\mx$ and covariance function $\kx$. 
$\mx$ gives the average function value at $\bm{x}$. Values at different positions $\bm{x}$ and $\bm{x}'$ are correlated by an amount $\kx$, also called the kernel. In practice, this kernel is used to ensure that points close to each other in hyper-parameter space are likely to have similar values, which follows from an assumption about the behaviour of $f(\bm{x})$.
Intuitively, we expect small changes in the objective function if the hyper-parameters are changed slightly. 

\begin{figure}
	\centering
	\includegraphics[width=0.6\columnwidth]{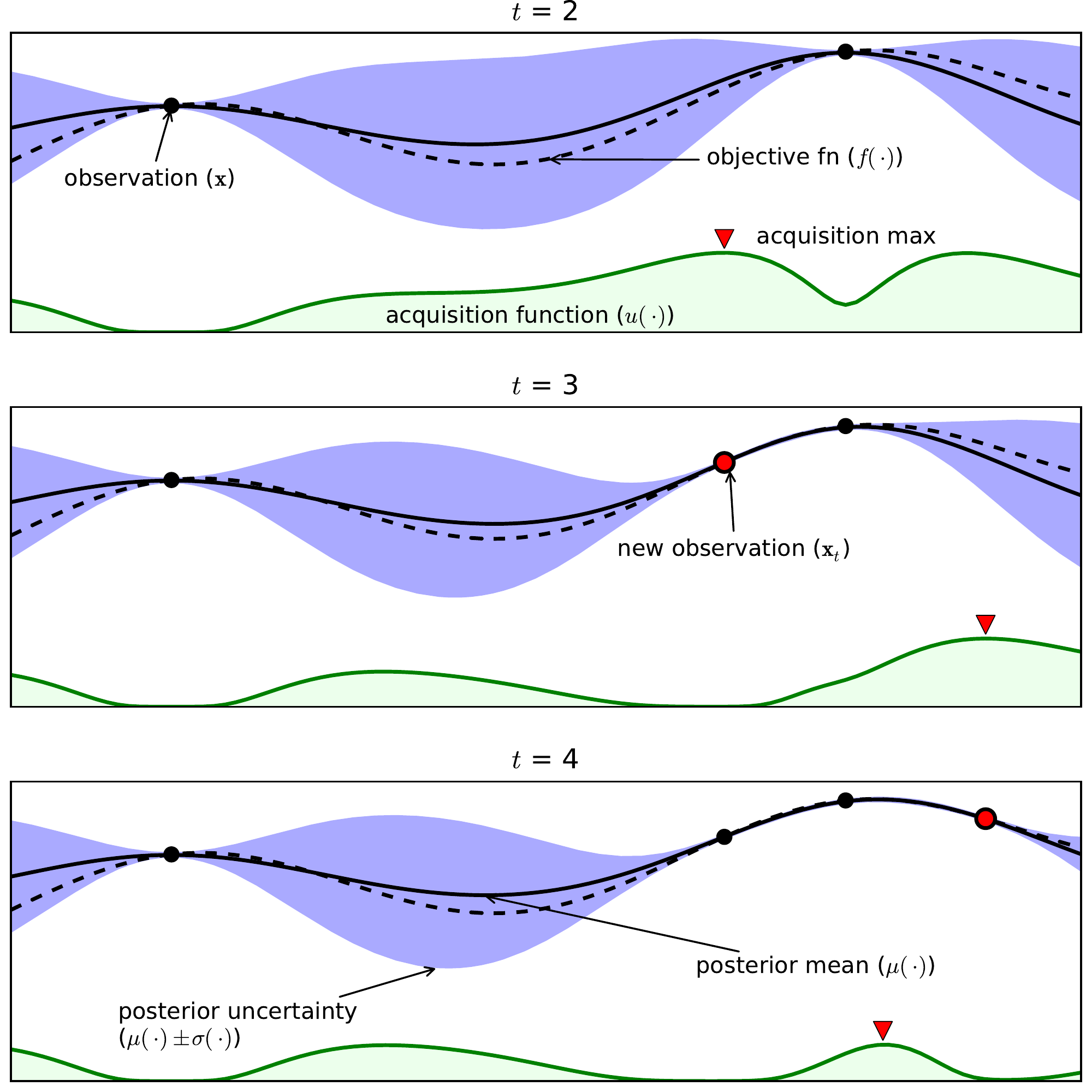}
	\caption{Illustration of a Gaussian process with Bayesian updating, and the expected improvement acquisition function. The ``time'' $t$ denotes the number of measurements of the error function. Measurements are denoted by points.
		The dashed line is the real function that needs to be fitted. The black solid line is the average of the GP, and the purple band is the variance on the predicted average. Note that the variance is zero at these points, which is the special case of noiseless measurements. In the case of noisy measurements there would be a finite variance at the measurement points. On the bottom of each figure is the acquisition function, which determines where to perform a new measurement (denoted by a red triangle). Note that the acquisition function does not choose to sample at the point with the highest average, but also accounts for the reduction in variance. Reprinted with permission from \cite{brochu2010tutorial}.}
	\label{fig:gp_tutorial}
\end{figure}

A GP that models $f(\bm{x})$ well is achieved by performing measurements of $f(\bm{x})$, and consequently updating $\mx$ and $\kx$ to take the obtained information into account. This updating is done with Bayesian statistics \cite{sivia2006data}. Before any measurements are performed, one needs to set a prior, i.e., guess a reasonable form for $m(\bm{x})$ and $k(\bm{x}, \bm{x}')$. 
An example would be $\mx = 0$ for all $\bm{x}$ and the covariance function
\begin{equation}\label{eq:kexample}
k(\bm{x}_i, \bm{x}_j) = \exp \left( - \frac{1}{2 \xi^2} |\bm{x}_i - \bm{x}_j|^2 \right),
\end{equation}
with $\xi$ an adjustable parameter. This covariance function is a popular choice, although it is too smooth for most machine learning problems. In this paper, the automatic relevance detection Mat\'{e}rn 5/2 kernel is used \cite{NIPS2012_4522}. 
After each measurement one can update $\mx$ and $\kx$ to obtain the posterior GP. An illustration of a GP and its evolution when measurements are performed is shown in Figure \ref{fig:gp_tutorial}. We note that it is possible to incorporate the effect of noise in the measurement process. 

One can analytically calculate the mean and variance of the function distribution at each $\bm{x}$. The choice of where to perform a new measurement is not only determined by the improvement in the average of $f(\bm{x})$ (\textit{exploitation}). It is also influenced by how much new information we gain by performing the measurement, which is determined by the reduction in variance (\textit{exploration}). The balance between exploitation and exploration is determined by the acquisition function, which needs to be chosen by the researcher. The acquisition function used here is the expected improvement. An example of this acquisition function is shown in Figure \ref{fig:gp_tutorial}. A discussion of several acquisition functions can be found in \cite{brochu2010tutorial}.

The hyper-parameters of the GP itself, such as $\xi$ in Eq.~\eqref{eq:kexample}, can also be optimized. A more practical method is to average these parameters out, e.g.~with Bayesian statistics. This leads to a completely parameter free algorithm. The implementation of the GP with Bayesian updating is done with Spearmint \cite{NIPS2012_4522}. This library contains several non-trivial extensions, such as averaging out the hyper-parameters of the GP with Bayesian statics, and input warping \cite{snoek2014input}. Input warping deals with the problem of $f(\bm{x})$ being non-stationary, i.e., the covariance function $k(\bm{x}_i, \bm{x}_j)$ is not invariant under translations in the input space. Error functions in machine learning are generally non-stationary: different regions in the hyper-parameter space have different length scales of function variation. Therefore, a choice of covariance function such as in Eq.~\eqref{eq:kexample}, which is invariant under translations, cannot model most error functions well. Input warping can be seen as applying a particular kind of non-stationary kernel to the original data, which solves the problem of the non-stationarity of the error function.

Our final choice for the optimal hyper-parameters is the best set found during the search. An interesting question is which functions can be fitted with a GP. A discussion on the convergence of the GP to the average of the underlying function $f(\bm{x})$ can be found in \cite{williams2006gaussian}. There are two necessary conditions for convergence in the limit of an infinite number of measurements: (1) the covariance function $\kx$ should be non-degenerate, which is the case for us; (2) the average of the underlying function should be sufficiently well-behaved, so that it can be represented by a generalized Fourier series.
Of more practical importance is whether the search for the minimum converges. This depends on the acquisition function, the function to be estimated, and the prior on the GP. Sufficient conditions for the convergence of GPs with Bayesian optimization were given by Mo\v{c}kus \cite{mockus1994application}. His assumptions are quite general, so are expected to apply to a large number of error functions, although it is difficult to numerically check whether they are fulfilled. Depending on the implementation, the expected improvement acquisition function can converge near-optimally, although it is possible that the algorithm does not converge at all \cite{bull2011convergence}. We note that, for practical purposes, convergence to the absolute minimum is not necessary. Rather, we are interested in a ``good enough'' minimum. Besides asymptotic behaviour, one is also interested in the rate of convergence. Recent work has shown that for the upper confidence bound acquisition function, there are algorithms for which convergence of a GP to the optimum is exponentially fast \cite{kawaguchi2015bayesian,de2012exponential}. As discussed in Section \ref{sec:NDNres}, the convergence behaviour we observe is also approximately exponential.

We now briefly discuss the differences with other optimization methods.
The fit for unseen parameter regions is non-local: it takes into account the information of all measurements. This should be contrasted with SGD, where only local information of the error function determines the parameter update. There is also no exploration component in SGD. If the connection weights of the reservoir output are high, the SGD learning rate becomes exponentially slow \cite{jaeger2007optimization}. This problem does not occur for GPs.  
When optimizing hyper-parameters step by step \cite{lukovsevivcius2012practical}, the hyper-parameters are optimized with all others fixed. The interdependency of the hyper-parameters is therefore not fully exploited. This inevitably leads to a sub-optimal end result. In contrast, such interdependency can be represented and taken advantage of with a GP.
In contrast to random search, the new measurements are directed towards interesting regions, which means we need far fewer measurements to get good results. This compensates for the slight increase in computation time needed to calculate and update the GP.

%%%%%%%%%%%%%%%
%% TASKS
%%%%%%%%%%%%%%%

\section{Benchmark Tasks}\label{sec:tasks}
In this section we describe the data sets we use to test the performance of the algorithm.
\subsection{Santa Fe}

The Santa Fe data set consists of the output of a chaotic laser system \cite{makridakis1994time}. We use the \textit{A.cont} data set, which can be obtained from \cite{sfonline}. The goal is to predict the next step of the time series.

\subsection{NARMA 10}

The Nonlinear Auto-Regressive Moving Average (NARMA) of order 10 was originally introduced for use as a timeseries prediction benchmark in \cite{atiya2000new}. It is generated by
\begin{equation}
y(t+1) = 0.3y(t) + 0.05y(t) \sum_{i=0}^9 y(t-i) + 1.5 s(t-9)s(t) + 0.1,
\end{equation}
with $s(t)$ a random term with uniform distribution in $[0, 0.5]$. The goal is to do a one-step ahead prediction of $y(t)$.

\subsection{Nonlinear channel equalization}\label{sec:Nonlinear}
This task was first introduced in \cite{mathews1994adaptive} and has since been used as a benchmark several times \cite{jaeger2004harnessing,paquot2012optoelectronic,rodan2010simple}. The data set is created as follows:
An i.i.d.~sequence $d(t)$ is generated by randomly choosing values from $\{-3,-1,1,3\}$. This signal is then passed through a linear channel described by:
\begin{align*}
    q(n) = &0.08 d\left(n + 2\right) - 0.12 d\left(n + 1\right) + d\left(n\right) 
    + 0.18 d\left(n - 1\right) - 0.1 d\left(n - 2\right) + 0.091 d\left(n - 3\right) \\
    &- 0.05 d\left(n - 4\right) + 0.04 d\left(n - 5\right) + 0.03 d\left(n - 6\right) + 0.01 d\left(n - 7\right),
\end{align*}
which in turn is passed through a nonlinear channel:
\begin{equation*}
u(n) = q(n) + 0.036 q(n)^2 - 0.011 q(n)^3 + \nu (n),
\end{equation*}
where $\nu(n)$ is an i.d.d.~Gaussian noise with zero mean and a standard deviation adjusted to yield signal-to-noise ratios (SNR) ranging from $12$ to $32$ dB. 
The objective is to reconstruct $d(n - 2)$ given $u(n)$, for several values of the SNR (so $u(n)$ is the input and the target is $y(n) = d(n-2)$). Following \cite{jaeger2004harnessing, rodan2011minimum}, we shifted $u(n)$ by +30. The quality measure of the algorithm is given by the Symbol Error Rate (SER), which is the fraction of incorrect symbols obtained.

%%%%%%%%%%%%%%%%%%%
%% IMPLEMENTATION
%%%%%%%%%%%%%%%%%%%

\section{Implementation}\label{sec:implement}

Before discussing the benchmarks, we give some details on how the algorithms are implemented.

\subsection{Mackey-Glass dynamics implementation}\label{sec:mgimplement}

If $p$ is uneven, the MG dynamics Eq.~\eqref{eq:MGdynamic} is unstable for negative values. We therefore rescale $u(k)$ so that it is always positive. We furthermore add an offset parameter $\delta \geq 1$ to the mask $M \in \{ -1 + \delta, 1 + \delta \}^N$. 

For comparison purposes, the number of nodes $N$ is fixed. The (time) distance between the nodes is denoted by $\theta$. A larger $\theta$ gives a longer relaxation time between the nodes. By definition one has that $\tau = N \theta$.

We employ Euler integration to solve the MG equation. The number of integration steps between adjacent nodes determines the precision of the obtained solution. However, the method used to solve the differential equation does not necessarily influence the quality of the reservoir. Put differently, one does not need an exact solution of the differential equation to have a high-quality reservoir. The essential property of the reservoir is a ``good'' nonlinear projection to a high-dimensional space, which is not necessarily the exact solution of the differential equation.
We perform Euler integration with either several integration steps between adjacent nodes (number of steps $2 |\lfloor \theta / 0.1 - 1 \rfloor| + 2$), or with one integration step between adjacent nodes. We refer to these methods as, respectively, multi-step integration (MSI) and one-step integration (OSI). Both choices give good results, with the latter method obviously being faster.

For all NDN implementations, ridge regression is performed with sci-kit learn \cite{scikit}.

\subsection{Echo state network}

To implement the echo state networks we made use of the software library Oger \cite{verstraeten2012oger}, which in turn builds on the Modular toolkit for Data Processing (MDP) \cite{zito2009modular}.

\subsection{Bayesian optimization}

For the implementation of the GP we used the Spearmint library, available at \url{https://github.com/HIPS/Spearmint}. A fork of this framework, as well as a step-by-step tutorial detailing both installation and a practical example (NDN MG on NARMA 10), is available at \url{https://bitbucket.org/uhasseltmachinelearning/spearmint/}. This code replaces the MongoDB backend with an SQLite one. This removes the need for a background server process, which makes it easier to deploy in a cluster environment or as part of a Jupyter Notebook.

%%%%%%%%%%%%%%%
%% RESULTS
%%%%%%%%%%%%%%%

\section{Results}\label{sec:NDNres}

\subsection{Santa Fe laser data}

For the MG NDN, we compare our results with those of \cite{appeltant2012reservoir}. Details of the implementation can be found in \ref{app:sf}. As in \cite{appeltant2012reservoir}, we take $N=200$.
Before the other parameters are optimized, ten different masks are generated, and the best performing one is selected. 
There are six parameters to optimize: $p$, $\gamma$, $\eta$, $\theta$, $\delta$, and the regularization parameter $\lambda$. 
Spearmint runs are performed for both OSI and MSI. One needs to specify the boundaries of the hyper-parameter search space. We therefore start by running Spearmint for different boundary settings, to find reasonable areas for the hyper-parameter search. Correct order of magnitudes for the boundaries can also be estimated from previously published results and physical arguments.
For different boundary values, the optimum for $p$ always converged to a value close to 1. Because a smaller number of variables implies a smaller search space, especially in high dimensions, we set $p=1$ in our final run.

\begin{figure}[!t]
\centering
\includegraphics[width=0.5\columnwidth]{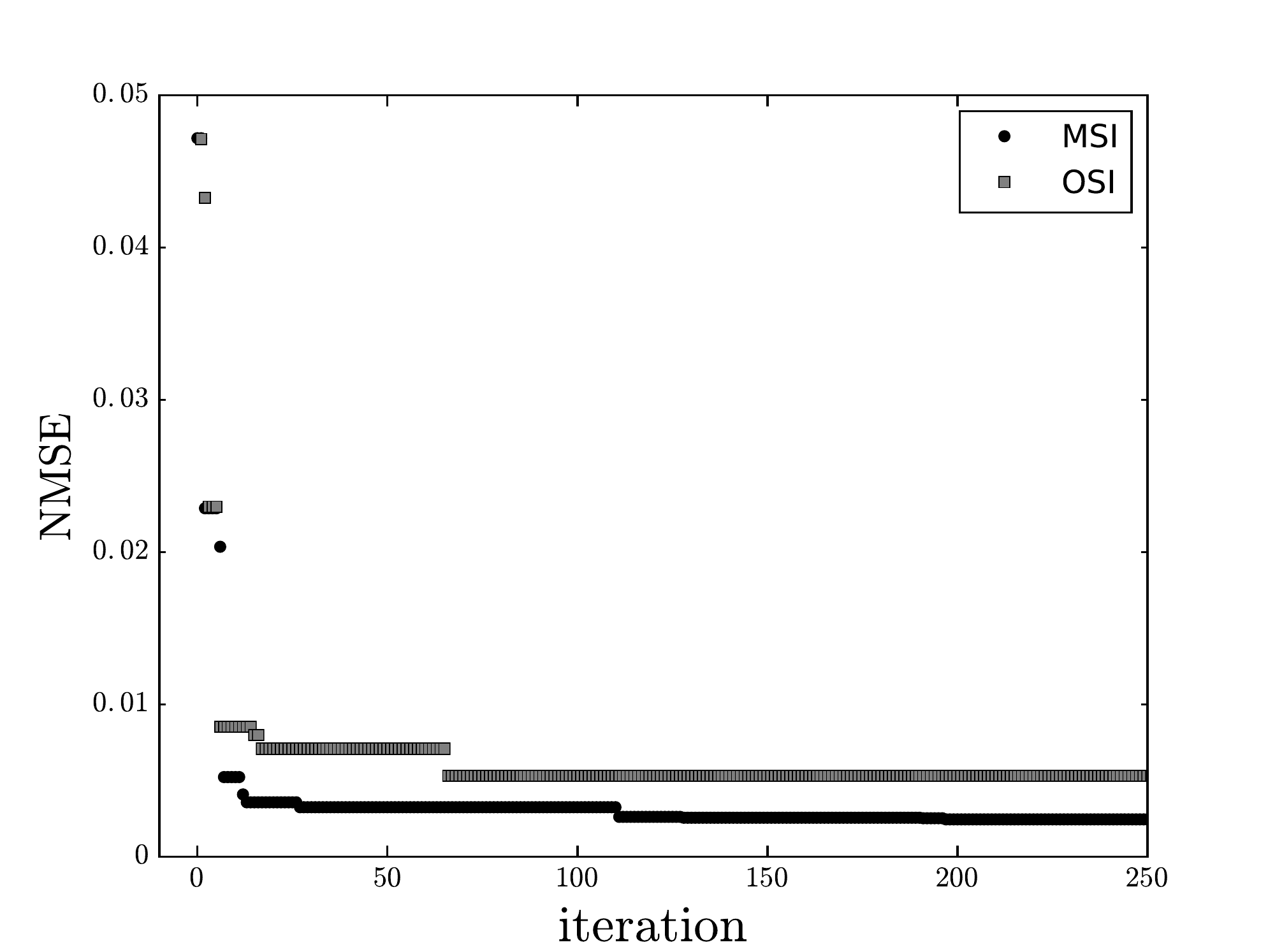}
\caption{NMSE as a function of the number iterations of the algorithm, for the Mackey-Glass NDN on the Santa Fe laser task.}
\label{fig:sfspear}
\end{figure}

The final results are shown in Table \ref{tab:santafe}.
The NMSE for the MSI is an order of magnitude lower than \cite{appeltant2012reservoir}, while the OSI result is four times lower. A plot of the NMSE on the validation set as a function of the number of Spearmint iterations is shown in Figure \ref{fig:sfspear}. If we start with good hyper-parameter boundaries, the first runs show a strong decrease in error, after which further improvements come in steps. This behaviour qualitatively resembles the exponential convergence speed found in analytical work, cf.~Section \ref{sec:GP}. It should be noted that different runs with small changes in the boundaries of the search space lead to parameter sets that are quite different, but have the same performance.

Some error surfaces are shown in \ref{app:sfhm}. One sees that the error surface is smooth around the optimum parameter values for both OSI and MSI. The error surfaces of the OSI and MSI differ significantly.
The error surface is non-stationary. For example, the variation in the $\gamma$ direction is low near the optimum, Figure \ref{fig:santafe_gammap}, while it is high away from the optimum, Figure \ref{fig:santafe_gammap_bad}. A difference in spatial variation between different parameters can be seen in Figure \ref{fig:santafe_gammap_MSI}, where $\gamma$ exhibits a slow variation and $p$ a faster variation. This difference in variation is taken care of by Spearmint, as discussed in Section \ref{sec:GP}. Note that $\delta$ has little to no influence on the performance.
Given the smoothness of the error surface, we expect good convergence behaviour from the GP, as is witnessed in the results. 
% From the demo, we see that uninteresting regions are discarded with one measurement. This is a strong point of the GP compared to random or grid search. It gives a qualitative understanding of the convergence behaviour observed in Figure \ref{fig:sfspear}.

\begin{table}
\begin{tabular}{l | c c c}
  & OSI & MSI & Appeltant \cite{appeltant2012reservoir} \\ \hline
  NMSE & 0.0047 & 0.0027 & 0.02 \\
  $p$ & 1 & 1 & 1 \\
  $\gamma$ & 0.018 & 0.0025 & 0.001 \\
  $\eta$ & 1.26 & 2 & 0.4 \\
  $\theta$ & 0.77 & 0.44 & 0.2 \\
  $\delta$ & 1.01 & 2.42 & ? \\
  $\log_{10} \lambda$ & -6.58 & -12.32 & ? \\
\end{tabular}
\caption{Optimal hyper-parameters of the Mackey-Glass NDN, for the Santa Fe data set. The number of nodes is $N = 200$. Unknown parameter values are denoted by a question mark.}
\label{tab:santafe}
\end{table}

We now compare our optimal parameters to those from reference \cite{appeltant2012reservoir}. One should keep in mind that the integration schemes are slightly different from the implementation in \cite{appeltant2012reservoir}, even though Euler integration is also used there. Our programs are written in Python, while Matlab is used in \cite{appeltant2012reservoir}. For reference, our MSI method with Appeltant's parameters gives NMSE = 0.028, which is similar to his result NMSE = 0.02 (note that we don't know all the parameters of \cite{appeltant2012reservoir}, some are educated guesses).
All parameters differ at most one order of magnitude. It appears that higher values of $\gamma$ and $\eta$ are useful, although this seems to be specific to the Santa Fe task.
A more important conclusion is that the value of $\theta = 0.2$ is suboptimal. Larger $\theta$ values allow for better performance in most considered benchmark tasks. This is important, because the choice $\theta = 0.2$ has become popular in the literature \cite{larger2012photonic,soriano2015delay,appeltant2014constructing,soriano2013optoelectronic,ortin2015unified,brunner2013parallel,hicke2013information,nguimdo2014fast,nguimdo2015simultaneous,nguimdo2016reducing}.
A larger $\theta$ causes a longer calculation time for the MSI. The OSI calculation time is independent of $\theta$, and has the same order of magnitude improvement in NMSE.

In addition to the NDN, we studied an ESN to compare the performance of the Bayesian optimization algorithm to the results from \cite{wang2015optimizing}, who use binary PSO to optimize the ESN architecture. % Due to an irregularity in the data set, on which we elaborate further in \ref{app:sf}, we chose to adapt the partitioning as suggested by \cite{wang2015optimizing}, which puts the irregularity in the validation set.
They use $6400$ samples for training, $1700$ for validation and another $1700$ for testing. The optimized parameters were the input scaling (which is a scaling parameter of $\bm{W^{\textit{in}}}$ in Eq.~\eqref{eq:esnstate}), spectral radius $\alpha$ and the ridge regression parameter $\lambda$. The number of nodes in the network was set to $N=200$. The validation error was averaged over 10 independent realizations of the ESN. Using Bayesian optimization on these parameters we obtained a NMSE of $0.00694 \pm 0.00087$, which should be compared to $0.0284$ in \cite{wang2015optimizing} for their standard ESN, and $0.0143$ after applying their optimization algorithm. This is a significant improvement compared to their results. During testing an irregularity in the data set was found, on which we elaborate in \ref{app:sf}. In the partitioning we study, the irregularity is part of the validation set. It should be noted that the irregularity may have driven the optimization to a hyper-parameter set that minimizes the error caused by the irregularity, which isn't necessarily the optimal set for prediction on the test set, as no such deviations occur there. 
% This result uses /data/leuven/312/vsc31274/SpearExperiments/SantaFe/SF029/output/00000642.out
% and the software in the parent directory
In order to check the performance of the ESN on the Santa Fe laser task without the complications of the irregularity, we also performed some tests using the Appeltant division of the data set, as was used to test the performance of the delay node algorithm. The resulting NMSE was $0.00693 \pm 0.00097$. Even though the error is approximately the same as for the other data set division, the result is better since only 1000 points instead of 6400 points are used for training.

% This result uses /data/leuven/312/vsc31274/SpearExperiments/SantaFe/SF032/output/00000630.out
% and the software in the parent directory

\subsection{NARMA 10}

\begin{figure}[!t]
\centering
\includegraphics[width=0.5\columnwidth]{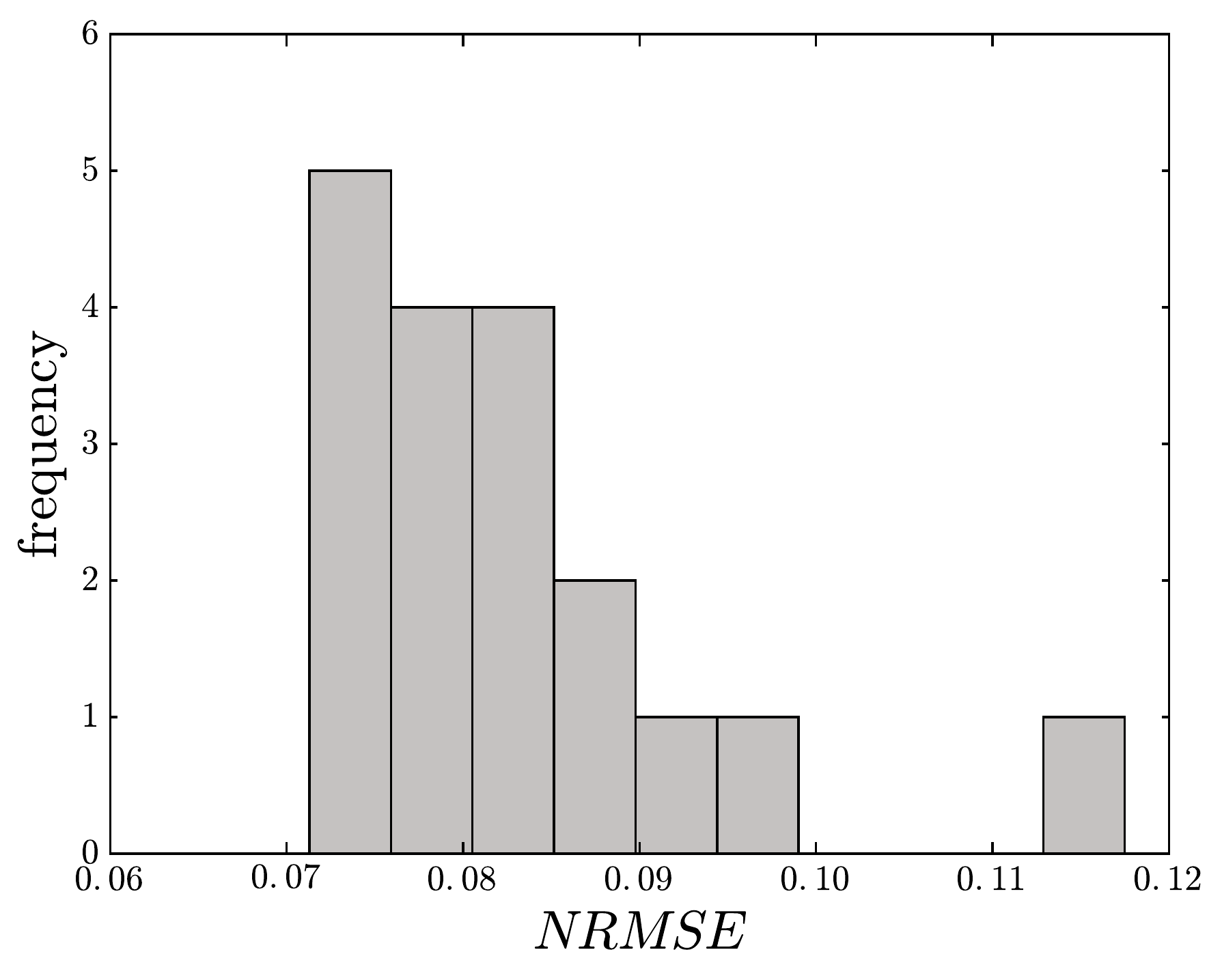}
\caption{Histogram of the NRMSE of 15 different NARMA 10 test sets of 2000 points each. The average is $\approx 0.083$.}
\label{fig:NARMAres}
\end{figure}

We used the same methods for the NARMA 10 time series. We train and validate on two fixed time series. The test error shows quite some variance, so we average the test error over 15 randomly generated time series of 2000 points each. We found an NRMSE $=0.08$ for the OSI scheme, which compares favorably to NRMSE $=0.12$ of Appeltant \cite{appeltant2012reservoir}. The histogram of test errors is shown in Figure \ref{fig:NARMAres}. Most values lie around the average, but there are some outliers with higher NRMSE. We don't know if such an average was taken in \cite{appeltant2012reservoir}. The MSI scheme gives NRMSE $= 0.146$, which is worse than the result from \cite{appeltant2012reservoir}. Note that $\theta \approx 0.2$ for the optimal MSI scheme (compared to $\theta = 0.908$ for OSI). The MSI scheme for NARMA 10 was used in \cite{appeltant2012reservoir} to arrive at the choice of $\theta = 0.2$.

Because NARMA 10 is deterministic given the ten previous input steps, the linear regression weights of the output are extremely large (a factor 100 larger than Santa Fe, which itself has large weights because it is deterministic). Large regression weights occur when overfitting, but in this case they give good results. The regularization parameter should be set to zero $\lambda=0$, i.e., no regularization is needed. This ``overfitting'' issue also leads to a strong sensitivity on noise in the implementation, making hardware implementations inherently difficult \cite{appeltant2012reservoir}. It is possible that rounding errors in the MSI scheme are the reason for the worse performance, which makes the calculation of reservoir states less stable.

From the plots of the error surfaces in \ref{app:narmahm} one sees that there is a very low sensitivity for $\delta$. The error surface is less smooth than the Santa Fe error surface, but still rather smooth. While the MSI scheme is indeed noisier compared to the OSI scheme, the amount of noise seems small enough for Spearmint to handle (this amount of noise was not problematic for the demo runs, see Section \ref{sec:spdem}). The fact that the MSI error surface is noisier corroborates our idea that the worse performance of the MSI scheme is because of rounding error in the integration scheme. Away from the optimal parameter values, the error surface can be very noisy, cf.~Figure \ref{fig:narma_gammaeta_bad}. We finally note that the error surfaces are non-stationary here as well.

\begin{table}
\begin{tabular}{l | c c c}
  & one-step & multi-step & Appeltant \cite{appeltant2012reservoir} \\ \hline
  NRMSE & 0.08 & 0.146 & 0.12 \\
  $p$ & 1.01 & 1.04 & 1 \\
  $\gamma$ & 0.0005 & 0.0013 & 0.005 \\
  $\eta$ & 0.738 & 0.753 & 0.4 \\
  $\theta$ & 0.908 & 0.19 & 0.2 \\
  $\delta$ & 1.05 & 3.812 & ? \\
  $\lambda$ & 0 & 0 & ? \\
\end{tabular}
\caption{Optimal hyper-parameters of the Mackey-Glass NDN, for the NARMA 10 time series. The number of nodes is $N = 400$. Unknown parameter values are denoted by a question mark.}
\label{tab:NARMA}
\end{table}

\subsection{Nonlinear channel equalization}

We now assess the performance on the nonlinear channel equalization task (Section \ref{sec:Nonlinear}). In order to compare with the results found in \cite{paquot2012optoelectronic} we use the delay node system with the sine nonlinearity, as described in Section \ref{sec:Sine}. Following the setup in \cite{paquot2012optoelectronic} we generate $10$ different realizations of the data set, for each of the $6$ SNRs (ranging from $12$ to $32$ dB). The delay node was trained using an initial washout of $200$ samples, followed by $3000$ points of training, and was validated on $10^6$ validation points. This was done for each of the $10$ realizations of the data set, and the validation error was the average of the error on the individual realizations. To produce the output, ridge regression was used, after which the resulting values were binned to the values $\{-3,-1,1,3\}$. % Subsequently these were compared to the actual values to produce the Symbol Error Rate (SER).
The reservoir consists of $N=50$ nodes. 

\begin{figure}[ht]
\centering
\includegraphics[width=0.75 \columnwidth]{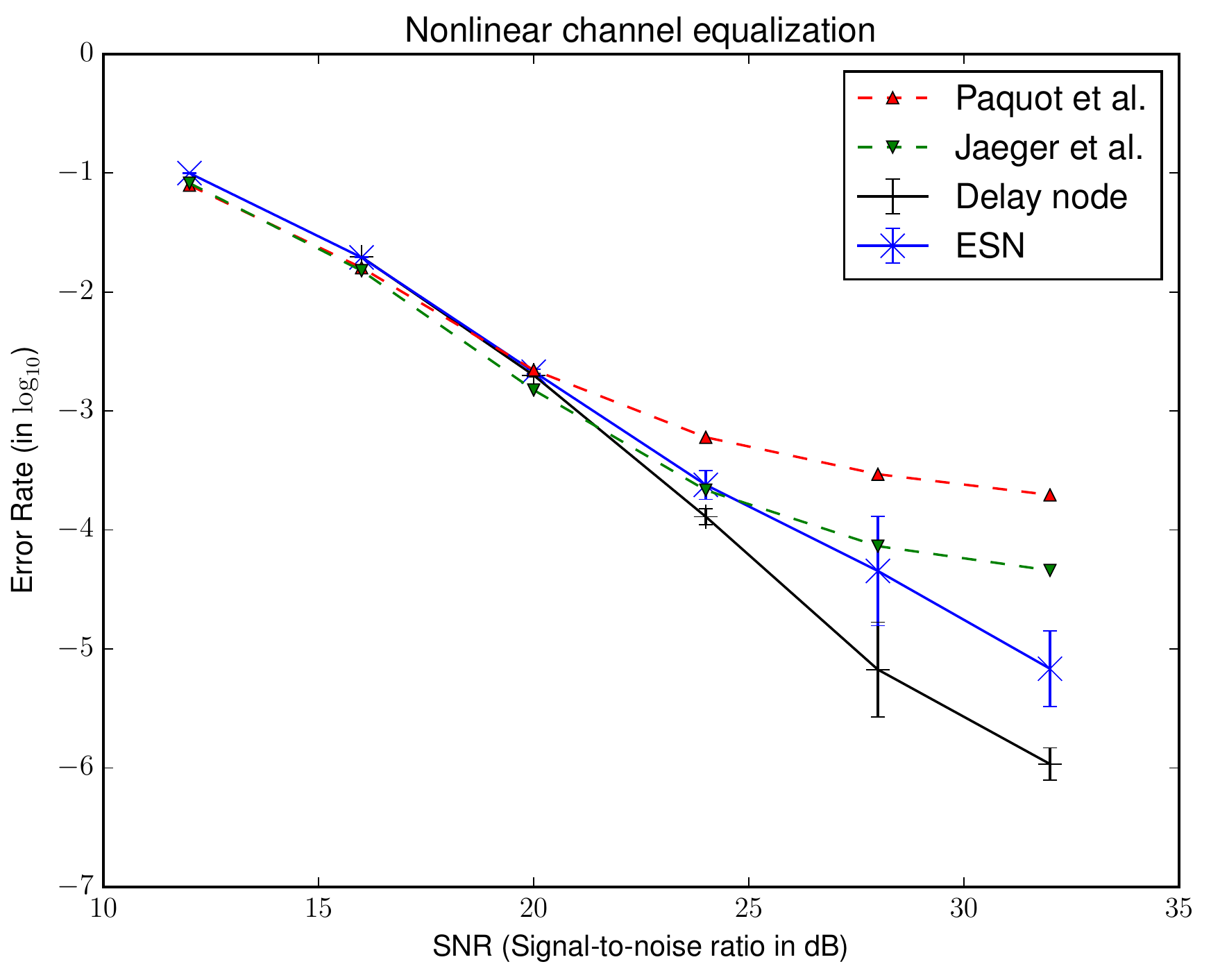}
\caption{Comparison with the literature on the the nonlinear channel equalization task. The results for the ESN architecture are displayed using cross markers, with the results of Jaeger et al.~\cite{jaeger2004harnessing} having dashed lines (green), and our results having solid lines (blue). The results of the sine delay node architecture are shown using plus markers, with the results of Paquot et al.~\cite{paquot2012optoelectronic} in dashed lines (red), and our results in solid line (black).}
\label{fig:nlc}
\end{figure}

The parameters that were optimized using Bayesian optimization are $\alpha$, $\beta$, $\gamma$ and the regularization parameter $\lambda$. These were optimized separately for each of the SNRs. The results are shown in Table \ref{tab:nonlin}. The results of the SER are plotted in Figure \ref{fig:nlc}. For low noise levels, Bayesian optimization finds parameters which result in a performance that is 2 orders of magnitude better than the literature. For high noise the performance is on par with \cite{paquot2012optoelectronic}. Presumably, the error surface at high levels of noise becomes less well behaved, and the assumption of smoothness breaks down, reducing the algorithm to random sampling. This is apparent at SNR 16, where the $\alpha$ parameter is larger by 2 orders of magnitude when compared to the other values of the SNR. 

We also used an ESN on this data set, and compare our results with \cite{jaeger2004harnessing}. Mimicking their setup we used an ESN with 47 nodes, used $200$ washout points, $5000$ training points and validated on $10^6$ points. The validation error was an average over the error on $10$ independent realizations of the network. Once the hyper-parameters were determined on the validation set, a test was performed on a test set of $10^6$ points, averaged over $100$ separate realizations of the network. This was done for $10$ instantiations of the data set, and the final result is the average of the performance on each of these data sets. The result is shown in Figure \ref{fig:nlc}. We observe a performance which is an order of magnitude better than \cite{jaeger2004harnessing} for high values of the SNR. Similar to the results with the NDN architecture with sine nonlinearity, our method converges to the results obtained in other works for the lower values of the SNR .

\begin{table}
\begin{tabular}{l | c c c c c}
  SNR (dB) & 16 & 20 & 24 & 28 & 32 \\ \hline
  $\alpha$ & 1.631484 & 0.076573 & 0.066186 & 0.090469 & 0.056437 \\
  $\beta$ & 0.001946 & 0.083608 & 0.082425 & 0.030920 & 0.091135\\
  $\phi$ & 0.086231 & 0.849972 & 0.934945 & 0.916553 & 0.238782 \\
  $\log_{10} \lambda$ & -12 & -11 & -13 & -15 & -11 \\
\end{tabular}
\caption{Hyper-parameters for the nonlinear channel equalization task using the sine delay node with $N=50$.}
\label{tab:nonlin}
\end{table}

% This result uses /data/leuven/312/vsc31274/SpearExperiments/ESN49-53 for the sine method and /data/leuven/312/vsc31274/SpearExperiments/Indiv/NLCESN09-13 for the ESN method.

\section{Spearmint demos}\label{sec:spdem}

\begin{figure}
\begin{subfigure}{0.42\textwidth}
\includegraphics[width=\linewidth]{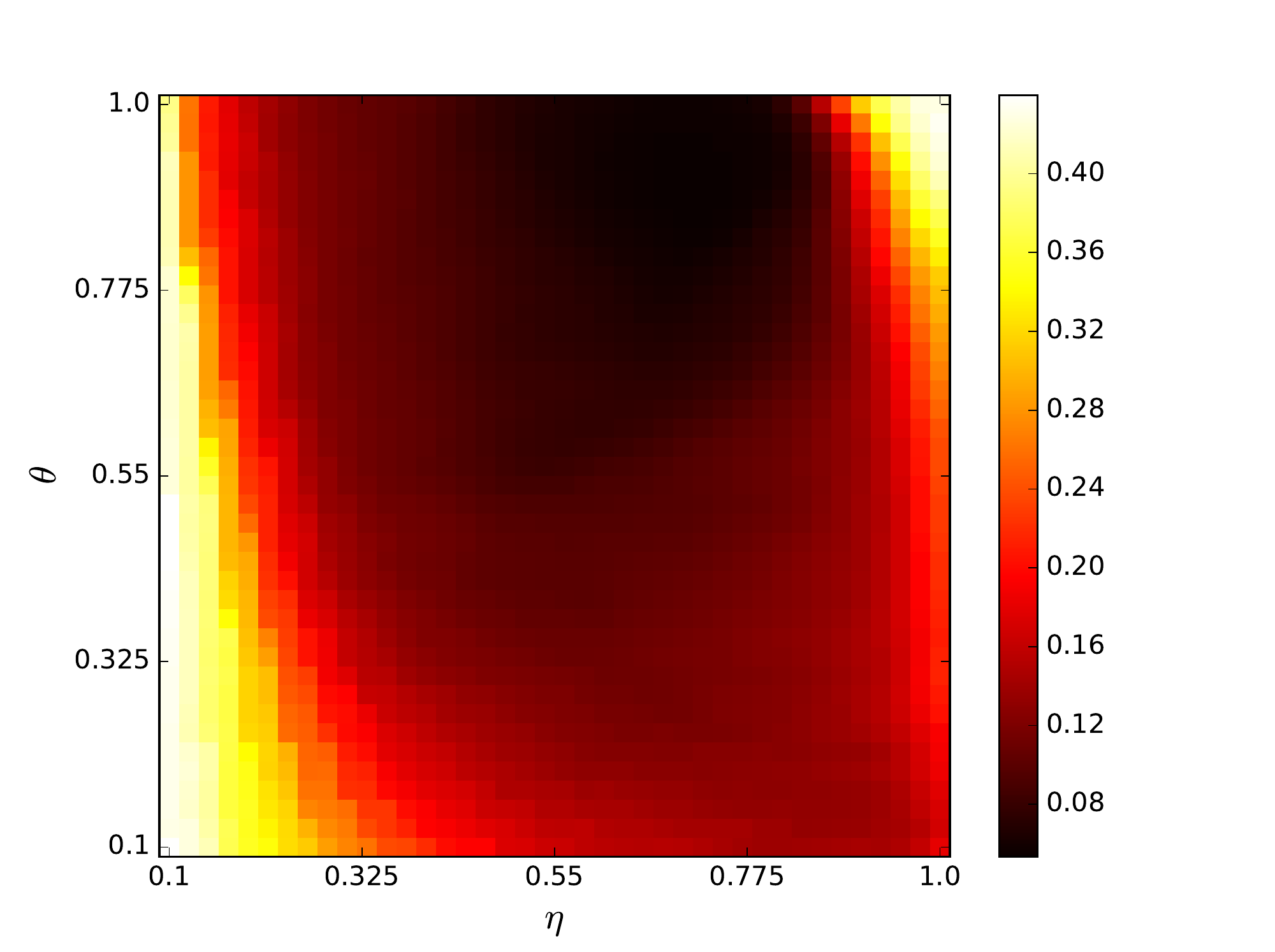}
%\caption{Caption of first subfigure}
\end{subfigure}
%\hspace*{\fill}
\begin{subfigure}{0.42\textwidth}
\includegraphics[width=\linewidth]{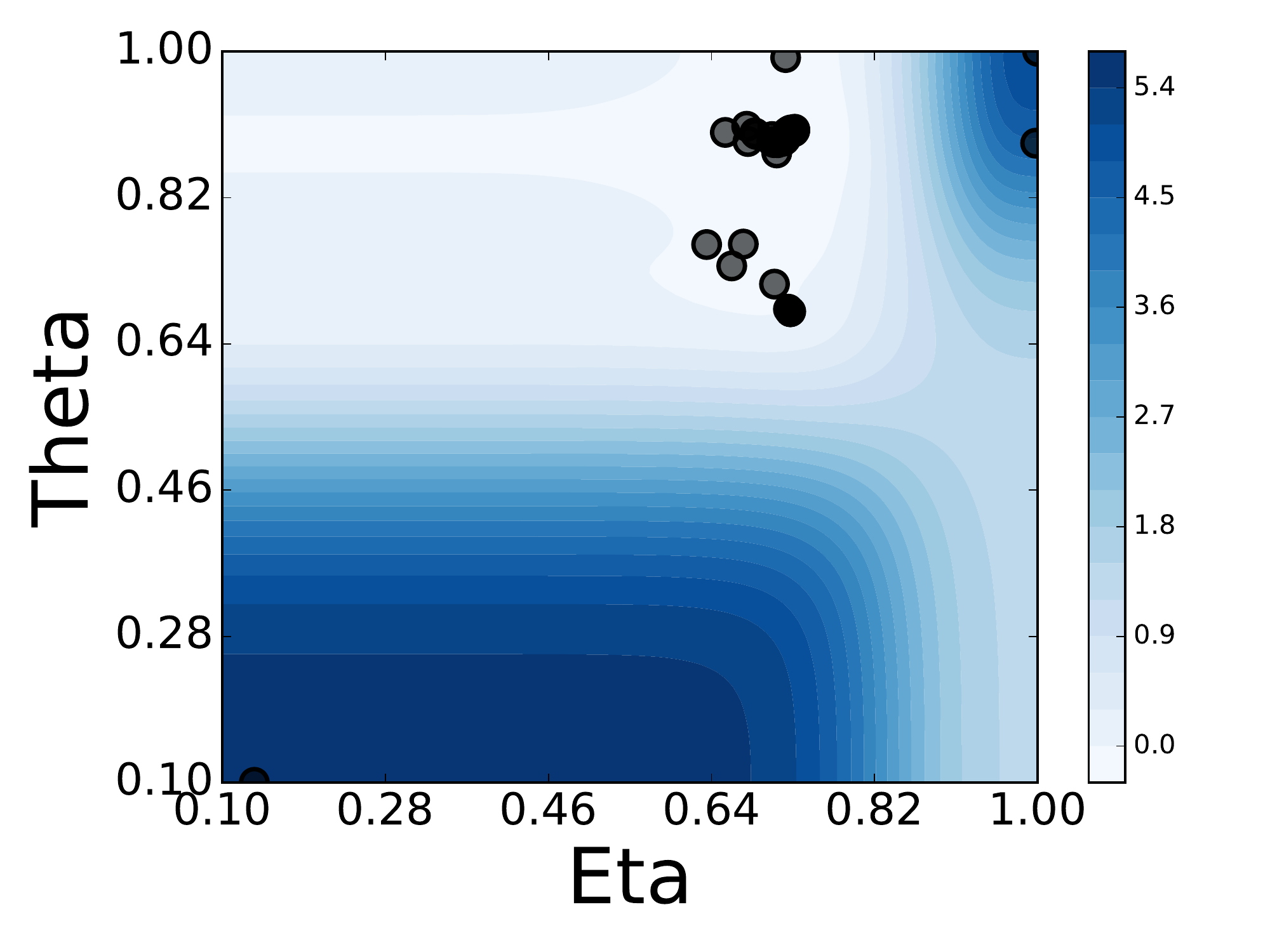}
%\caption{Caption of first subfigure}
\end{subfigure}
\caption{NARMA 10 task with MG NDN and OSI. (left) Grid plot of variables $\eta$ and $\theta$. (right) Spearmint search (black points are measurements).} \label{fig:narma_etatheta_spearmint}
\end{figure}

\begin{figure}
\begin{subfigure}{0.42\textwidth}
\includegraphics[width=\linewidth]{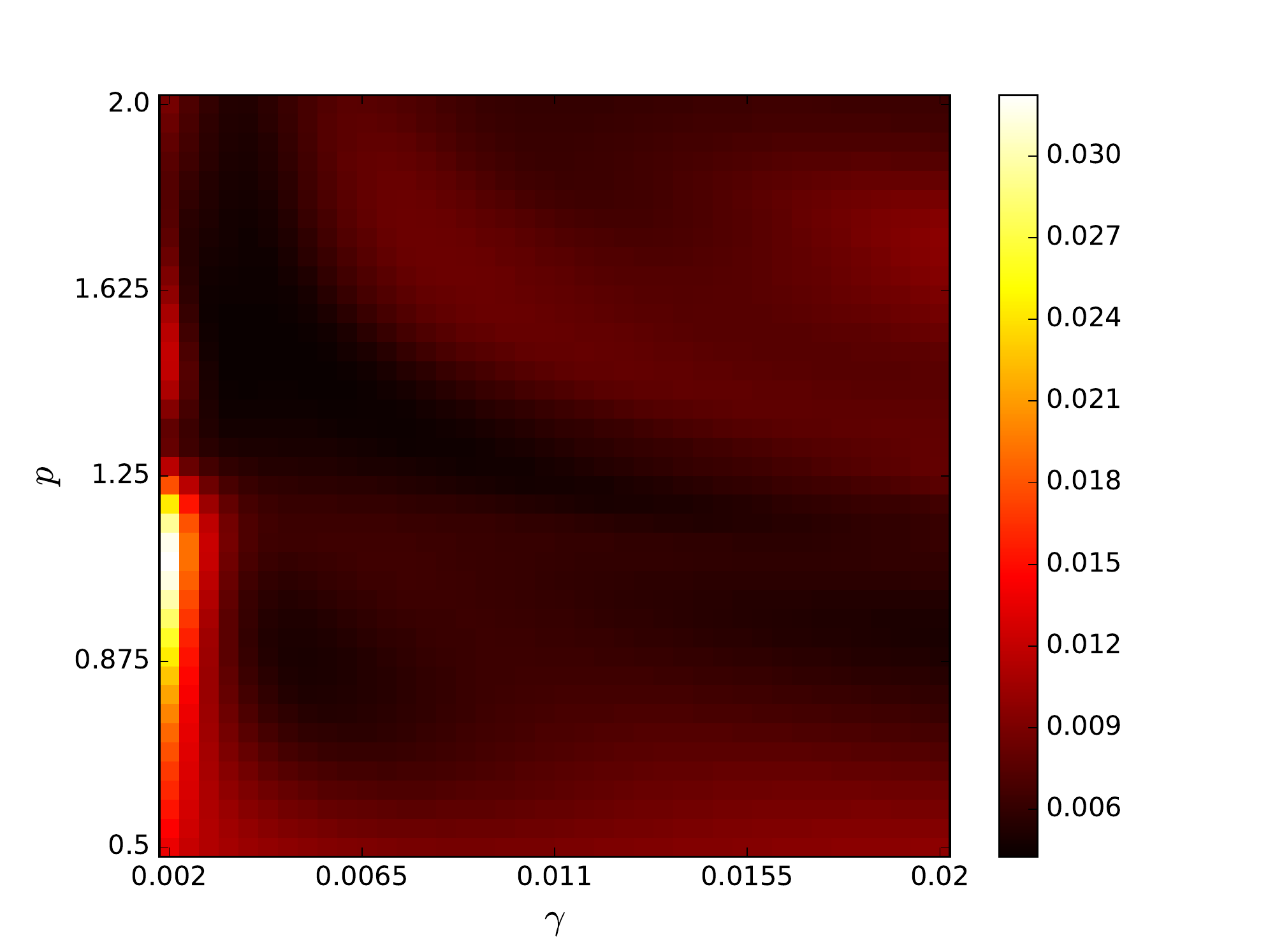}
%\caption{Caption of first subfigure}
\end{subfigure}
%\hspace*{\fill}
\begin{subfigure}{0.42\textwidth}
\includegraphics[width=\linewidth]{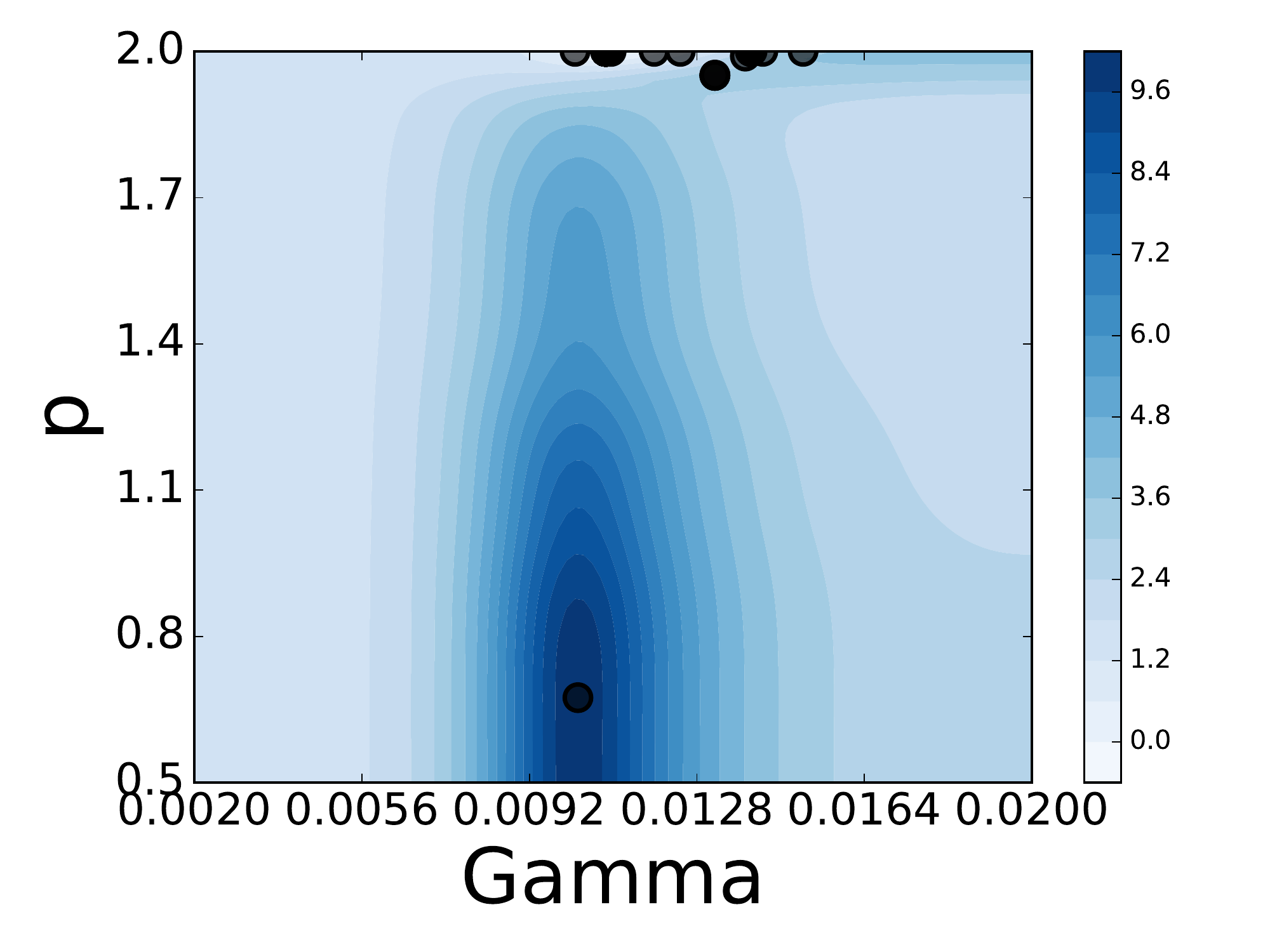}
%\caption{Caption of first subfigure}
\end{subfigure}
\caption{Santa Fe task with MG NDN and OSI. (left) Grid plot of variables $\gamma$ and $p$. (right) Spearmint search (black points are measurements).} \label{fig:santafe_gammap_spearmint}
\end{figure}

To get a better understanding of Spearmint's behaviour, we have visualized a few runs on two-dimensional parameter regions. Animations of the search process can be found in the supplementary material.
%Two runs are shown in Figures \ref{fig:narma_etatheta_spearmint} and \ref{fig:santafe_gammap_spearmint}, for the NARMA 10 and Santa Fe tasks, respectively. The other runs are shown in \ref{app:speardemo}.
One can conclude that Spearmint settles quite fast at a certain region. This region often contains the minimum, such as in Figure \ref{fig:narma_etatheta_spearmint} (see also Figures \ref{fig:NLC_sine_16}, \ref{fig:NLC_sine_32}, \ref{fig:NLC_esn_16}). In some cases, however, the region with the true minimum is missed, see Figure \ref{fig:santafe_gammap_spearmint}.
In Figure \ref{fig:NLC_esn_32} the region with the minimum is found, but is not fully investigated.
Spearmint does find the two regions with local minima in Figures \ref{fig:NLC_sine_16} and \ref{fig:NLC_sine_32} (the area at the bottom left and the strip above it). 
The fast fixation to a region indicates that the algorithm is quite ``eager'' to discard large hyper-parameter regions that it hasn't visited. This gives a qualitative understanding of the observed fast convergence behaviour: Spearmint discards large regions of the hyper-parameter space with only a few measurements, and after a while settles in a local minimum.

The estimate of the shape of the error surface is never really good, although in some cases a qualitative resemblance can be observed, cf.~Figures \ref{fig:narma_etatheta_spearmint} and \ref{fig:NLC_sine_32}. Sometimes there is almost no resemblance, cf.~Figure \ref{fig:santafe_gammap_spearmint}. This is because the algorithm has a strong focus on finding a good minimum as soon as possible; it simply discards regions that are not of interest instead of trying to model them well.

% For the nonlinear channel equalization task, we have found better results in the two-dimensional Spearmint searches with the starting parameters of the full runs. This means it could be advantageous to do specific searches at lower dimensions after the first runs. We have not found such improvements for the Santa Fe and NARMA 10 task, which is more smooth. This indicates that it is more advantageous to do extra Spearmint runs in lower dimensions if the error surface is less smooth.

\section{Conclusion and Outlook}\label{sec:conclusion}

To conclude, we have shown that Bayesian optimization of Gaussian processes is an excellent technique for automated hyper-parameter optimization in reservoir computing. 
The method was implemented with the Spearmint library. On all considered benchmarks, the method performed equally well or significantly better.

For the nonlinear delay nodes reservoirs, we showed that one does not need an exact solution of the differential equation to have a high-quality reservoir. Although this seems to be known, we are not aware of publications were this is explicitly discussed. We furthermore found that the popular choice of node distance $\theta = 0.2$ is often suboptimal. Because the algorithm is able to optimize over several parameters simultaneously, one can expect to find other such consistent optimal choices for other types of reservoirs.

Because the hyper-parameter search is automated, reservoir computing becomes a technique that requires little to no expert input. This could increase its appeal to machine learning practitioners. It remains to be seen whether this approach works well for all tasks and different types of reservoirs. We hope that the presented method is adopted by other researchers, so that a large variety of problems are studied.

Note that there is no optimization of other architectural features of the reservoir. The goal of this work is to compare the method with exactly the same architecture used in the literature. The methods presented here could possibly speed up the process of searching for new architectures, since the automated HO is able to find better results at a faster pace, without the need for a practitioner to acquire an intuition of the behaviour of the new architecture. 

It would be of interest to see the performance of Spearmint for HO of hardware implementations of an NDN/ESN \cite{larger2012photonic,paquot2012optoelectronic,brunner2013parallel,soriano2015delay,dong2016scaling,fischer2016photonic, duport2016fully, antonik2016towards, katumba2017multiple,soriano2015delay}, in which testing a hyper-parameter set could be more expensive compared to software implementations.

Another advantage of using Bayesian optimization to determine the hyper-parameters is that the comparison between different frameworks becomes less biased. When introducing a new framework it is not uncommon for researchers to put more effort into finding optimal parameters for the novel framework, than they do for the old method. When both methods are optimized using Bayesian optimization this bias becomes less of a factor.

\begin{appendix}

\section{Details Santa Fe implementation \label{app:sf}}

Appeltant \cite{appeltant2012reservoir} uses the first 4000 points of the data set: the first 1000 for training, the second for optimizing hyper-parameters, and the third for optimizing the regularization parameter $\lambda$. The NMSE is reported on the fourth set. We optimized the regularization parameter together with the other hyper-parameters on the second set, and measured the NMSE on the third and fourth set (last 2000 points).

One often uses initialization points to bring the reservoir to a stationary state. We use the first 200 points for initialization, and discard their reservoir states. So in fact we use 4200 points of the data set. We do not know if some initialization points are discarded in \cite{appeltant2012reservoir}. We checked that these small differences do not lead to qualitative differences in behaviour. 
The prediction to unseen data is robust, i.e., the first 1000 points and second 1000 points have similar errors. 

\begin{figure}[!t]
\centering
\includegraphics[width=0.75\columnwidth]{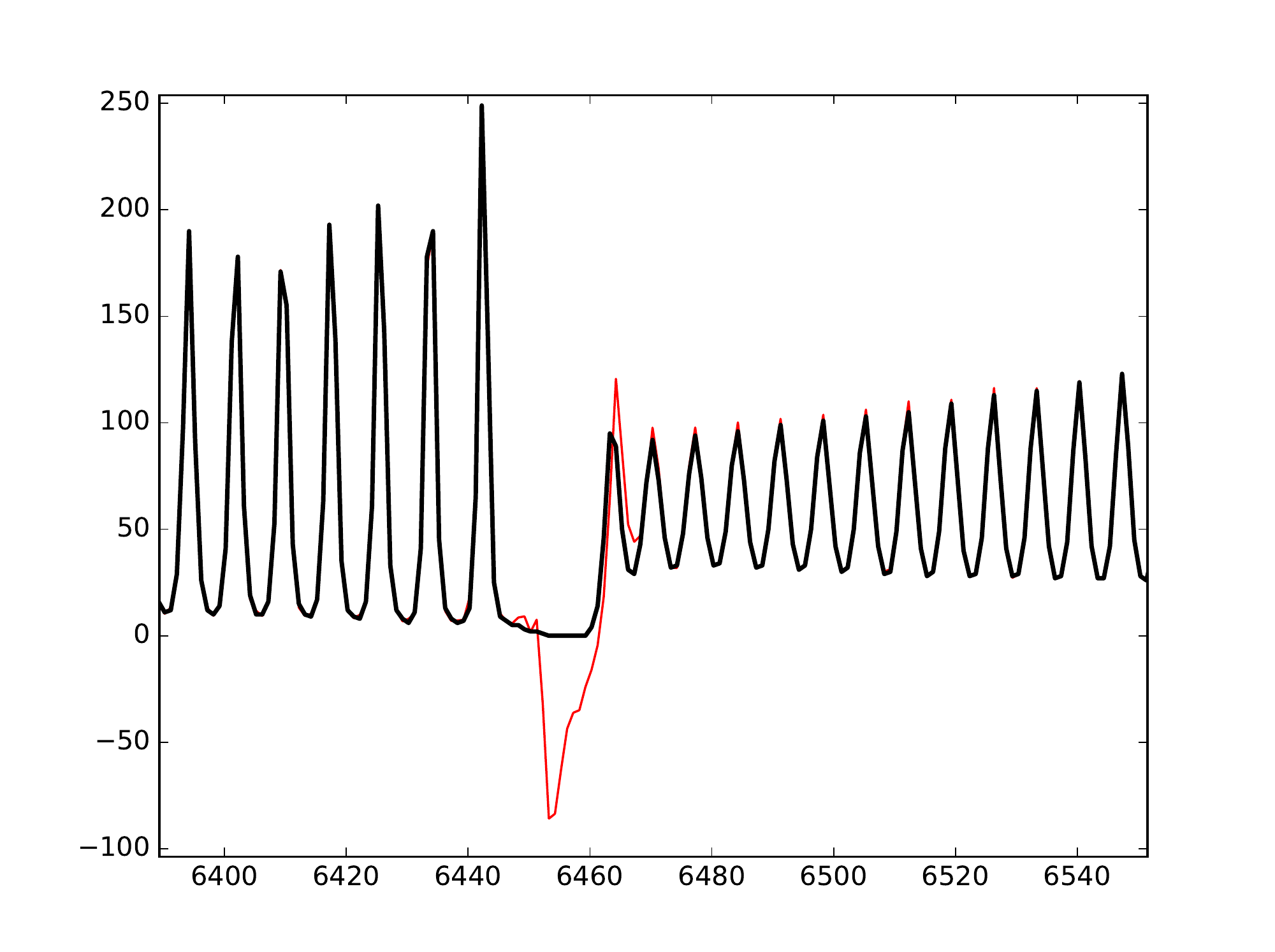}
\caption{The Santa Fe time series (thick black line) and the prediction of the MG NDN (red line). At points 6455-6461 the signal is exactly zero. As a result, the prediction diverges. The same divergent behaviour is observed for other types of networks.}
\label{fig:sfwrong}
\end{figure}

At points 6455-6461 in $A.cont$, the signal is exactly zero. This seems a measurement error, and the prediction of the reservoir diverges at those points, see Figure \ref{fig:sfwrong}. Several papers include this part of the data set in their test or validation set. In this case, the reservoir structure is not only optimized with respect to predicting the next point, but also towards a greater stability for a null input.
% This part of the data set is included in the test set of Tino and Rodan \cite{rodan2011minimum}. They do not mention this anomaly in the data set in their paper.

\section{Heat maps Santa Fe laser data and MG NDN}\label{app:sfhm}

We plot heat maps and one-dimensional slices of the error surface of the Santa Fe task. Parameter regions are near the optimal values unless stated otherwise.

\begin{figure}
\begin{subfigure}{0.32\textwidth}
\includegraphics[width=\linewidth]{figures/heatmap_santafe_gamma_p_flipud.pdf}
%\caption{Caption of first subfigure}
\end{subfigure}
\hspace*{\fill}
\begin{subfigure}{0.32\textwidth}
\includegraphics[width=\linewidth]{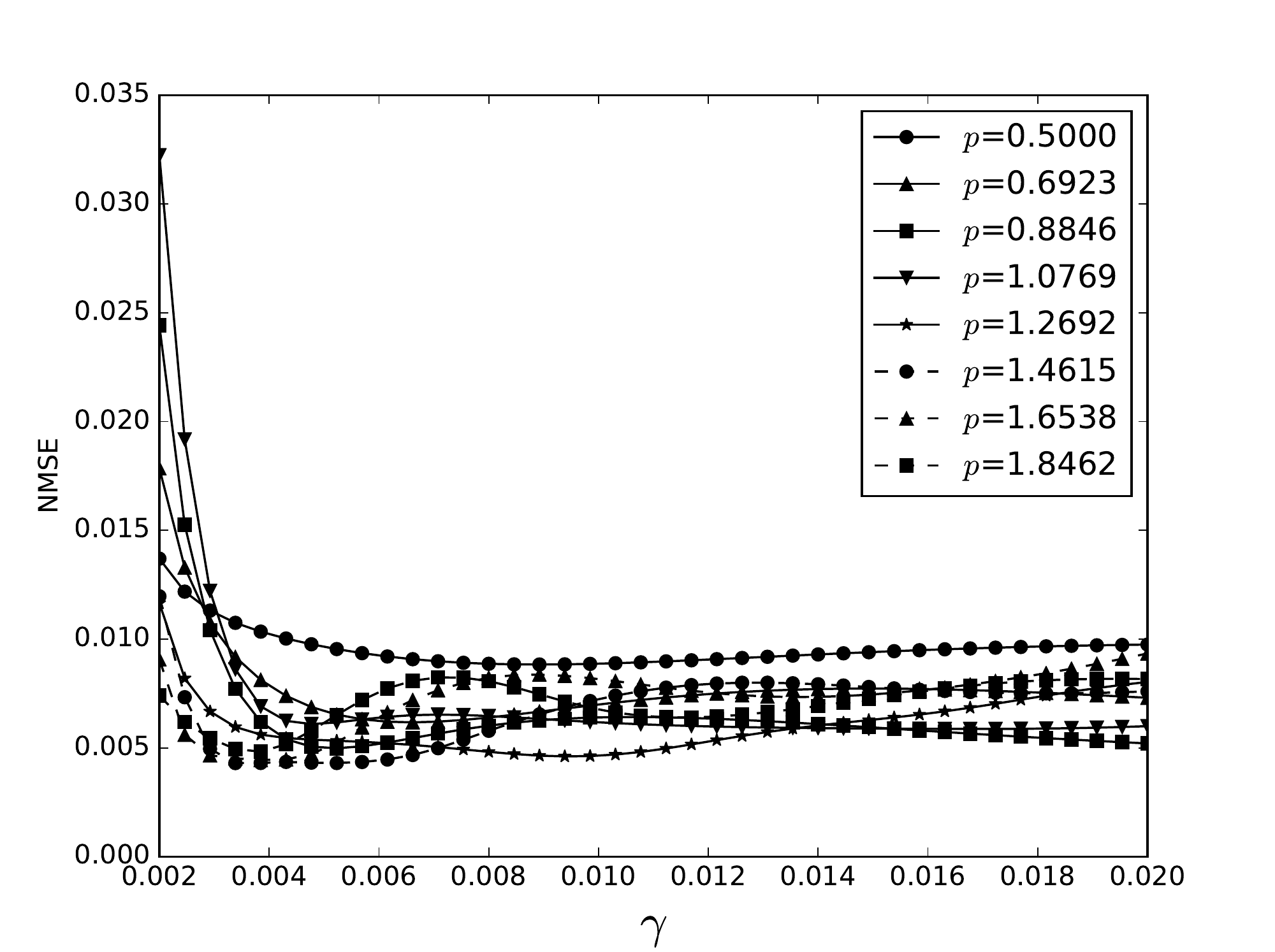}
%\caption{Caption of first subfigure}
\end{subfigure}
\hspace*{\fill}
\begin{subfigure}{0.32\textwidth}
\includegraphics[width=\linewidth]{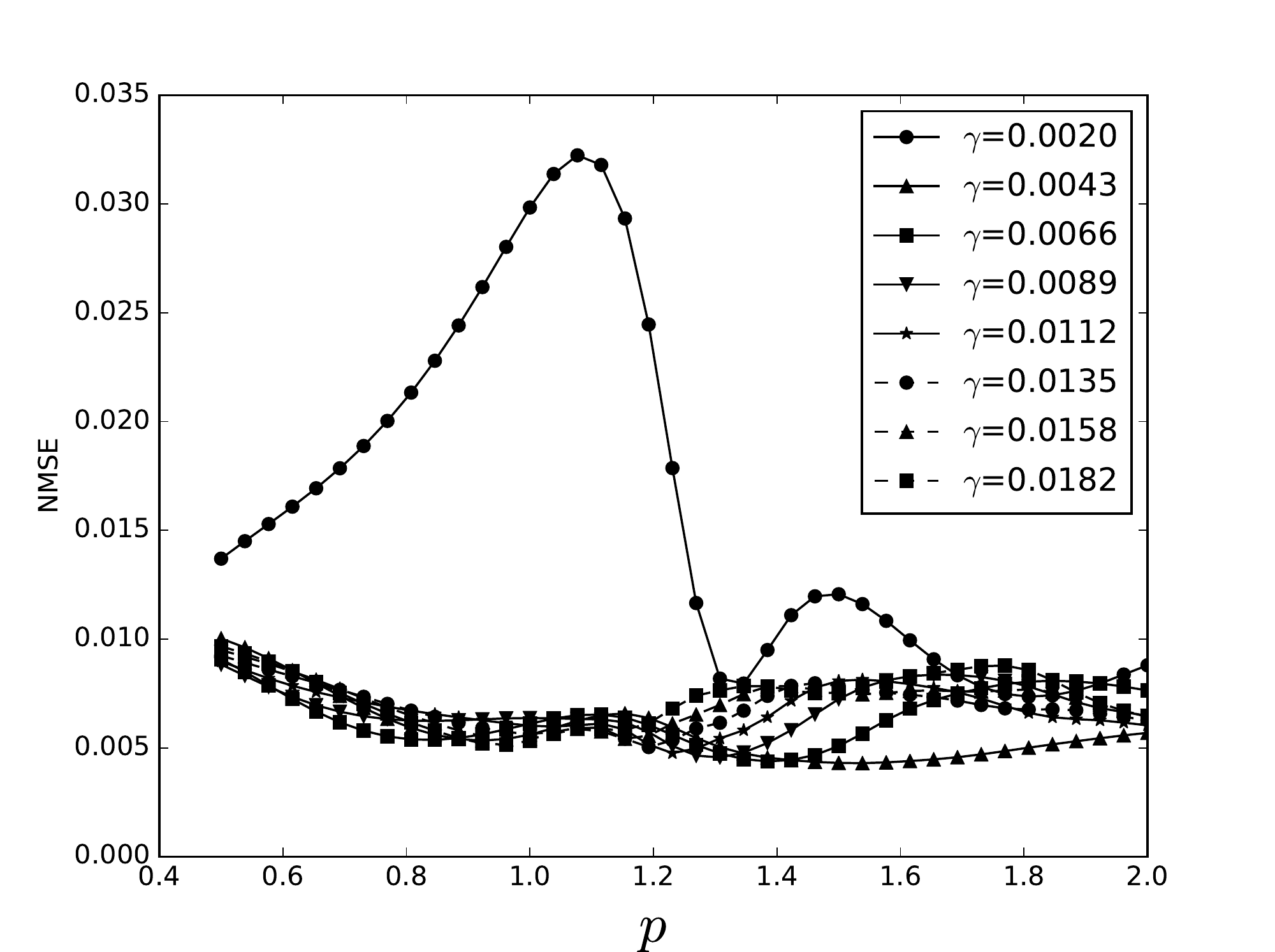}
%\caption{Caption of second subfigure}
\end{subfigure}
\caption{Santa Fe task with MG NDN and OSI, $\gamma$ and $p$.} \label{fig:santafe_gammap}
\end{figure}

\begin{figure}
\begin{subfigure}{0.32\textwidth}
\includegraphics[width=\linewidth]{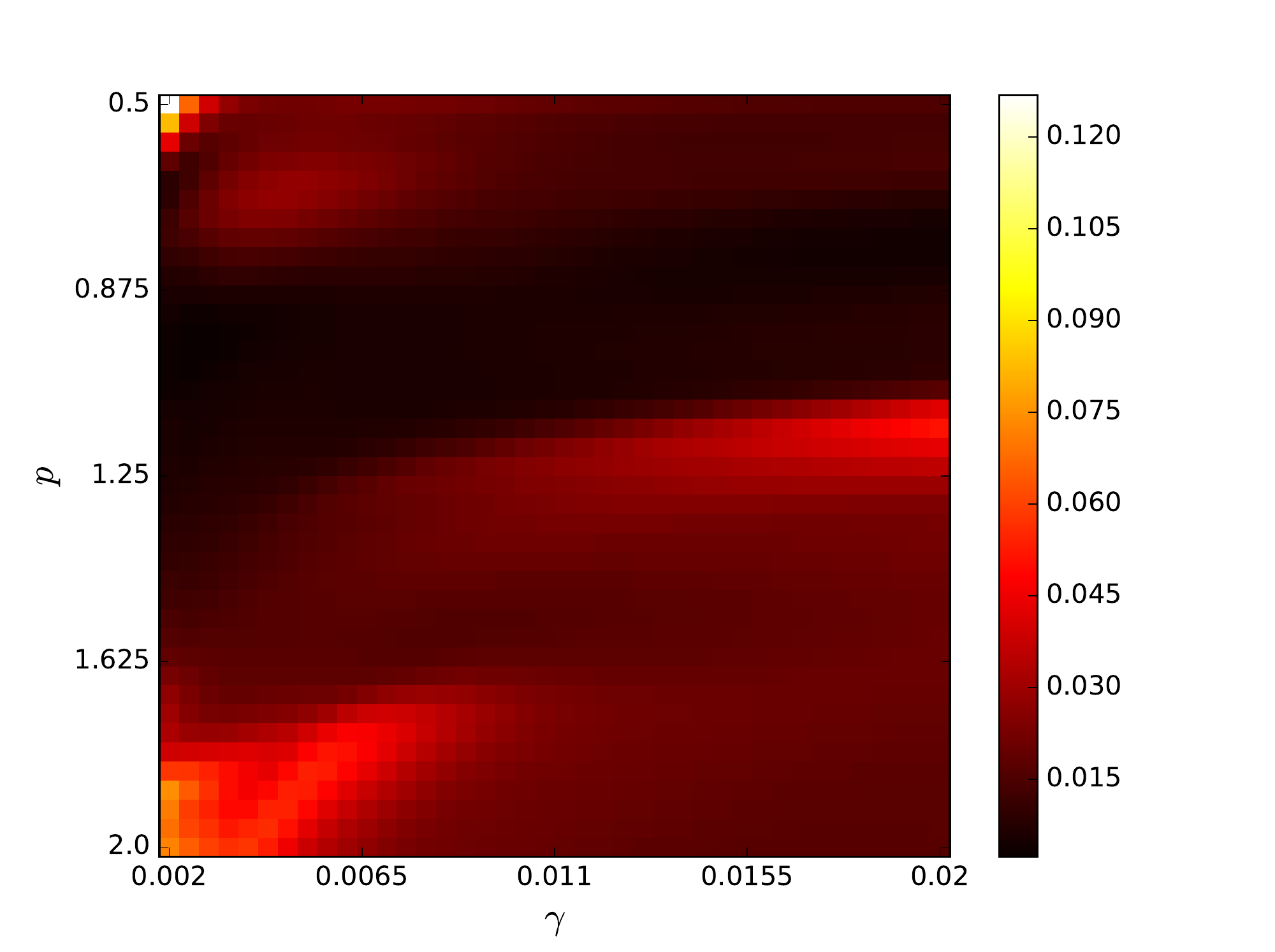}
%\caption{Caption of first subfigure}
\end{subfigure}
\hspace*{\fill}
\begin{subfigure}{0.32\textwidth}
\includegraphics[width=\linewidth]{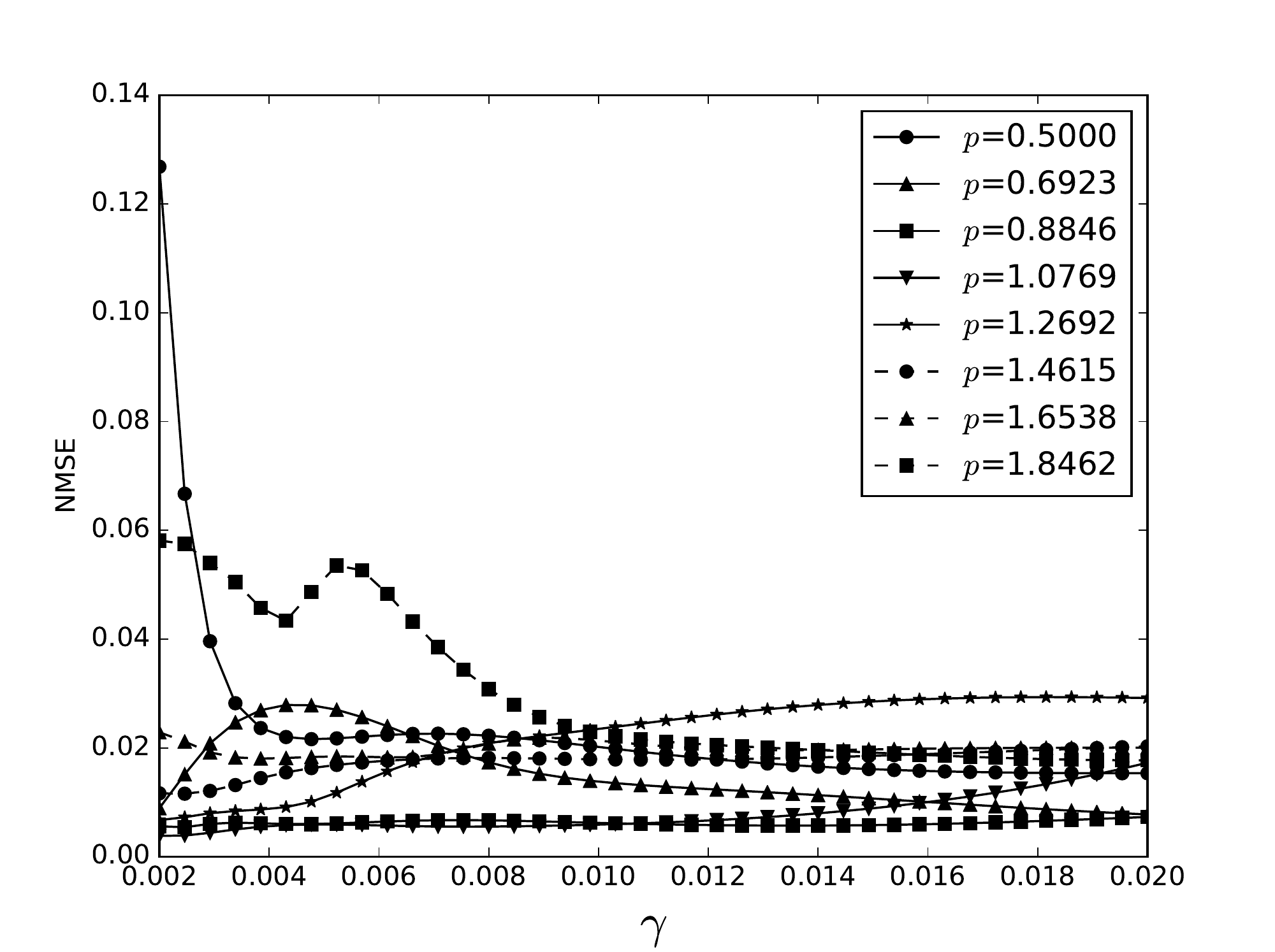}
%\caption{Caption of first subfigure}
\end{subfigure}
\hspace*{\fill}
\begin{subfigure}{0.32\textwidth}
\includegraphics[width=\linewidth]{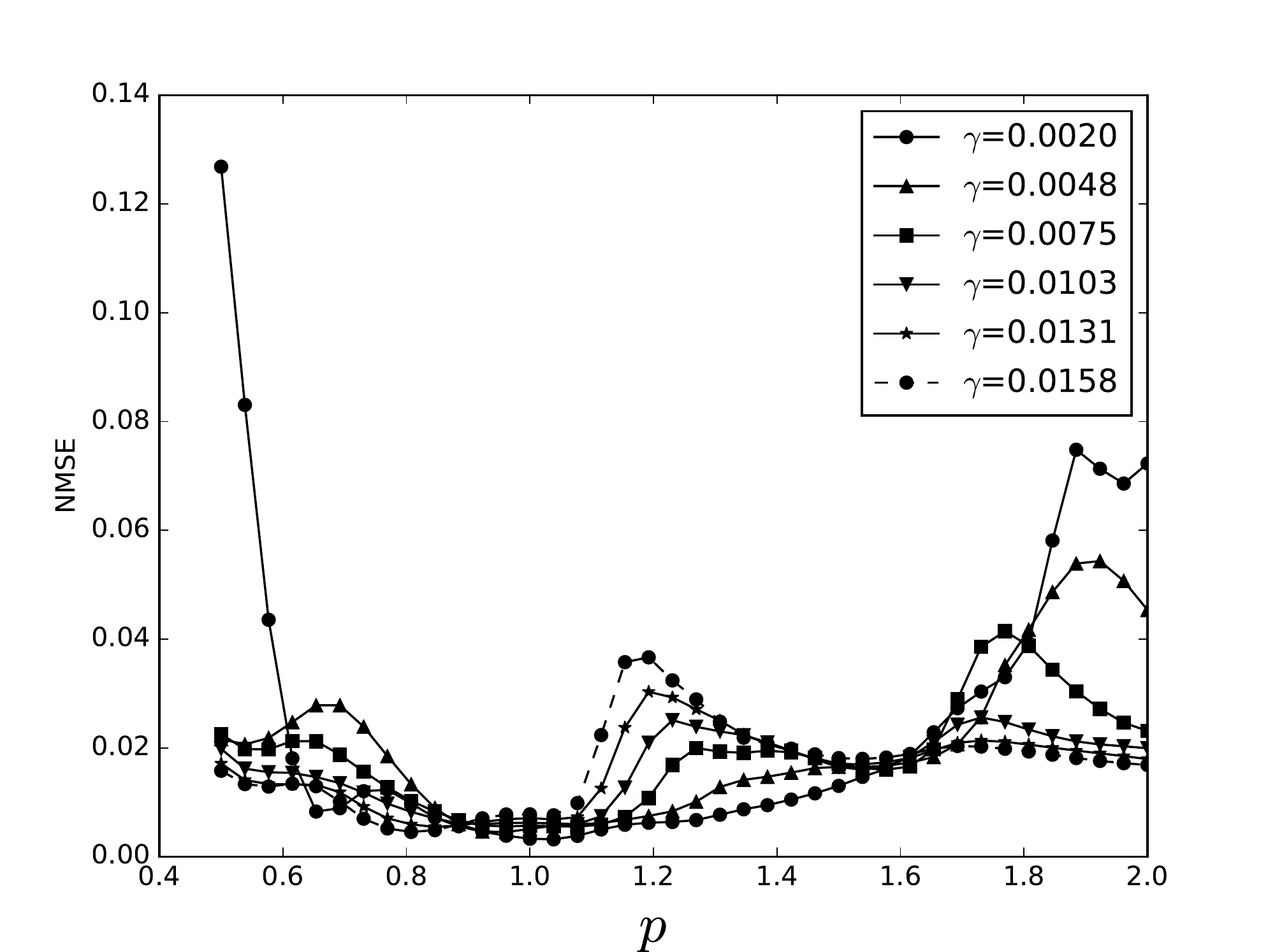}
%\caption{Caption of second subfigure}
\end{subfigure}
\caption{Santa Fe task with MG NDN and MSI, $\gamma$ and $p$.} \label{fig:santafe_gammap_MSI}
\end{figure}

\begin{figure}
\begin{subfigure}{0.32\textwidth}
\includegraphics[width=\linewidth]{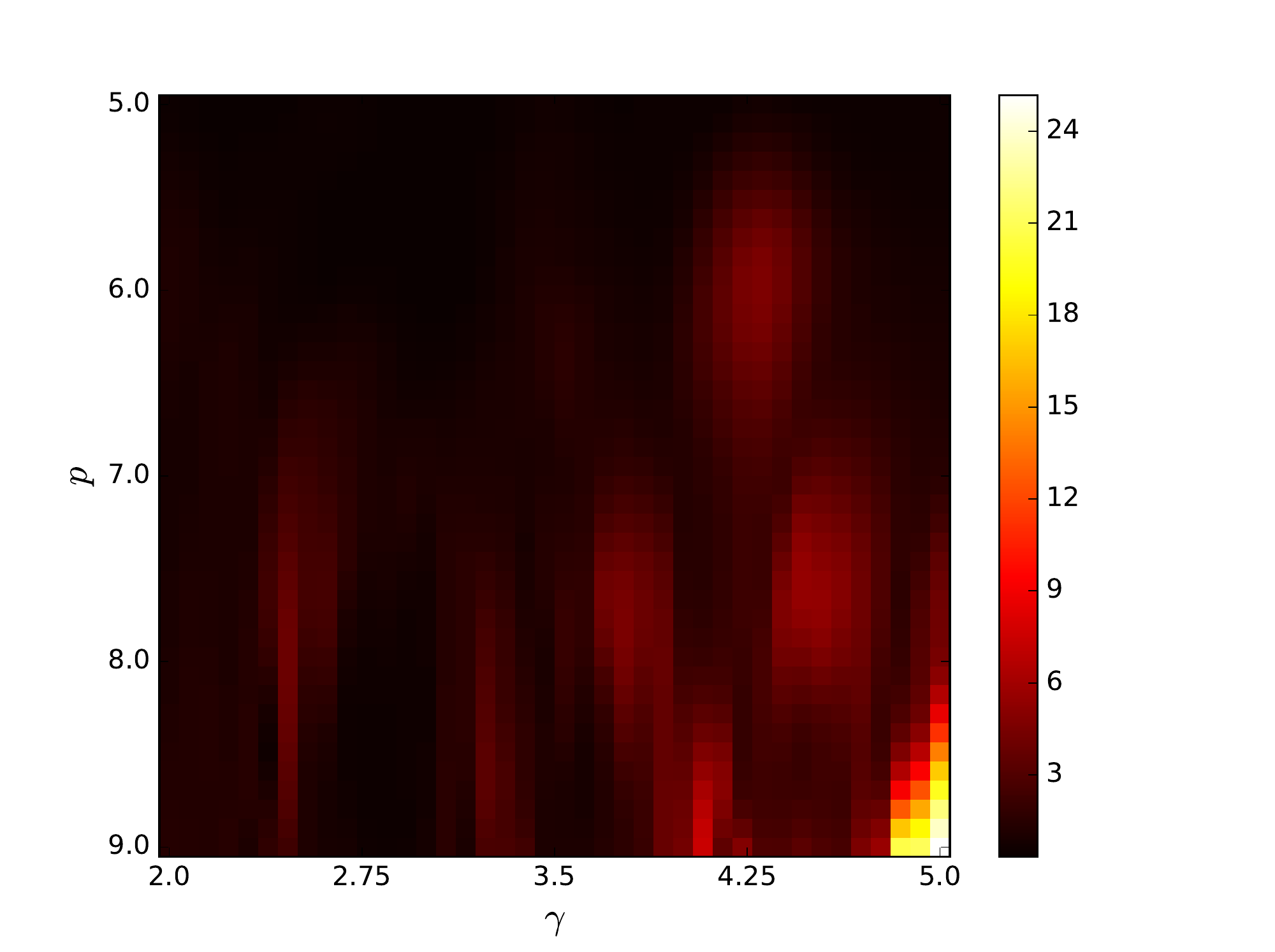}
%\caption{Caption of first subfigure}
\end{subfigure}
\hspace*{\fill}
\begin{subfigure}{0.32\textwidth}
\includegraphics[width=\linewidth]{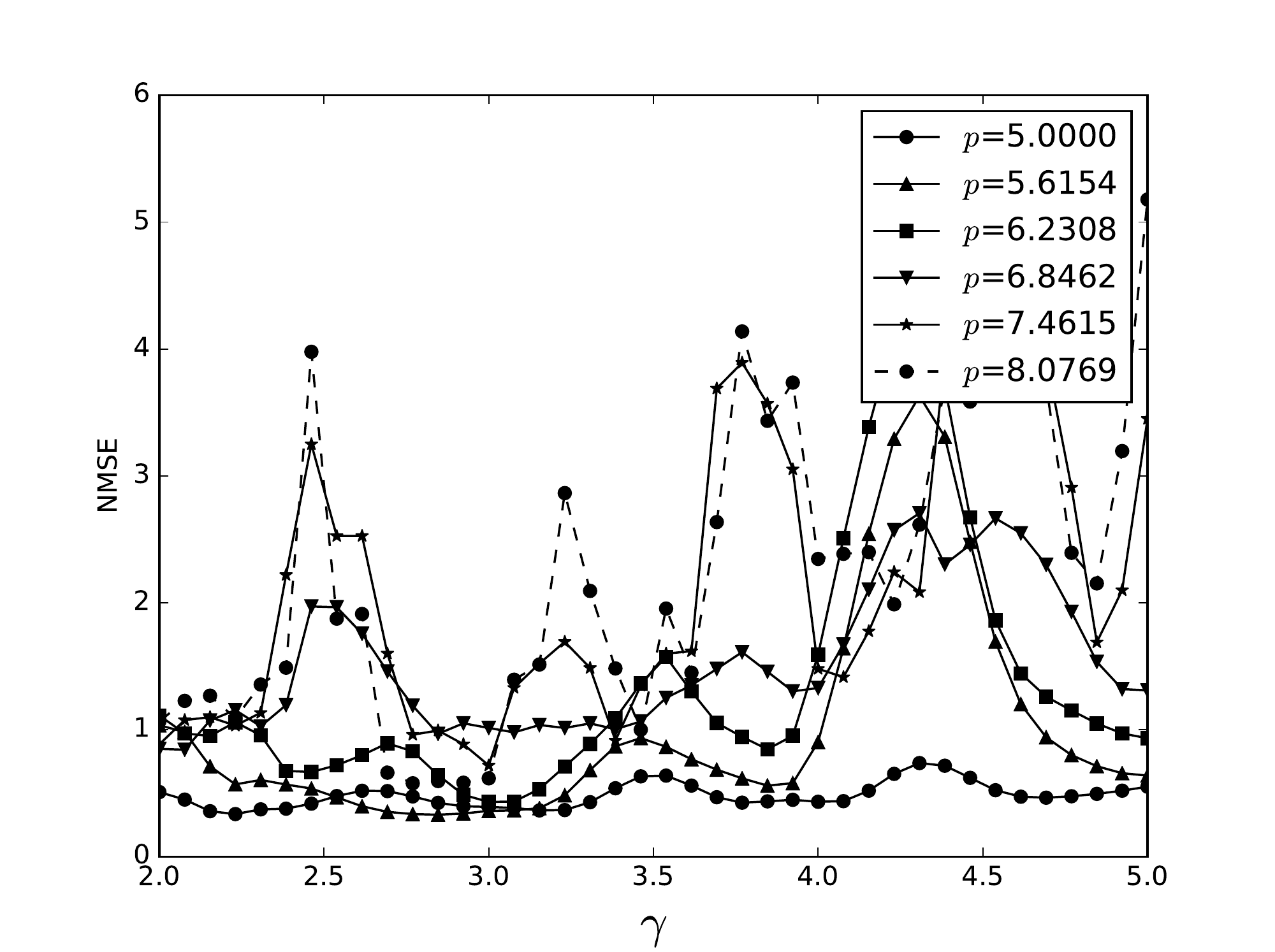}
%\caption{Caption of first subfigure}
\end{subfigure}
\hspace*{\fill}
\begin{subfigure}{0.32\textwidth}
\includegraphics[width=\linewidth]{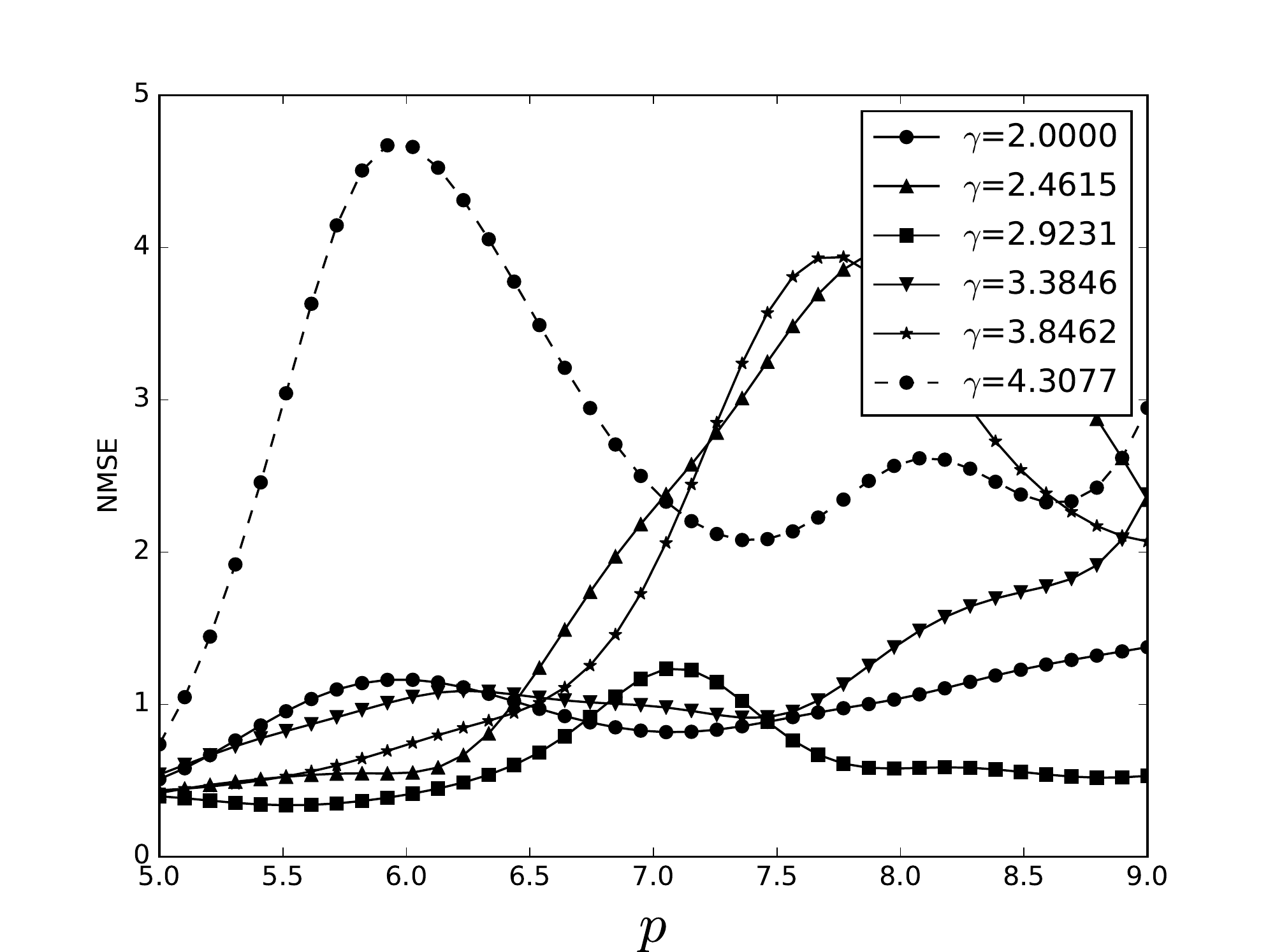}
%\caption{Caption of second subfigure}
\end{subfigure}
\caption{Santa Fe task with MG NDN and OSI, $\gamma$ and $p$ for a parameter region away from the optimum.} \label{fig:santafe_gammap_bad}
\end{figure}

\begin{figure}
\begin{subfigure}{0.32\textwidth}
\includegraphics[width=\linewidth]{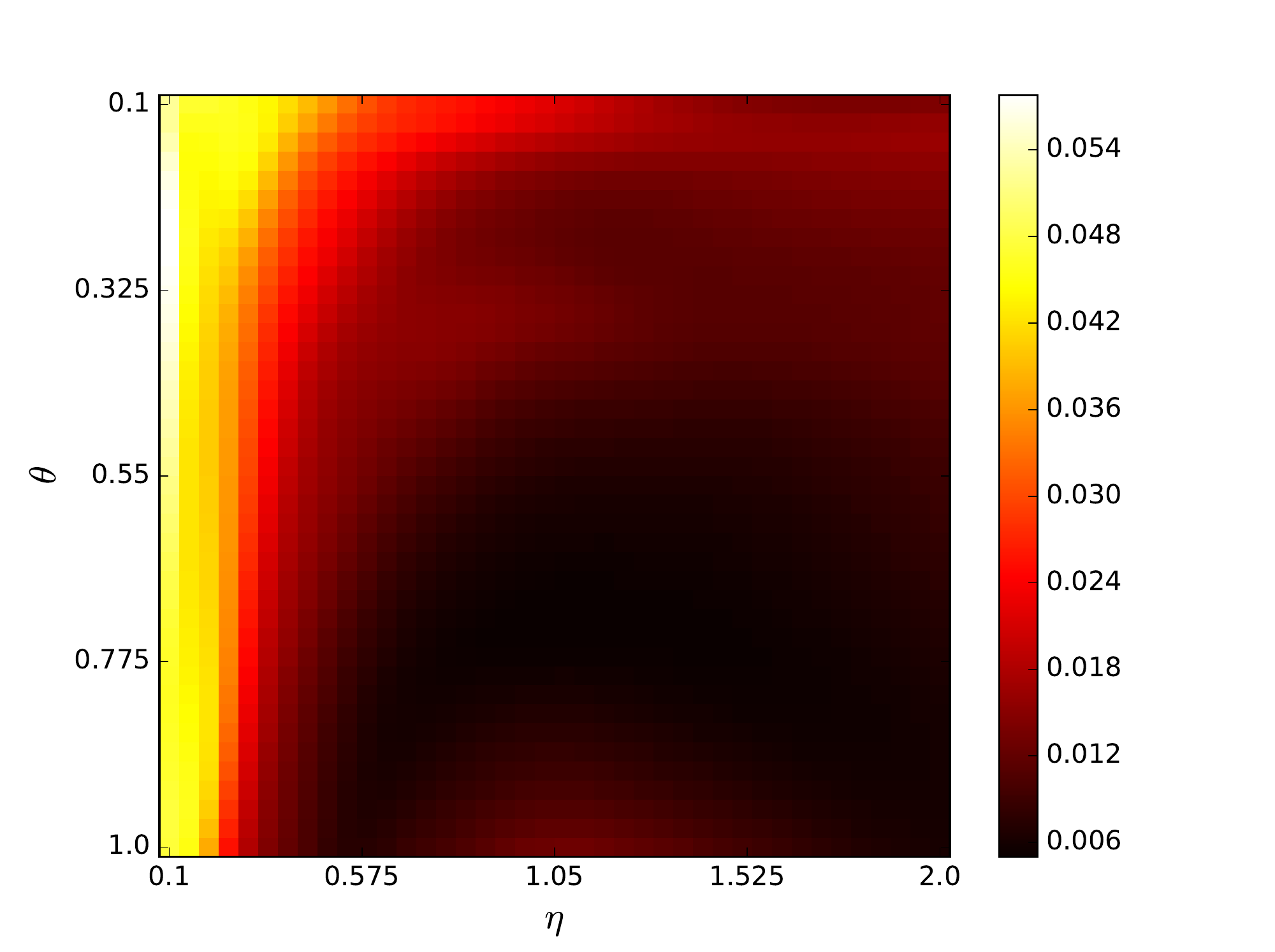}
%\caption{Caption of first subfigure}
\end{subfigure}
\hspace*{\fill}
\begin{subfigure}{0.32\textwidth}
\includegraphics[width=\linewidth]{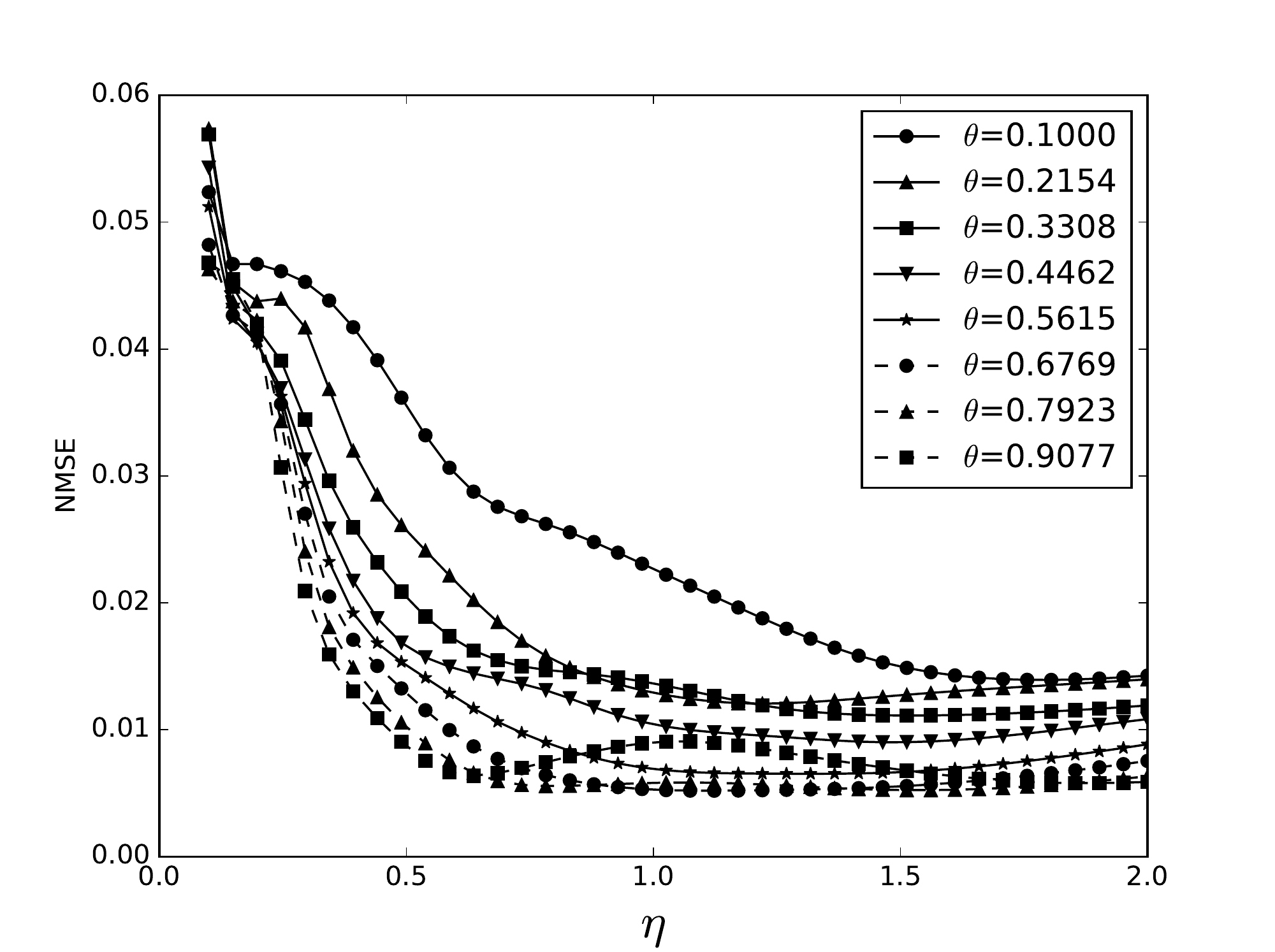}
%\caption{Caption of first subfigure}
\end{subfigure}
\hspace*{\fill}
\begin{subfigure}{0.32\textwidth}
\includegraphics[width=\linewidth]{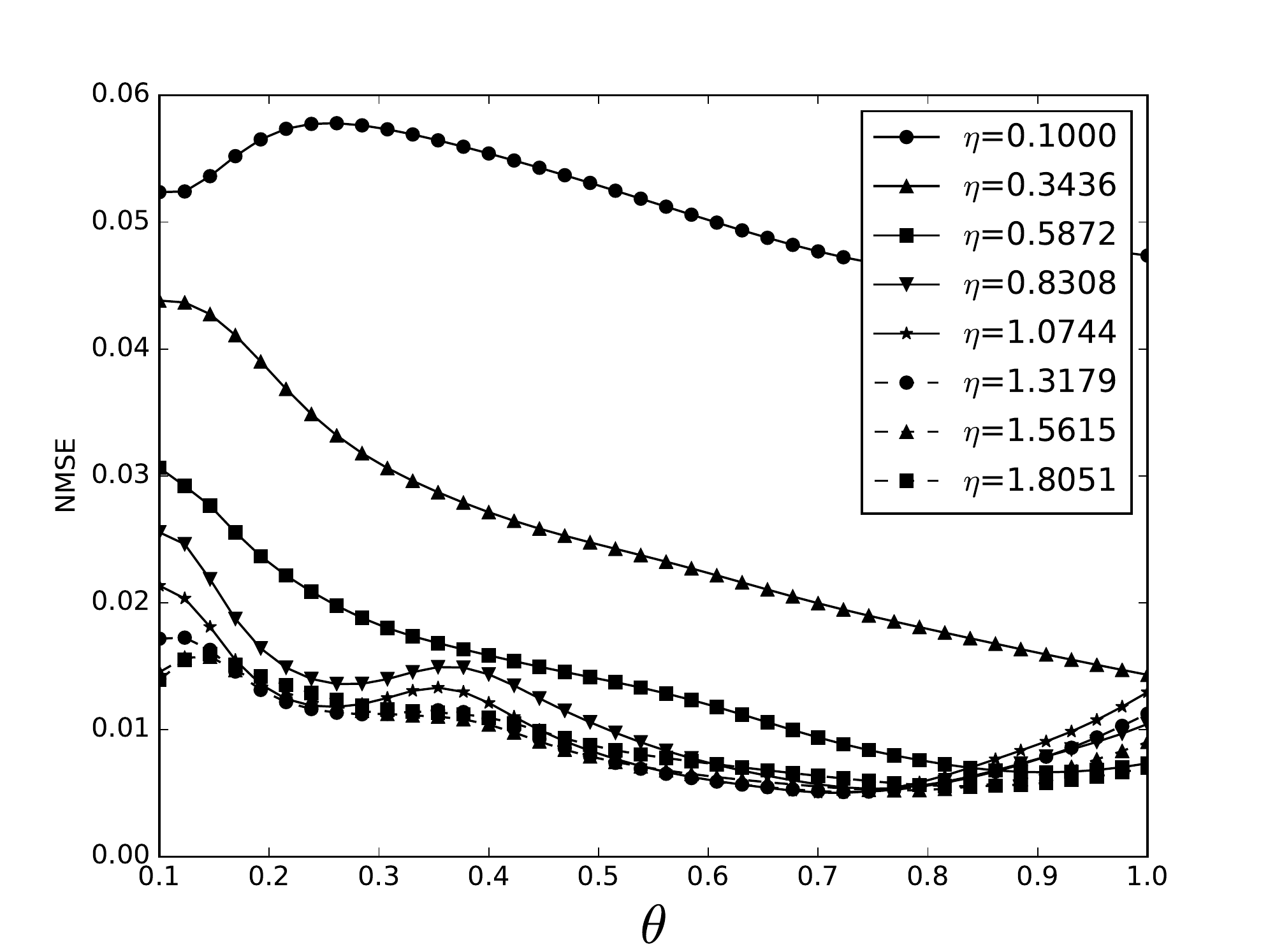}
%\caption{Caption of second subfigure}
\end{subfigure}
\caption{Santa Fe task with MG NDN and OSI, $\eta$ and $\theta$.} \label{fig:santafe_etatheta}
\end{figure}

\begin{figure}
\begin{subfigure}{0.32\textwidth}
\includegraphics[width=\linewidth]{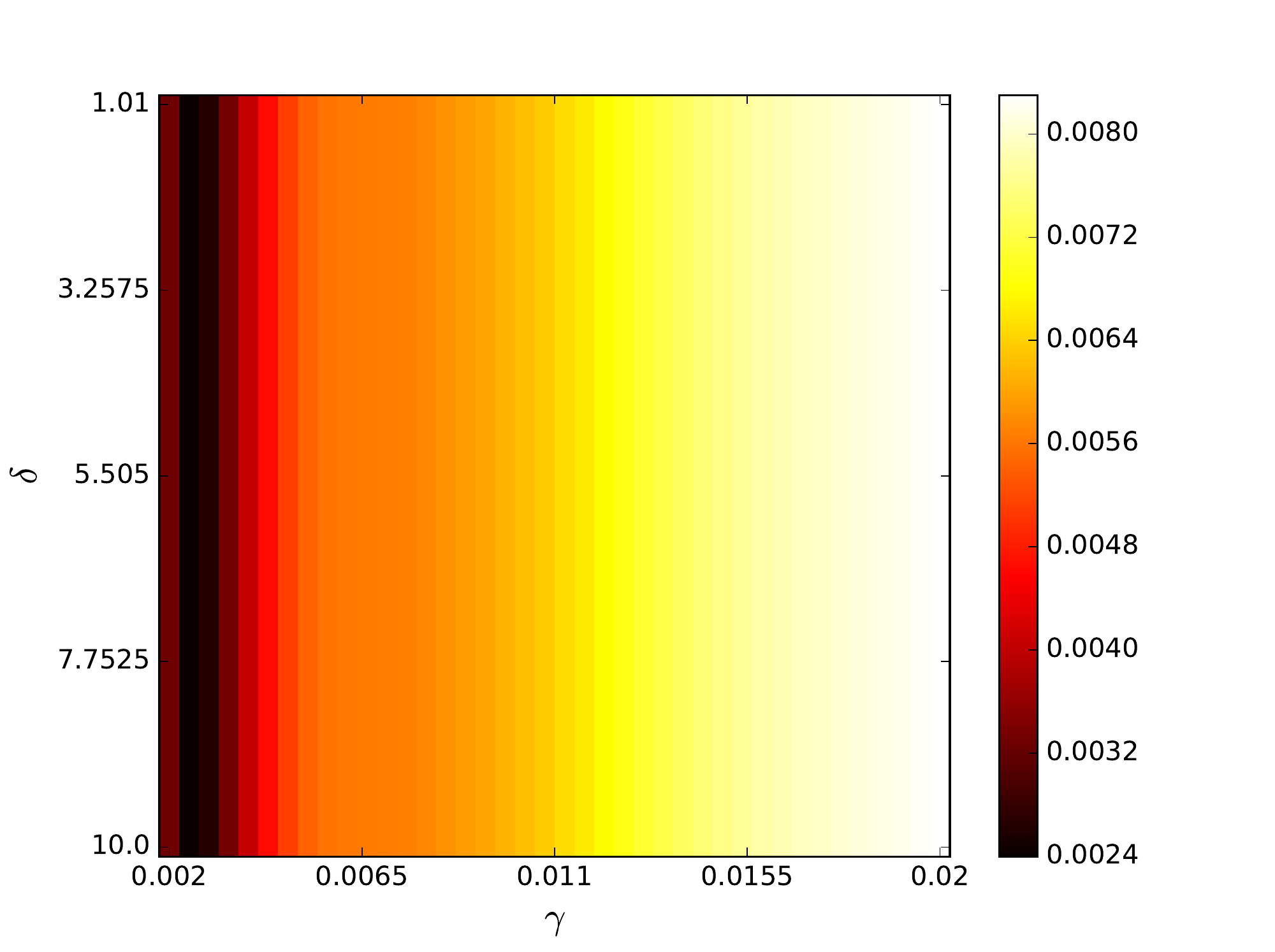}
%\caption{Caption of first subfigure}
\end{subfigure}
\hspace*{\fill}
\begin{subfigure}{0.32\textwidth}
\includegraphics[width=\linewidth]{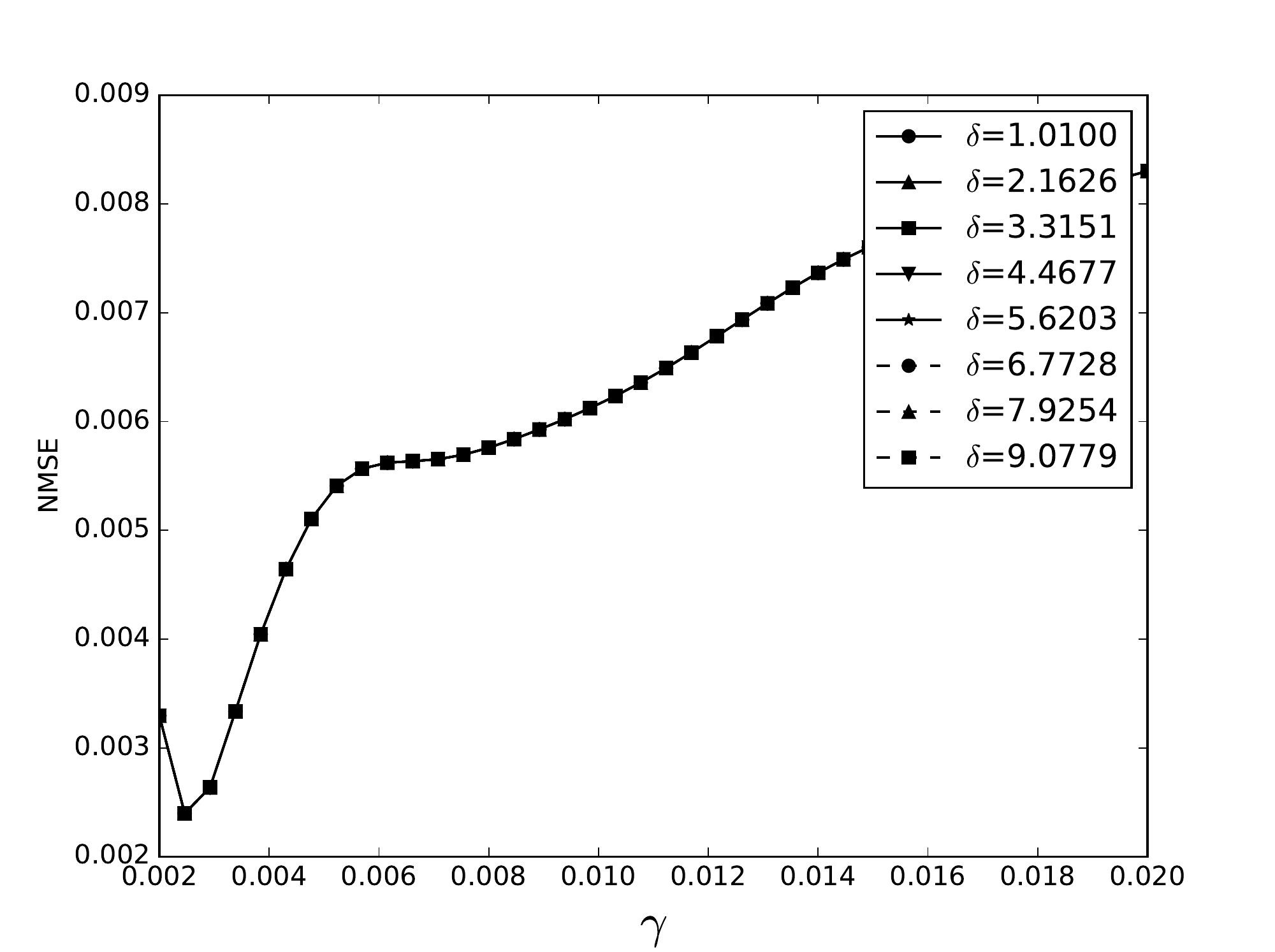}
%\caption{Caption of first subfigure}
\end{subfigure}
\hspace*{\fill}
\begin{subfigure}{0.32\textwidth}
\includegraphics[width=\linewidth]{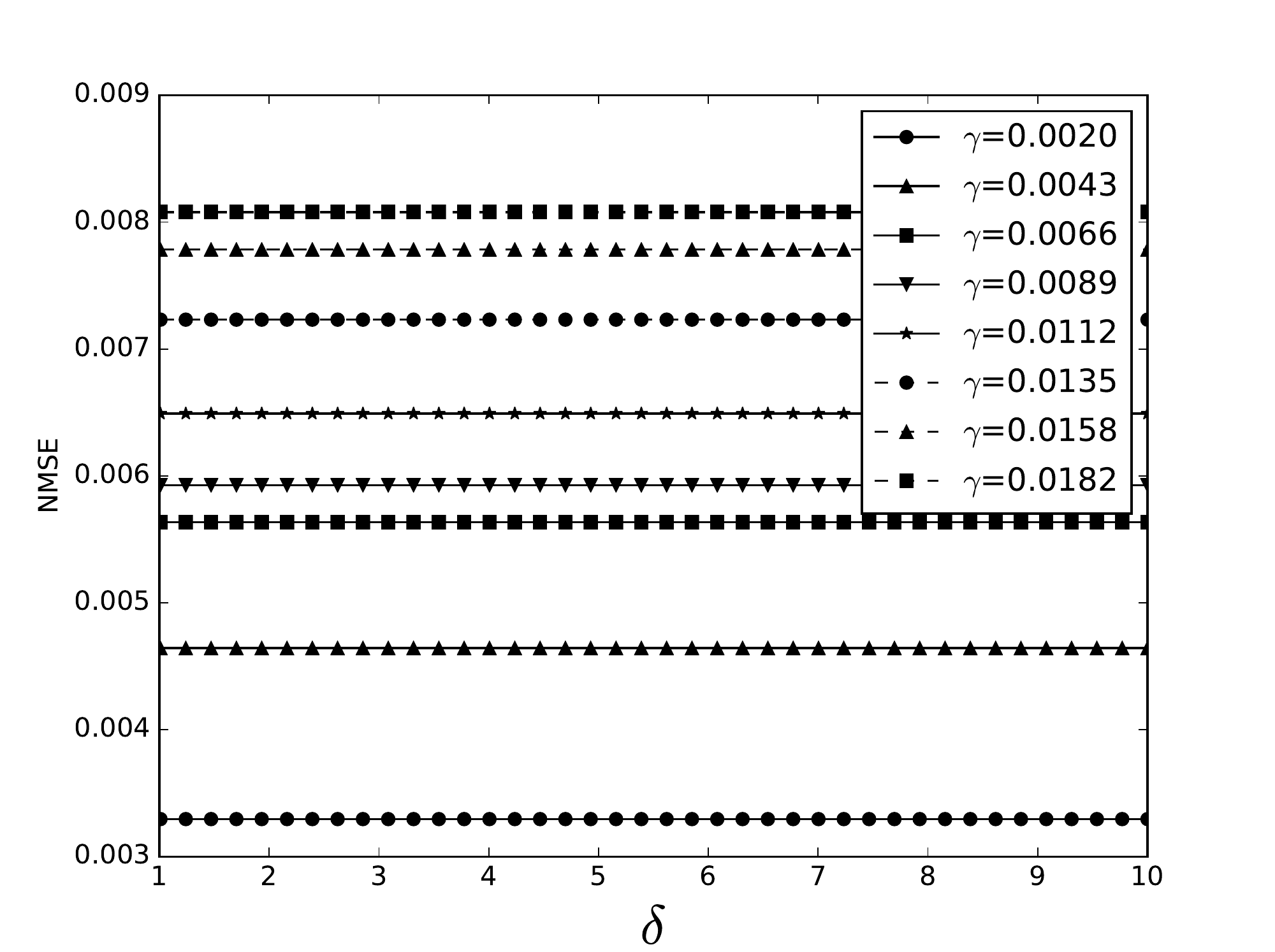}
%\caption{Caption of second subfigure}
\end{subfigure}
\caption{Santa Fe task with MG NDN and MSI, $\gamma$ and $\delta$.} \label{fig:santafe_gammadelta_MSI}
\end{figure}

\section{Heat maps NARMA 10 task and MG NDN}\label{app:narmahm}

We plot heat maps and one-dimensional slices of the error surface of the Santa Fe task. Parameter regions are near the optimal values unless stated otherwise.

\begin{figure}
\begin{subfigure}{0.32\textwidth}
\includegraphics[width=\linewidth]{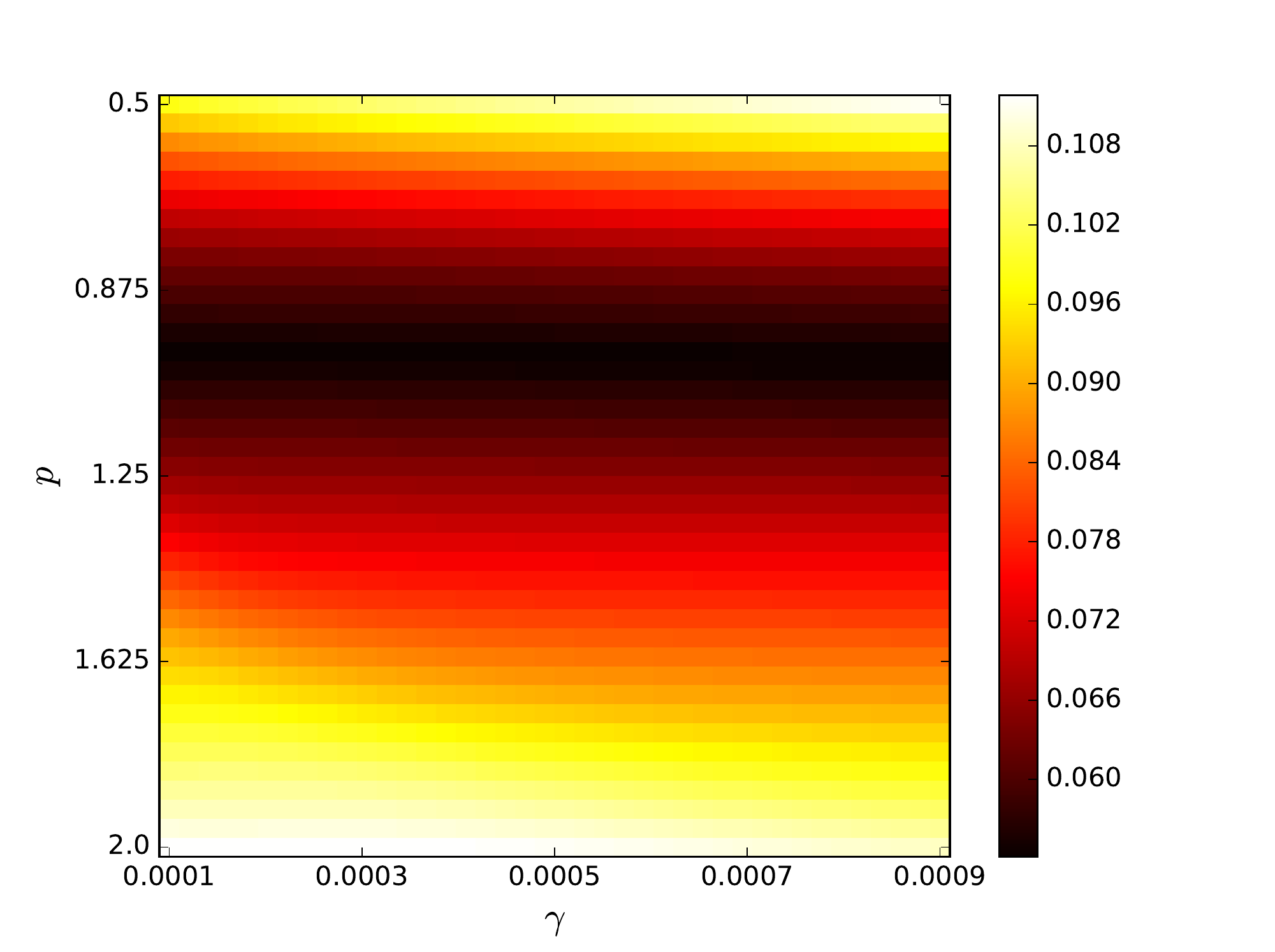}
%\caption{Caption of first subfigure}
\end{subfigure}
\hspace*{\fill}
\begin{subfigure}{0.32\textwidth}
\includegraphics[width=\linewidth]{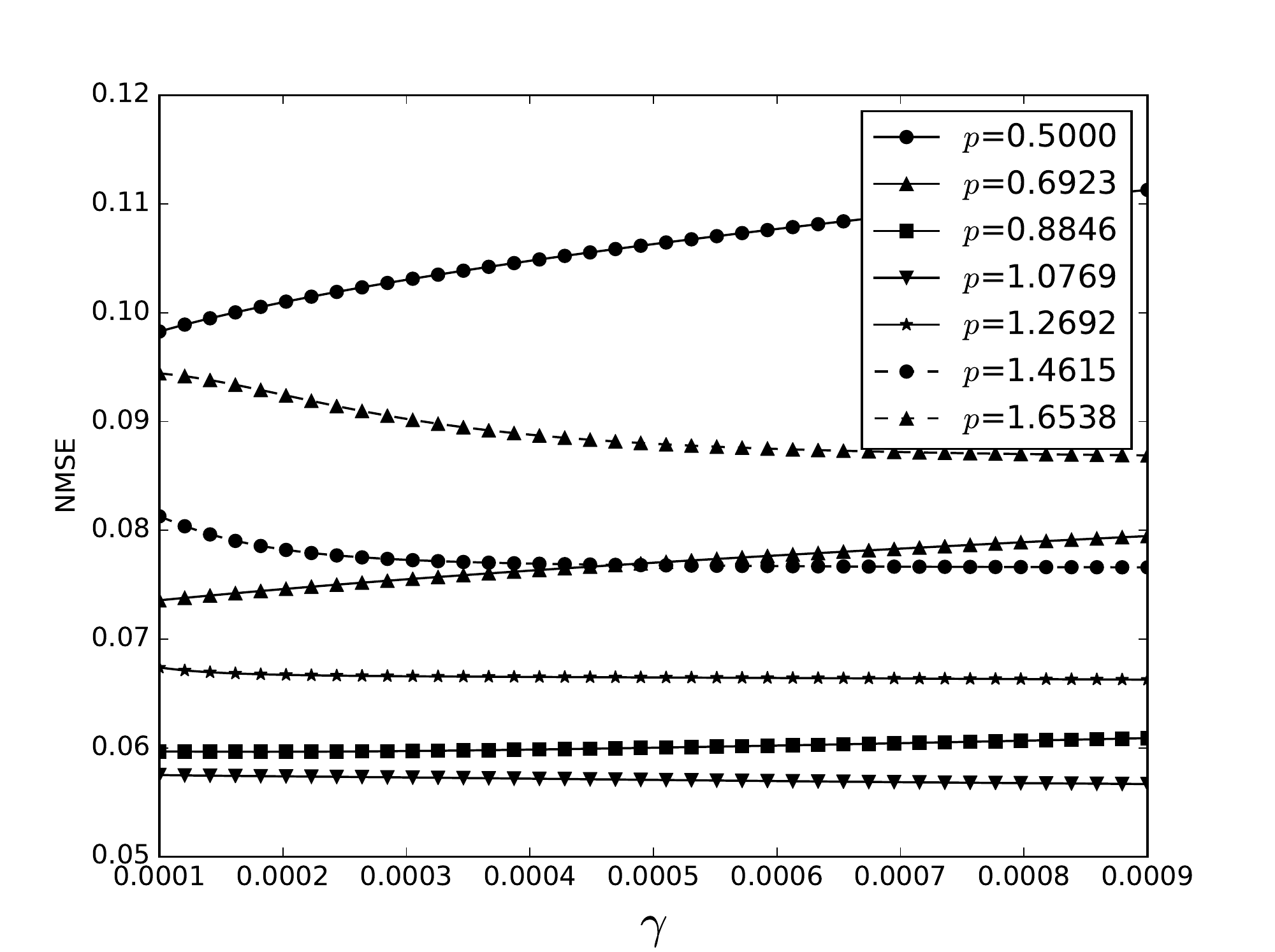}
%\caption{Caption of first subfigure}
\end{subfigure}
\hspace*{\fill}
\begin{subfigure}{0.32\textwidth}
\includegraphics[width=\linewidth]{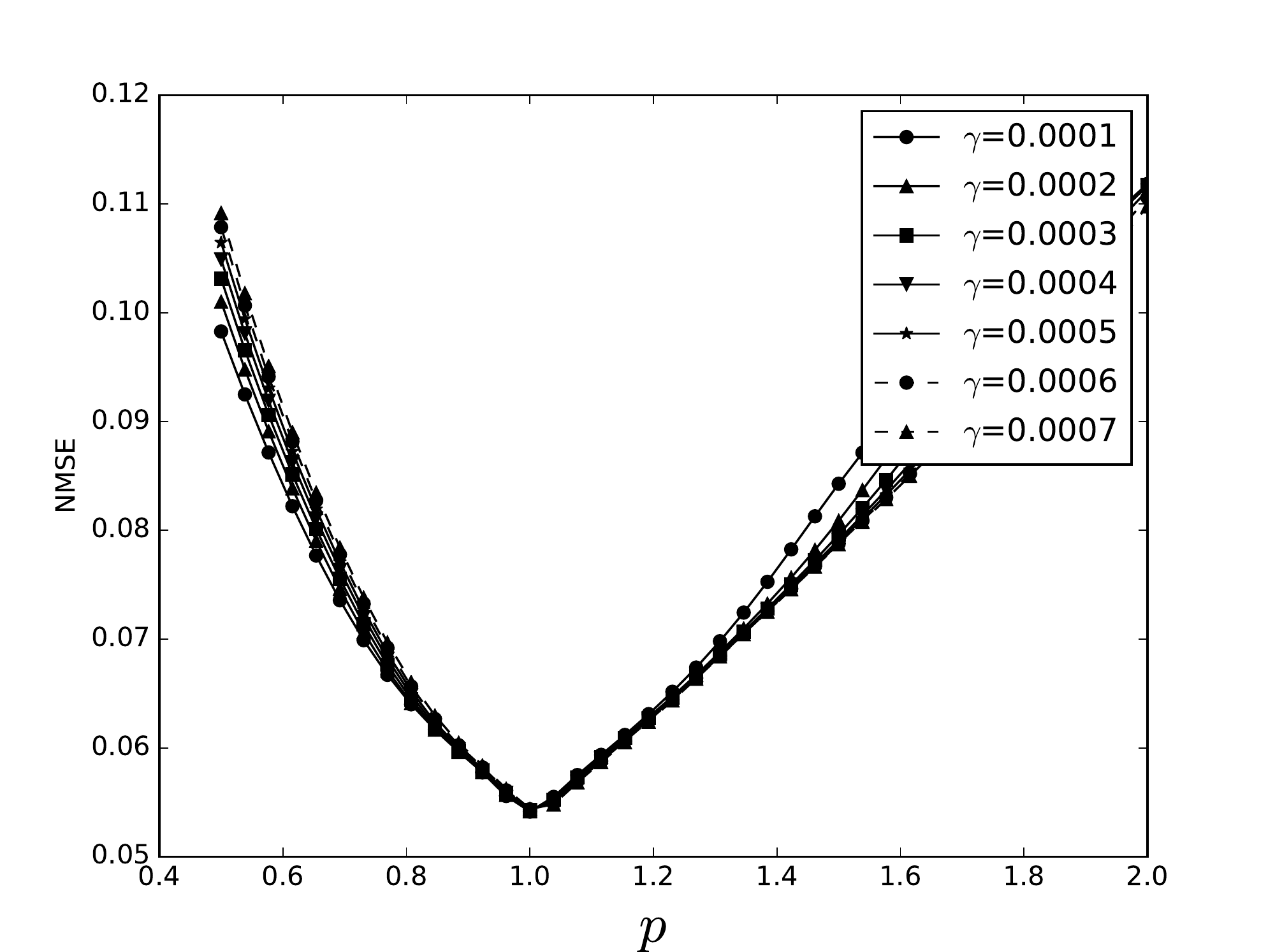}
%\caption{Caption of second subfigure}
\end{subfigure}
\caption{NARMA 10 task with MG NDN and OSI, $\gamma$ and $p$.} \label{fig:narma_gammap}
\end{figure}

\begin{figure}
\begin{subfigure}{0.32\textwidth}
\includegraphics[width=\linewidth]{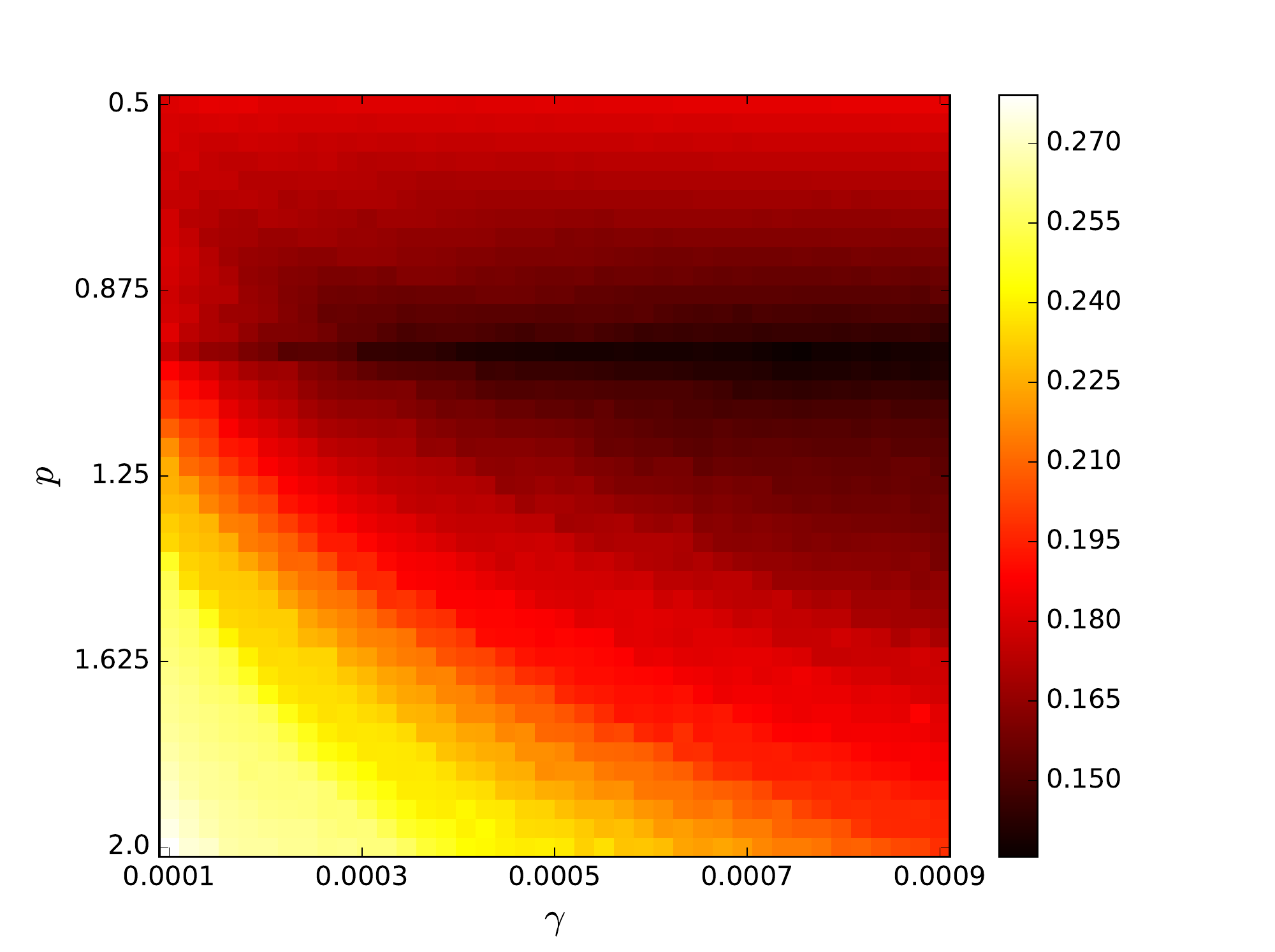}
%\caption{Caption of first subfigure}
\end{subfigure}
\hspace*{\fill}
\begin{subfigure}{0.32\textwidth}
\includegraphics[width=\linewidth]{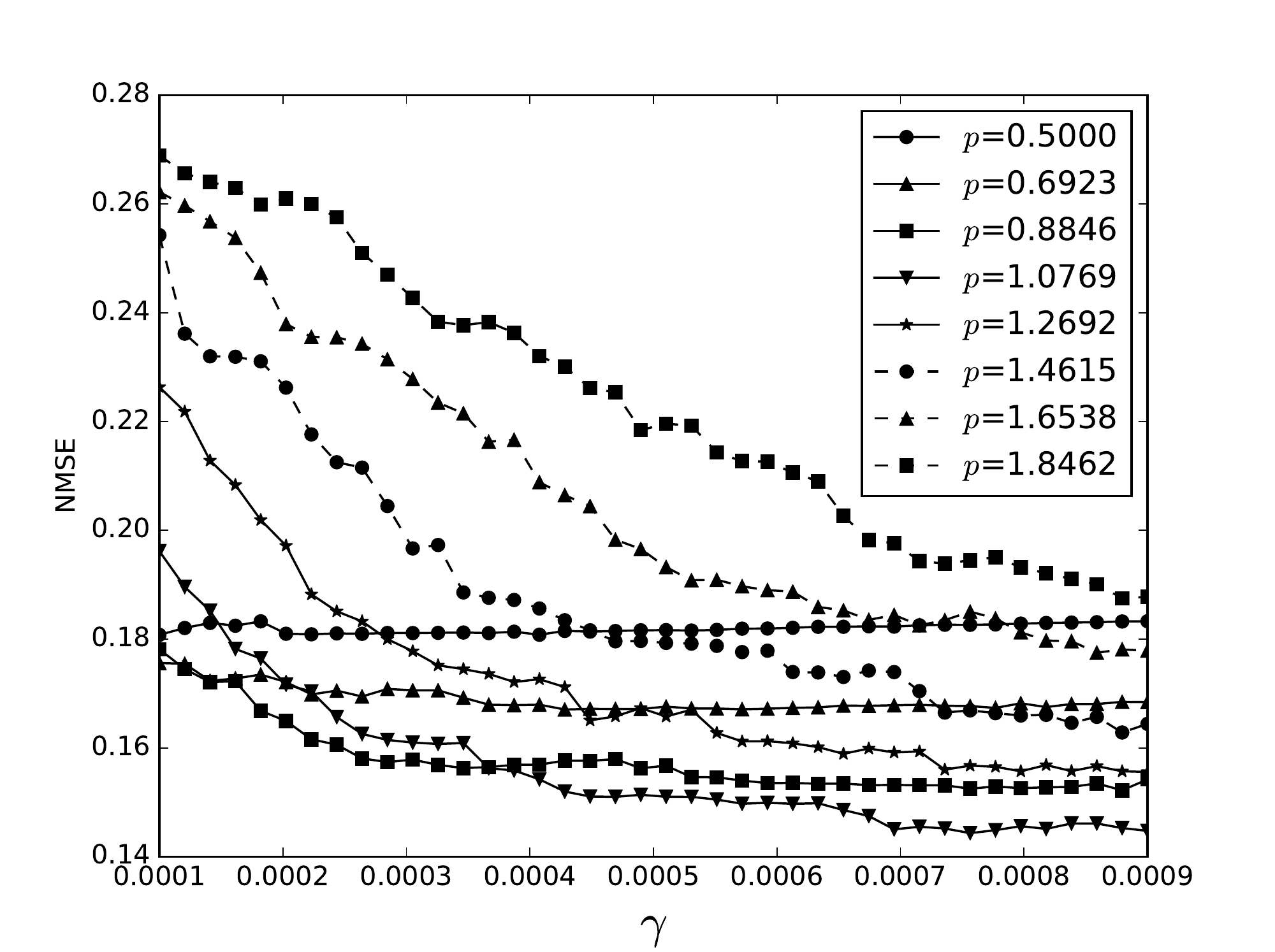}
%\caption{Caption of first subfigure}
\end{subfigure}
\hspace*{\fill}
\begin{subfigure}{0.32\textwidth}
\includegraphics[width=\linewidth]{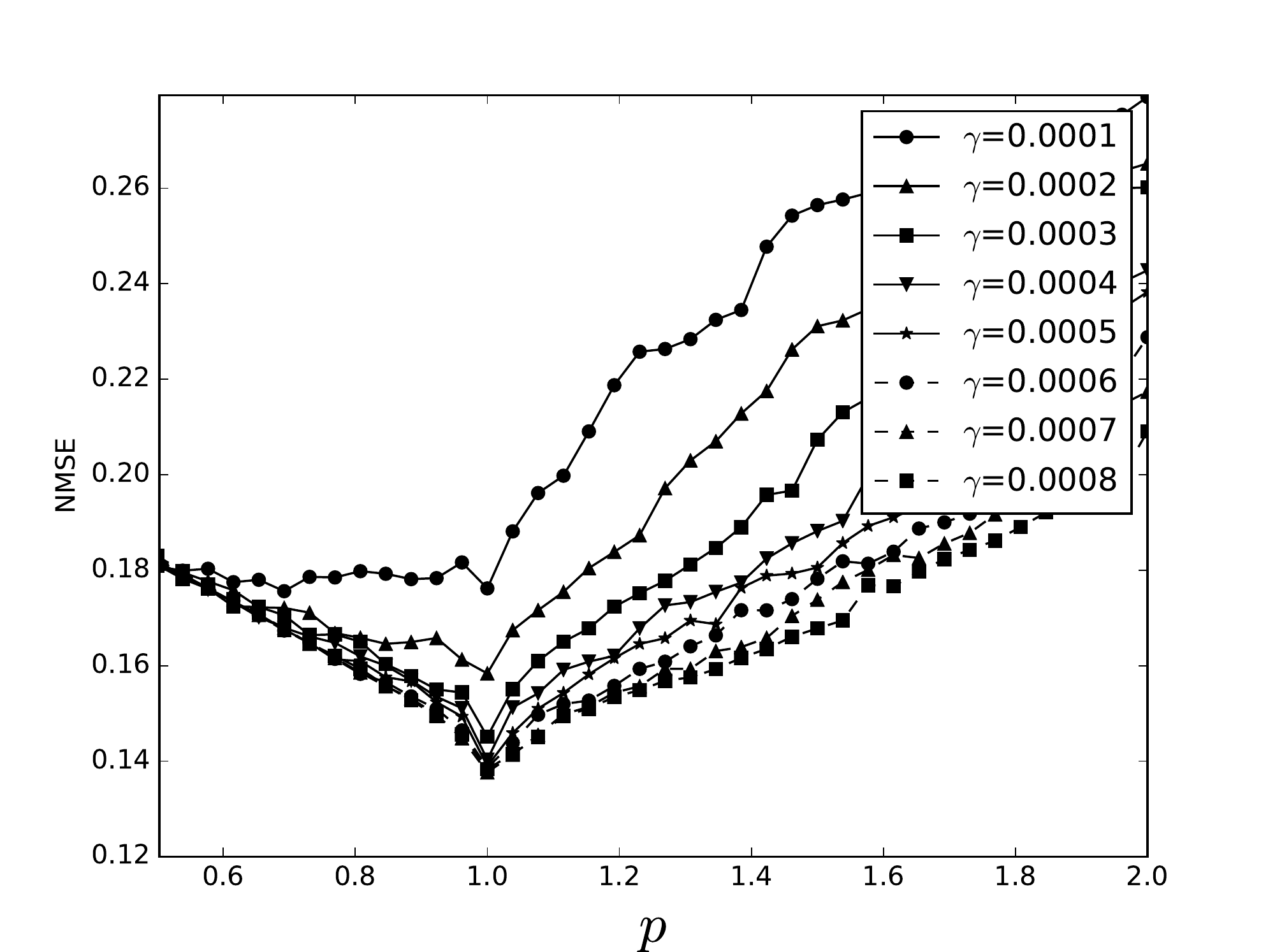}
%\caption{Caption of second subfigure}
\end{subfigure}
\caption{NARMA 10 with MG NDN and MSI, $\gamma$ and $p$.} \label{fig:narma_gammap_MSI}
\end{figure}

\begin{figure}
\begin{subfigure}{0.32\textwidth}
\includegraphics[width=\linewidth]{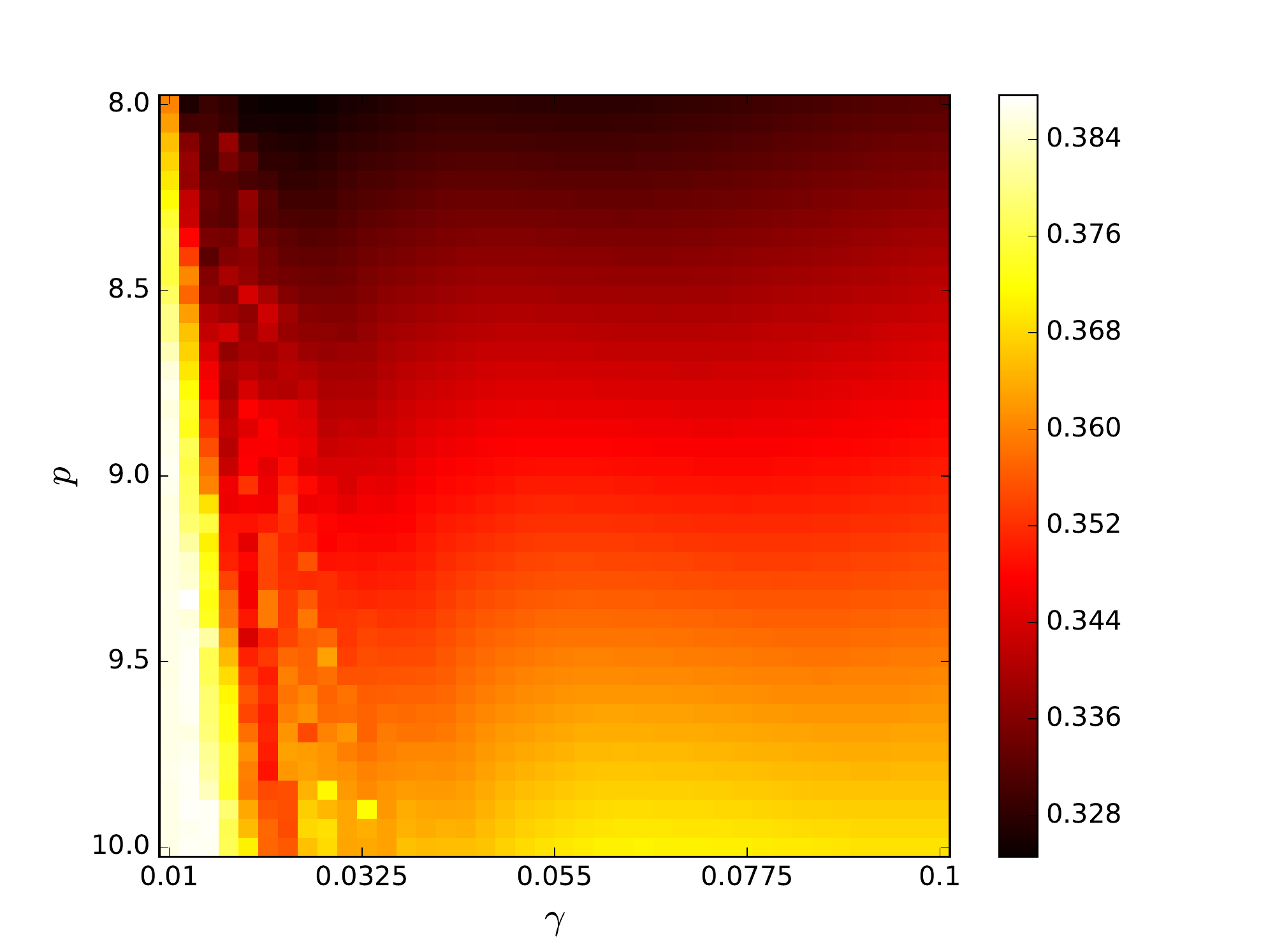}
%\caption{Caption of first subfigure}
\end{subfigure}
\hspace*{\fill}
\begin{subfigure}{0.32\textwidth}
\includegraphics[width=\linewidth]{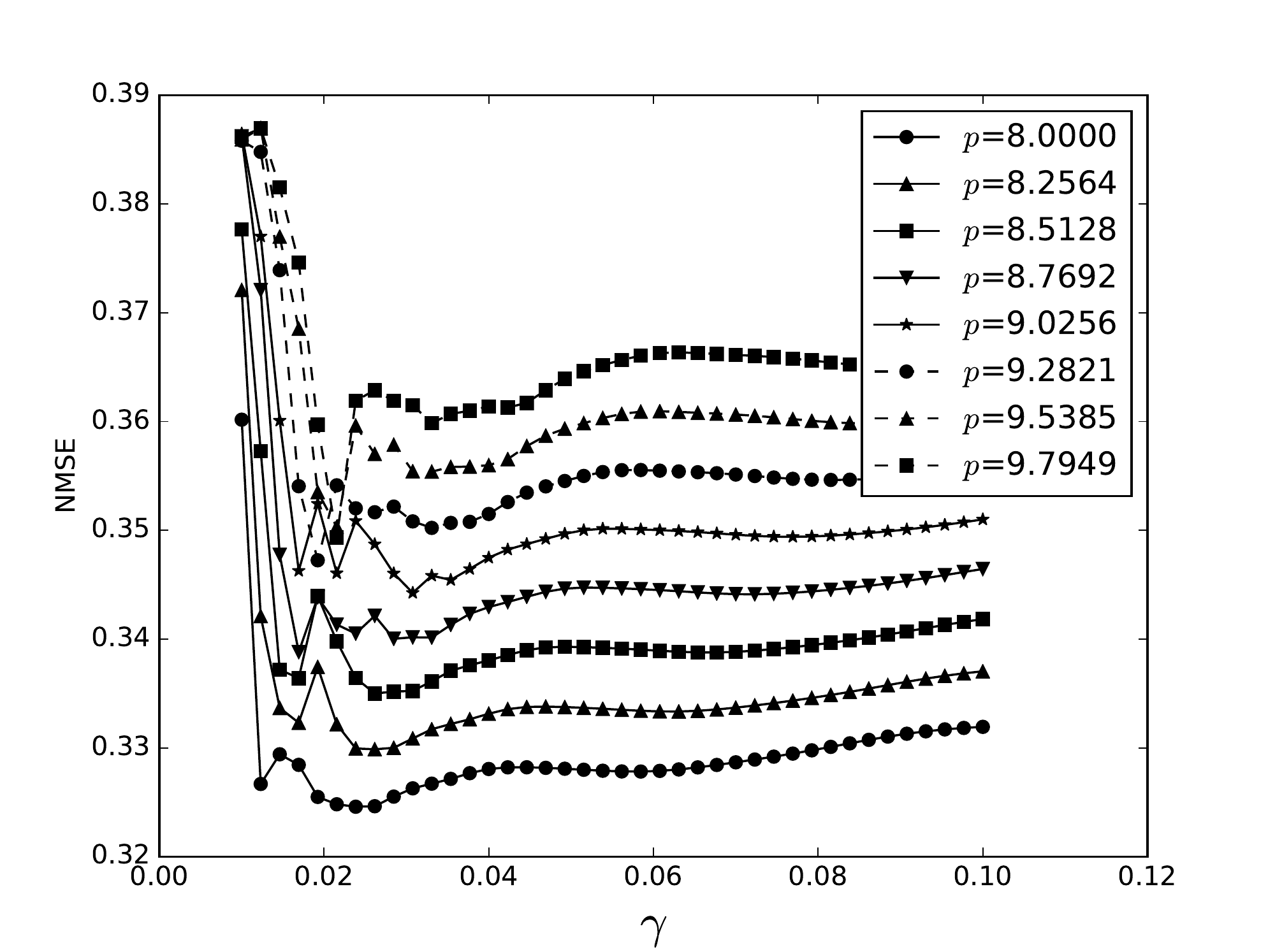}
%\caption{Caption of first subfigure}
\end{subfigure}
\hspace*{\fill}
\begin{subfigure}{0.32\textwidth}
\includegraphics[width=\linewidth]{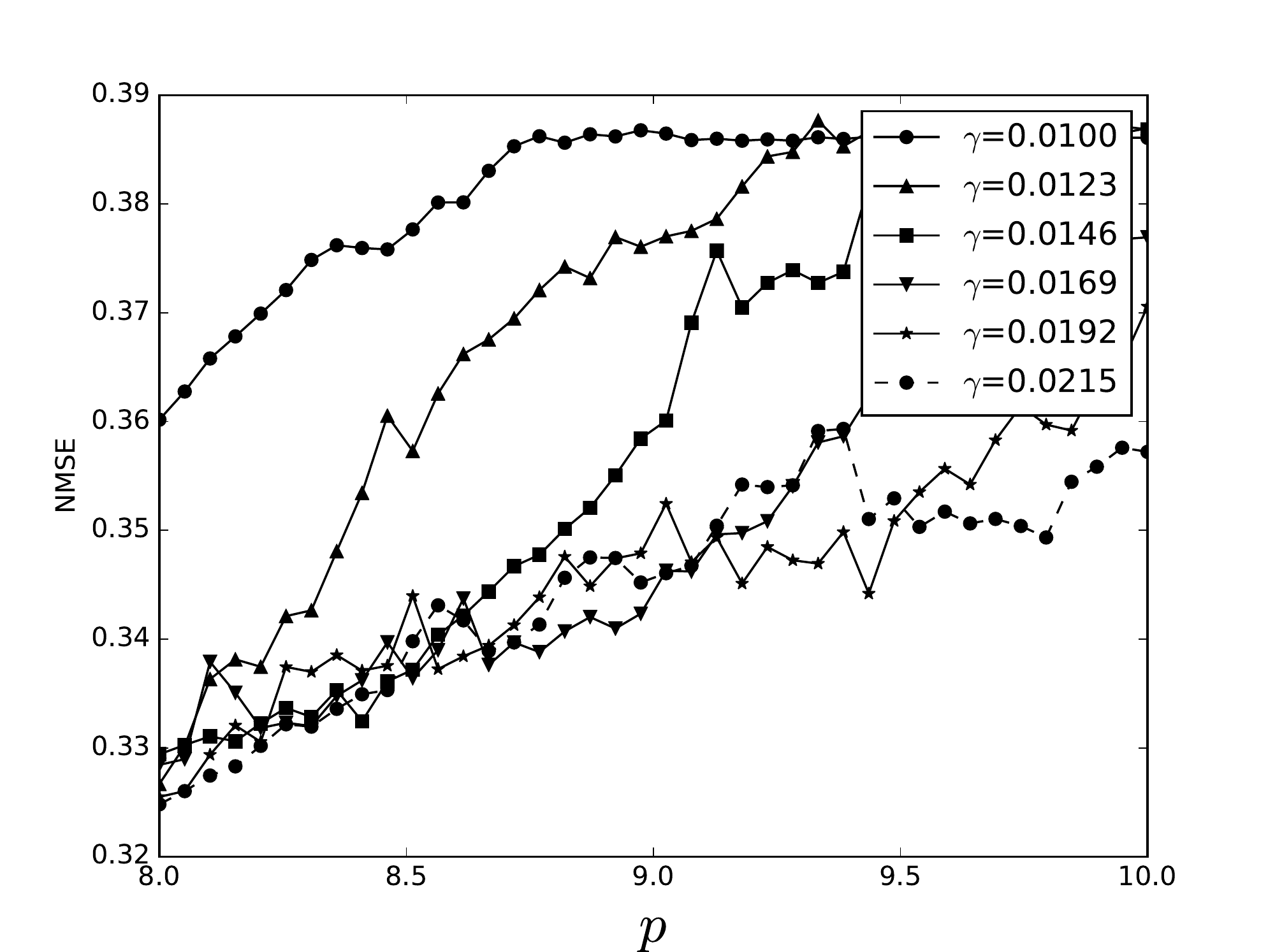}
%\caption{Caption of second subfigure}
\end{subfigure}
\caption{NARMA 10 task with MG NDN and OSI, $\gamma$ and $p$ for a parameter region away from the optimum.} \label{fig:narma_gammap_bad}
\end{figure}

\begin{figure}
\begin{subfigure}{0.32\textwidth}
\includegraphics[width=\linewidth]{figures/heatmap_narma_eta_theta_transform.pdf}
%\caption{Caption of first subfigure}
\end{subfigure}
\hspace*{\fill}
\begin{subfigure}{0.32\textwidth}
\includegraphics[width=\linewidth]{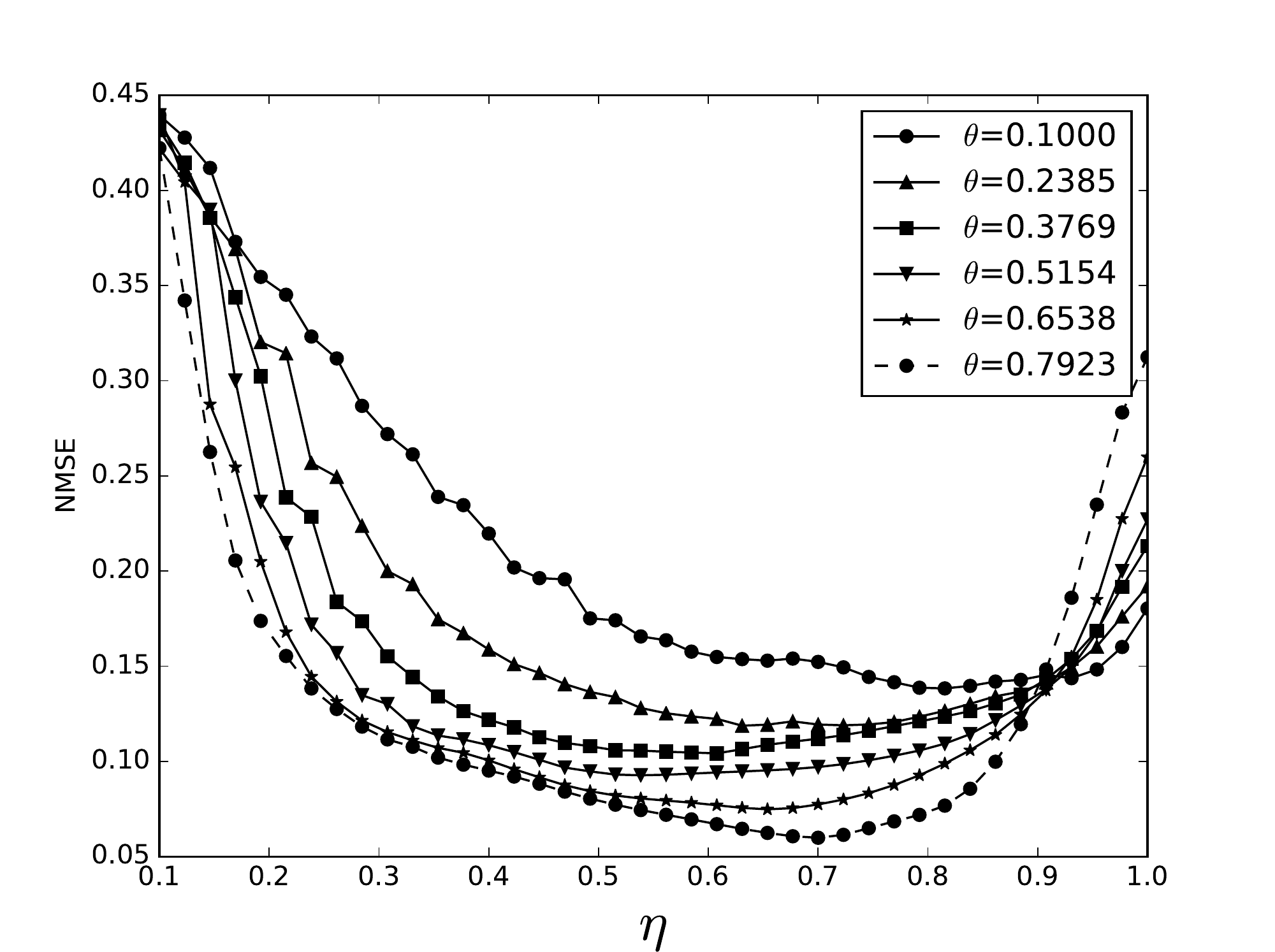}
%\caption{Caption of first subfigure}
\end{subfigure}
\hspace*{\fill}
\begin{subfigure}{0.32\textwidth}
\includegraphics[width=\linewidth]{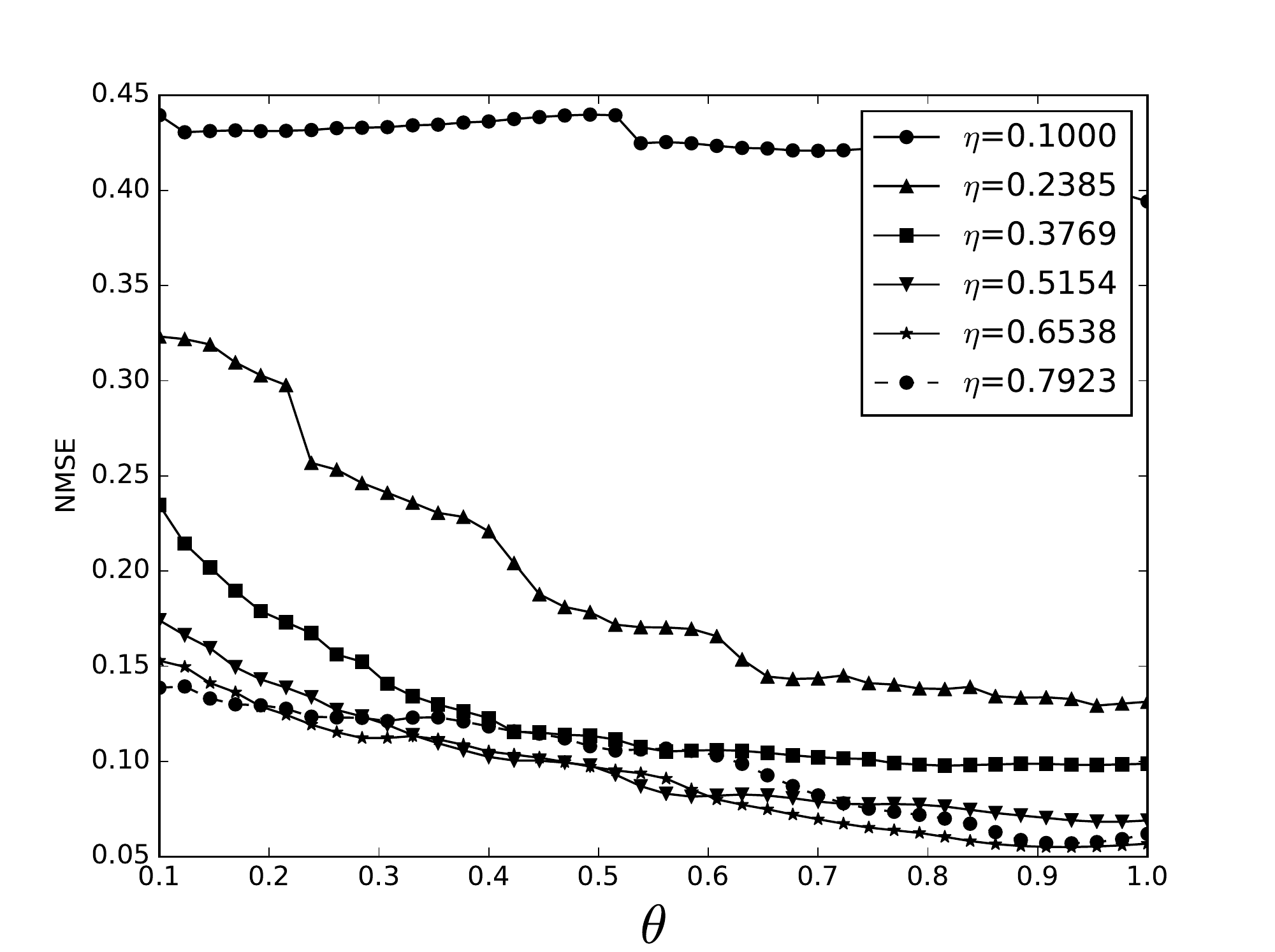}
%\caption{Caption of second subfigure}
\end{subfigure}
\caption{NARMA 10 task with MG NDN and OSI, $\eta$ and $\theta$.} \label{fig:narma_etatheta}
\end{figure}

\begin{figure}
\begin{subfigure}{0.32\textwidth}
\includegraphics[width=\linewidth]{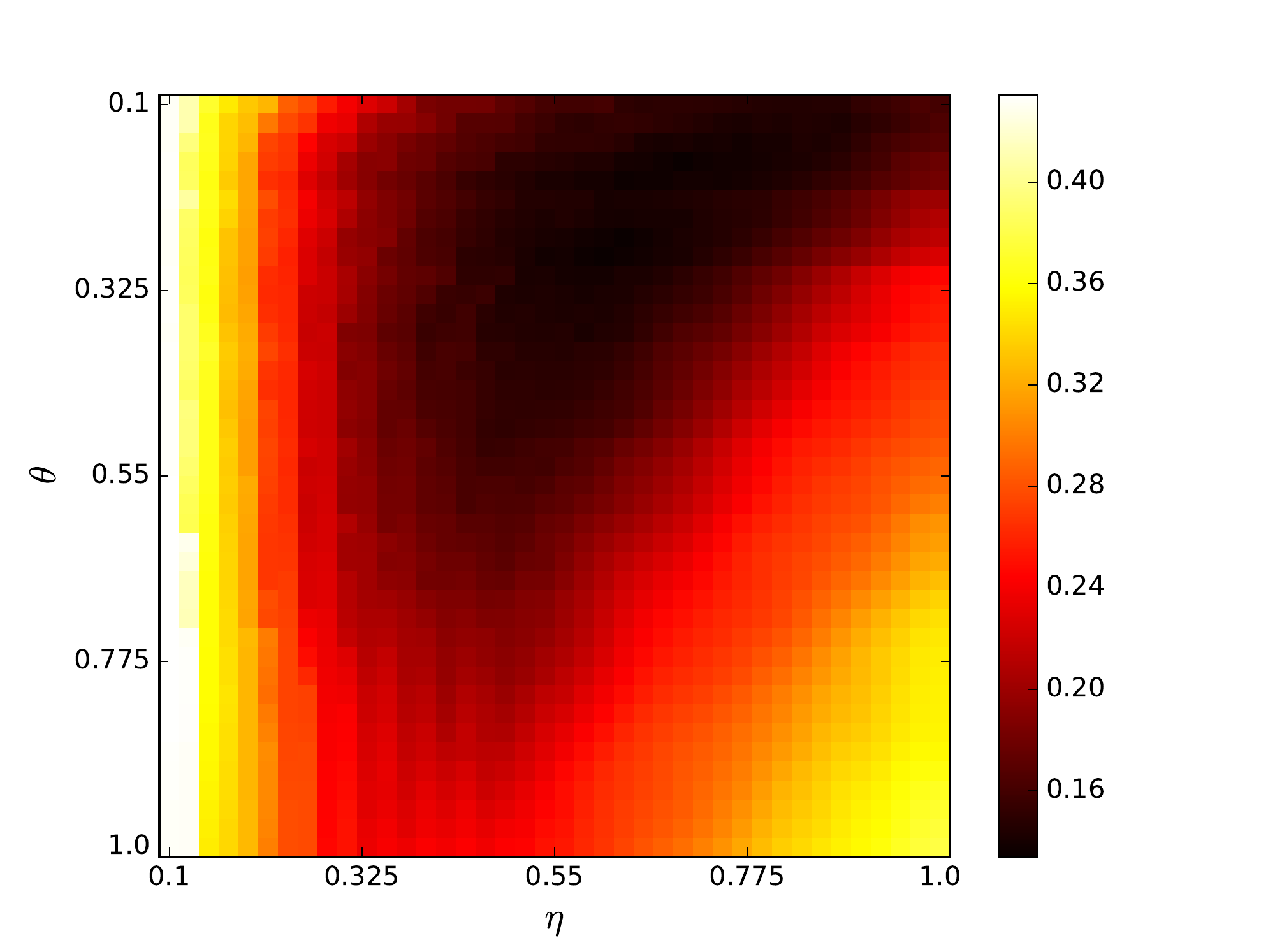}
%\caption{Caption of first subfigure}
\end{subfigure}
\hspace*{\fill}
\begin{subfigure}{0.32\textwidth}
\includegraphics[width=\linewidth]{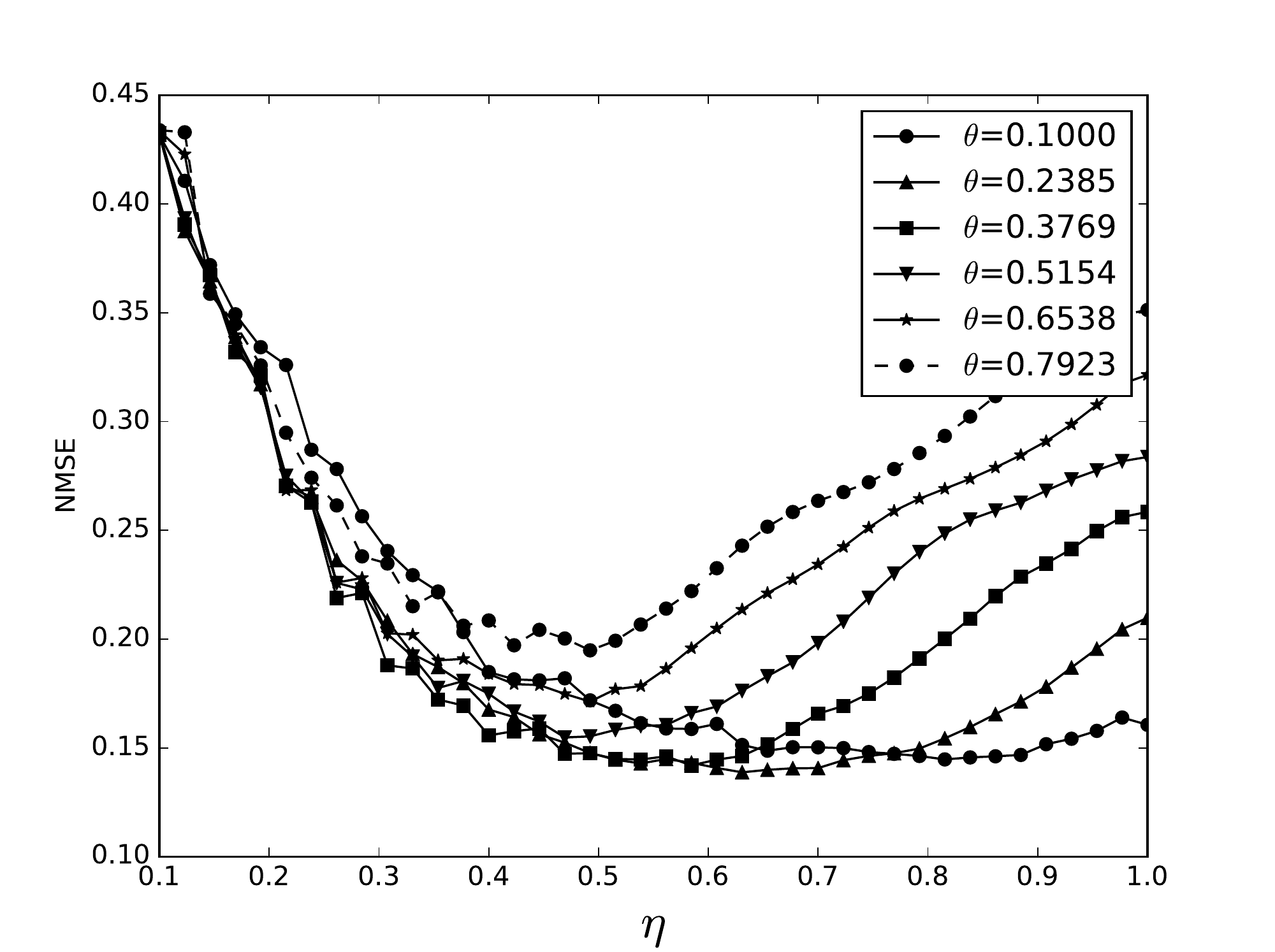}
%\caption{Caption of first subfigure}
\end{subfigure}
\hspace*{\fill}
\begin{subfigure}{0.32\textwidth}
\includegraphics[width=\linewidth]{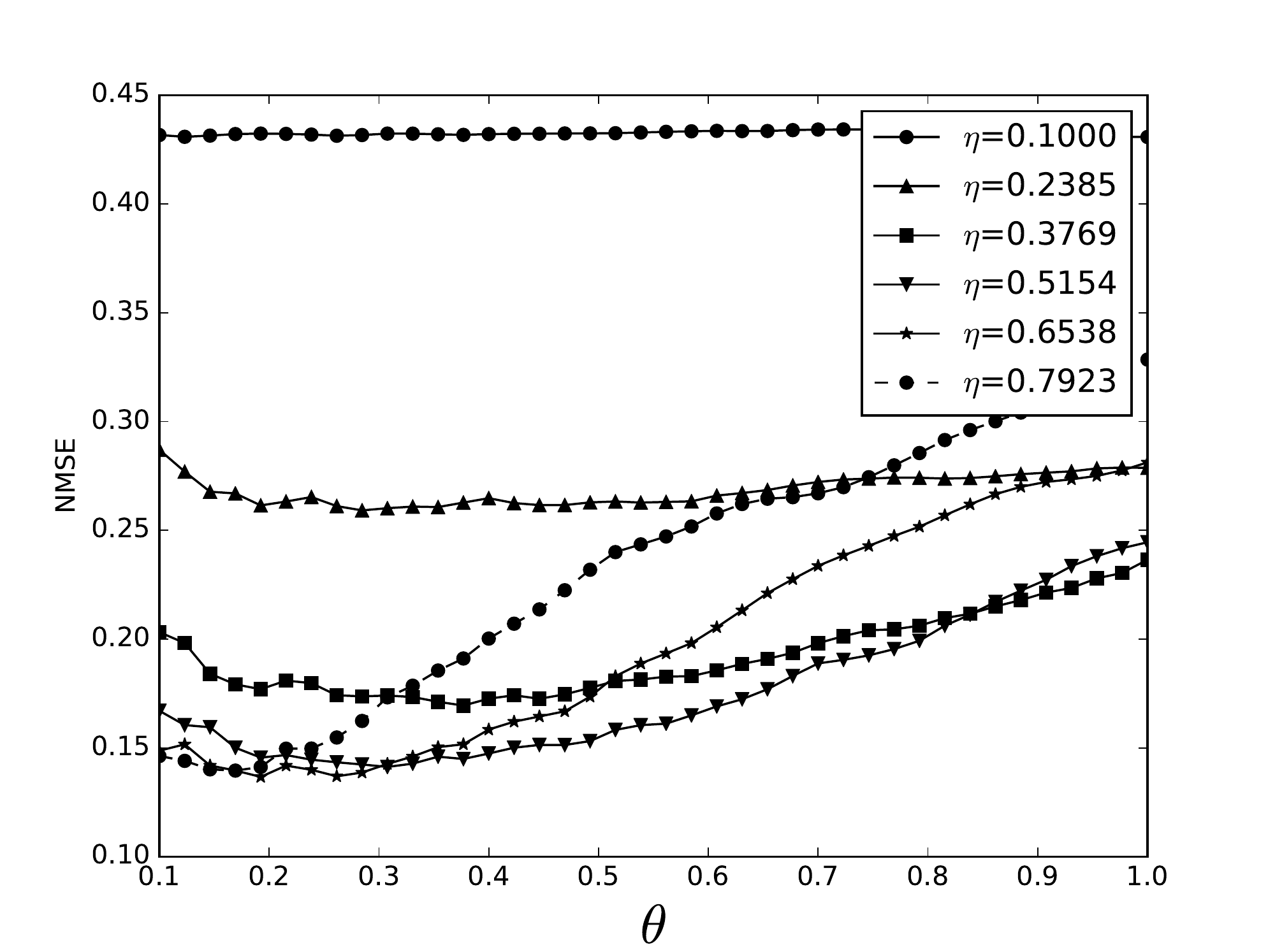}
%\caption{Caption of second subfigure}
\end{subfigure}
\caption{NARMA 10 task with MG NDN and MSI, $\eta$ and $\theta$.} \label{fig:narma_etatheta_MSI}
\end{figure}

\begin{figure}
\begin{subfigure}{0.32\textwidth}
\includegraphics[width=\linewidth]{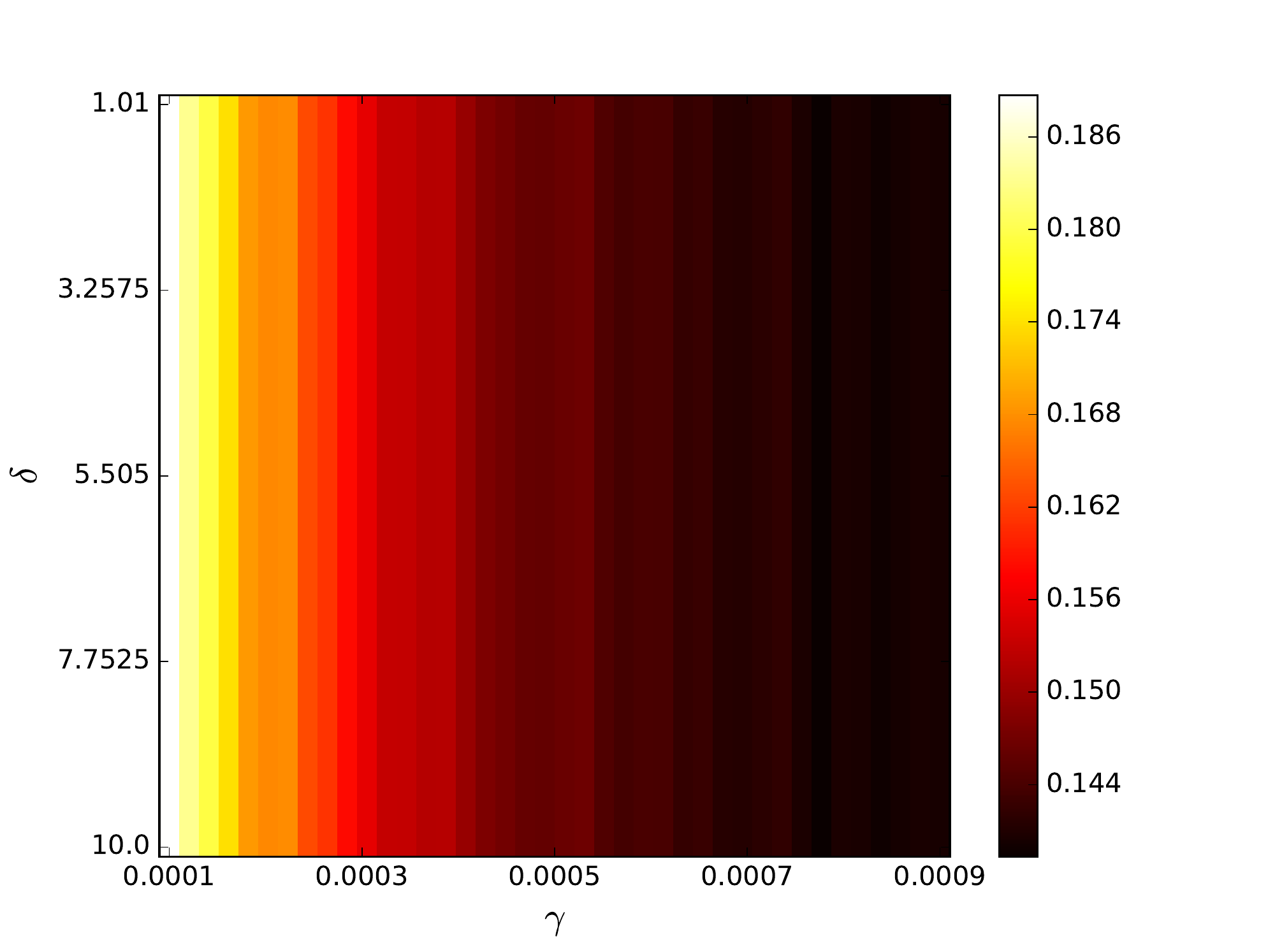}
%\caption{Caption of first subfigure}
\end{subfigure}
\hspace*{\fill}
\begin{subfigure}{0.32\textwidth}
\includegraphics[width=\linewidth]{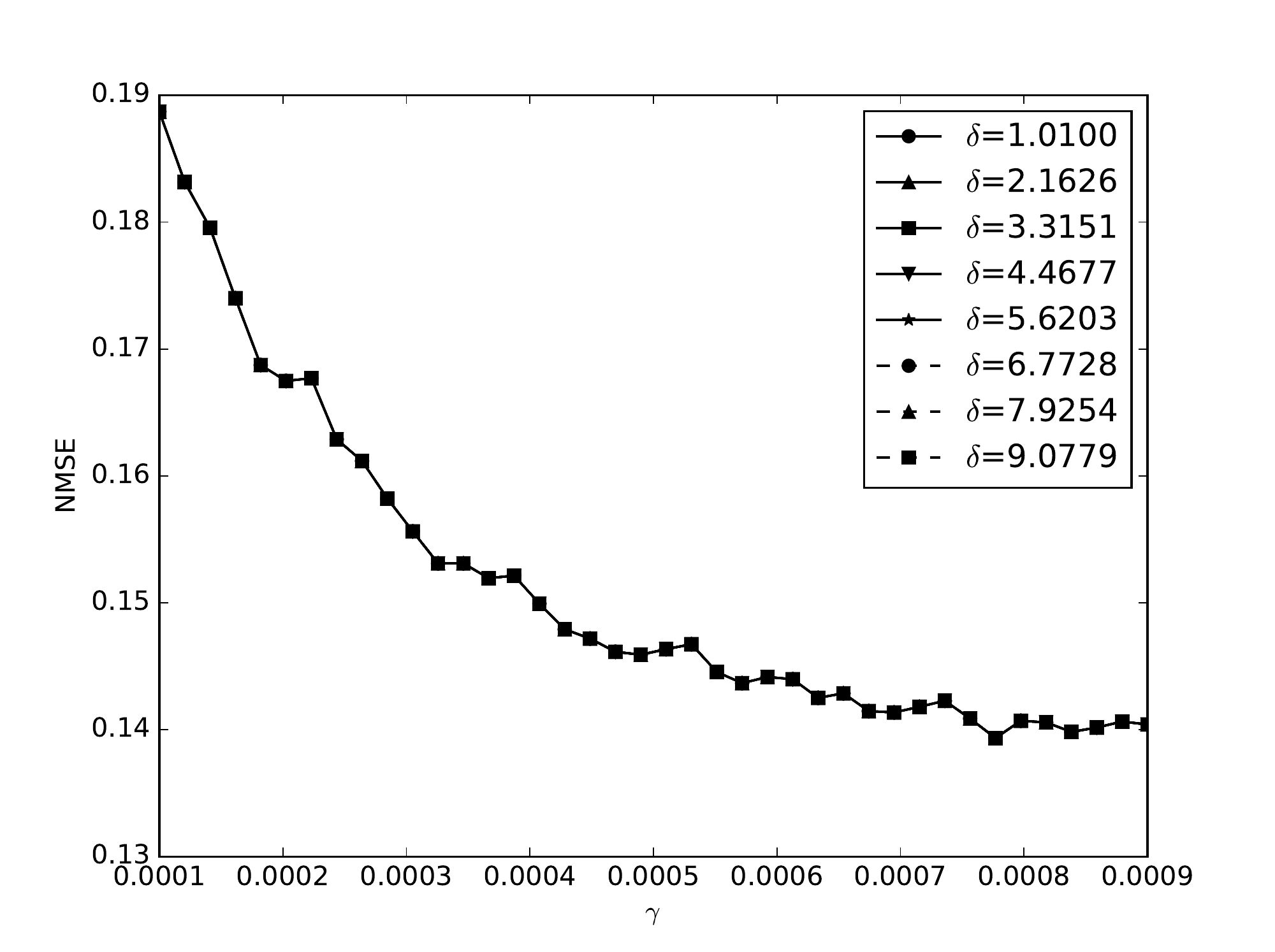}
%\caption{Caption of first subfigure}
\end{subfigure}
\hspace*{\fill}
\begin{subfigure}{0.32\textwidth}
\includegraphics[width=\linewidth]{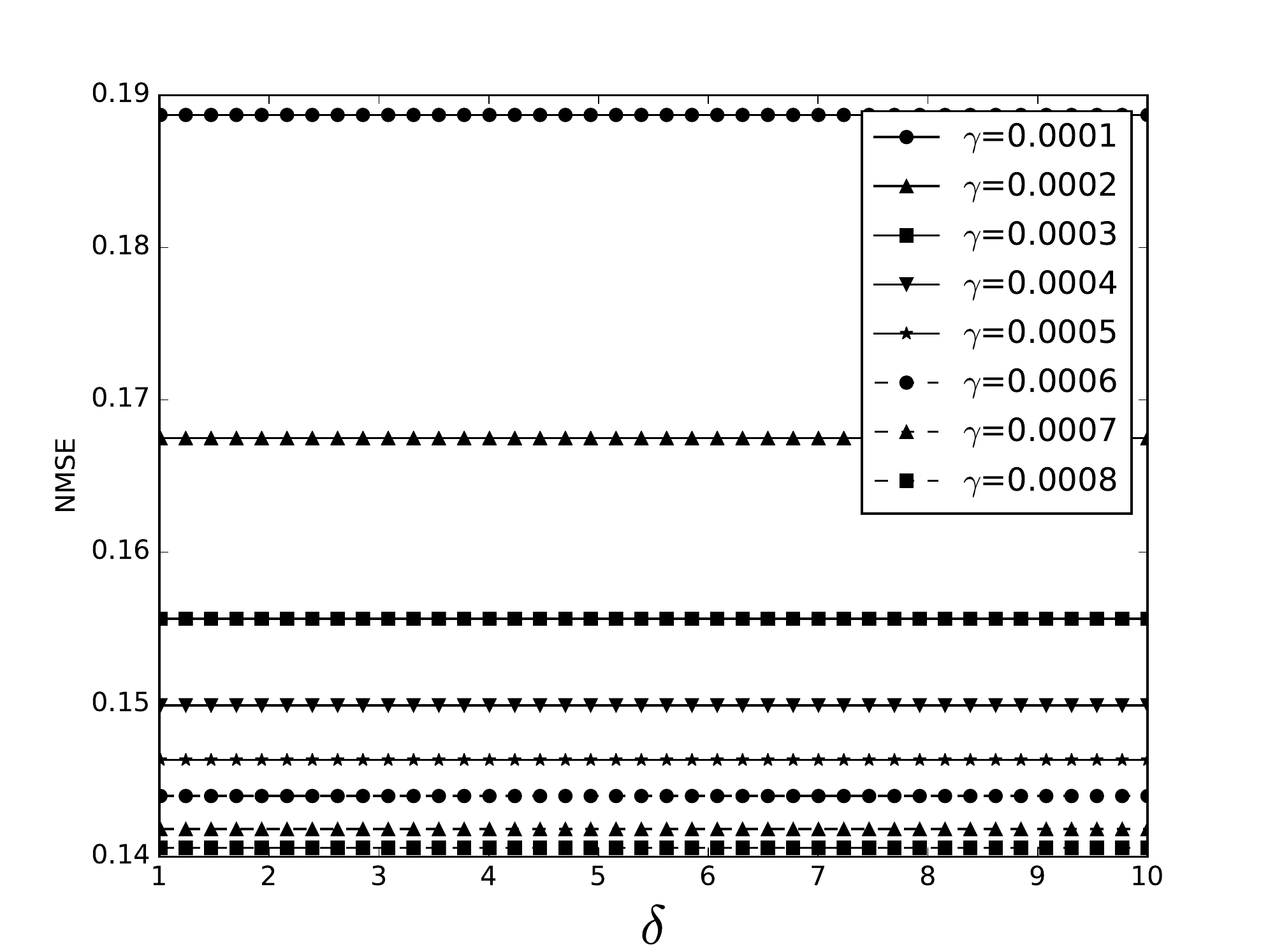}
%\caption{Caption of second subfigure}
\end{subfigure}
    \caption{NARMA 10 task with MG NDN and MSI, $\gamma$ and $\delta$.} \label{fig:narma_gammadelta_MSI}
\end{figure}

\begin{figure}
\begin{subfigure}{0.32\textwidth}
\includegraphics[width=\linewidth]{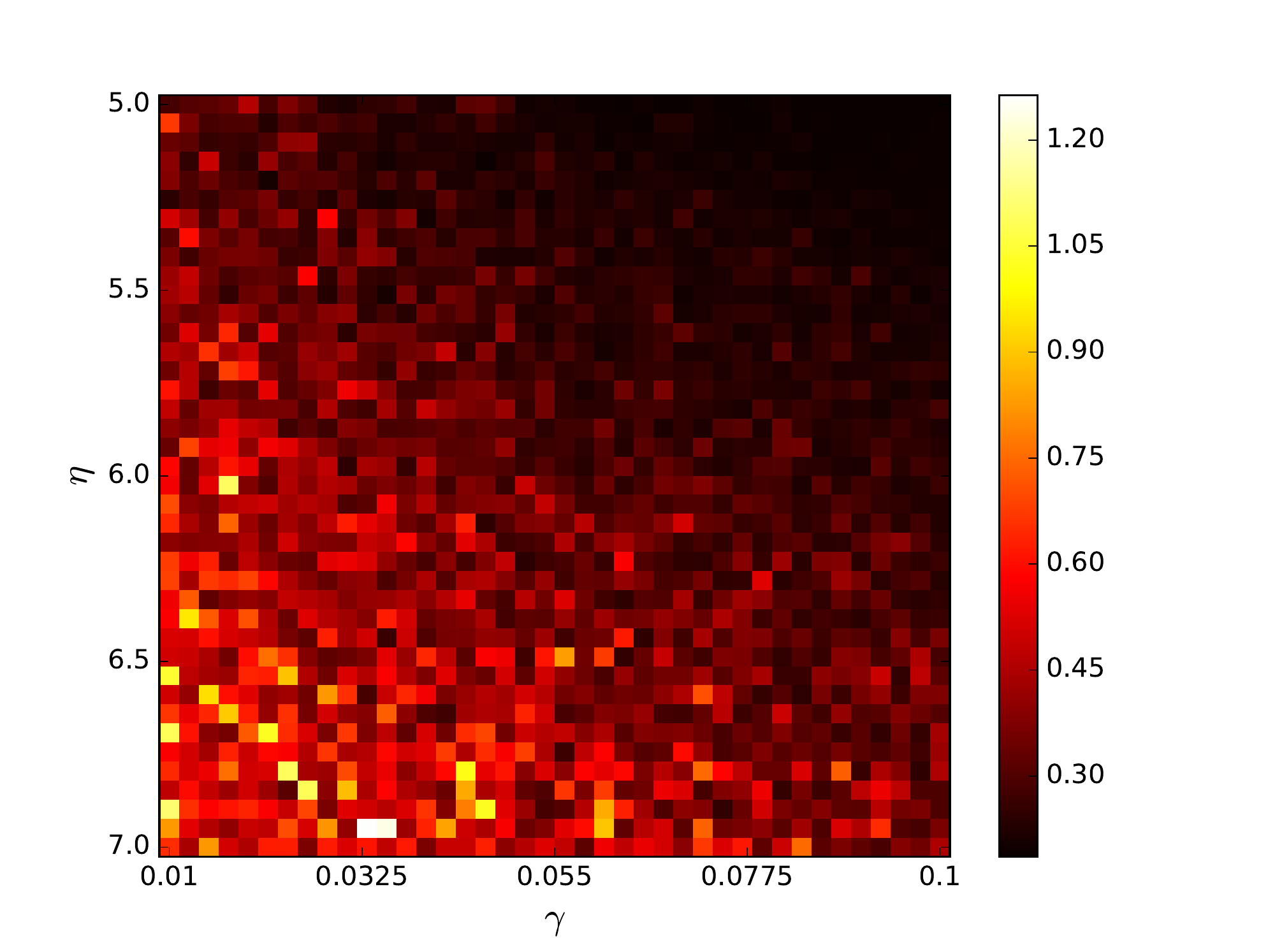}
%\caption{Caption of first subfigure}
\end{subfigure}
\hspace*{\fill}
\begin{subfigure}{0.32\textwidth}
\includegraphics[width=\linewidth]{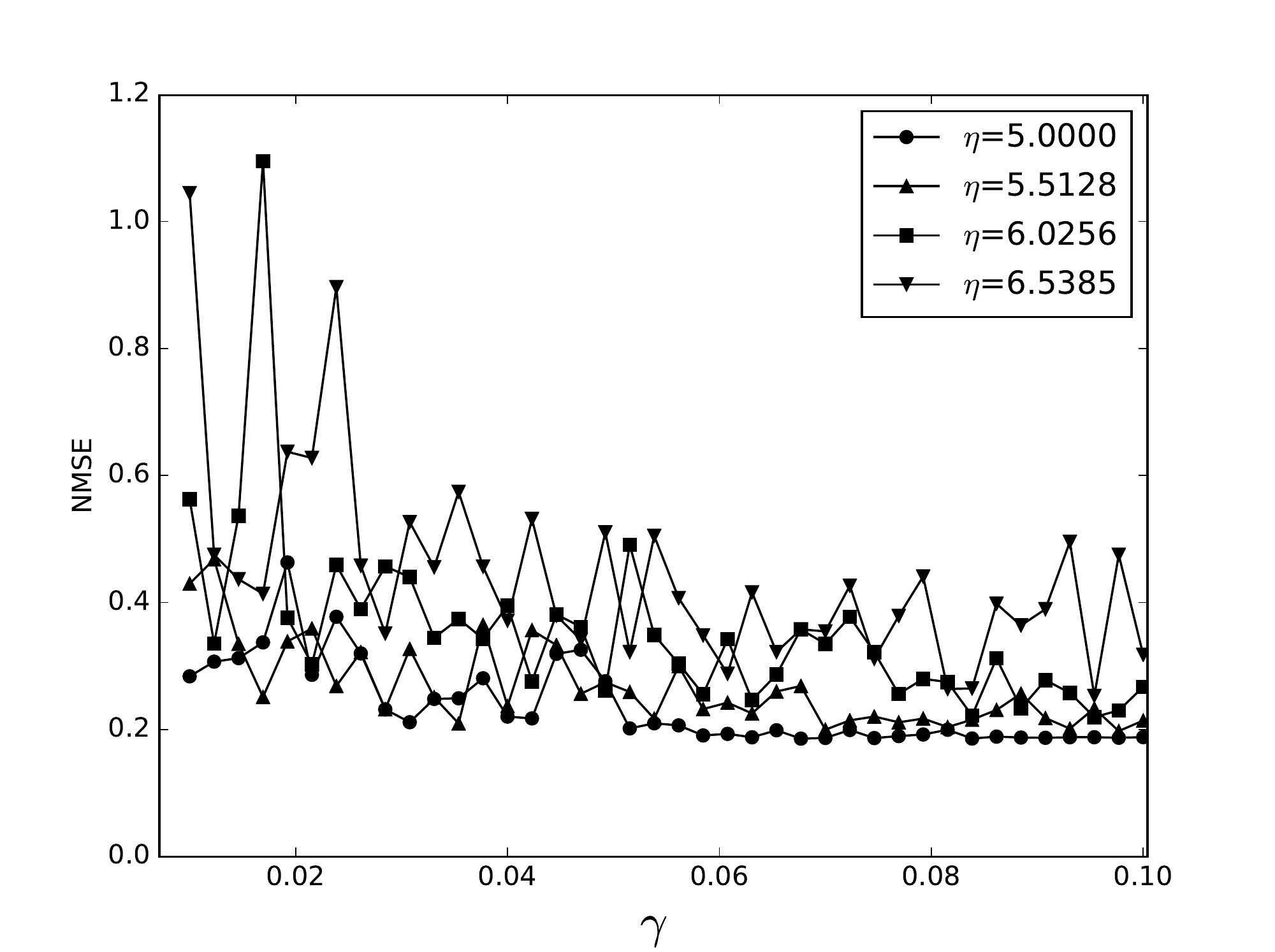}
%\caption{Caption of first subfigure}
\end{subfigure}
\hspace*{\fill}
\begin{subfigure}{0.32\textwidth}
\includegraphics[width=\linewidth]{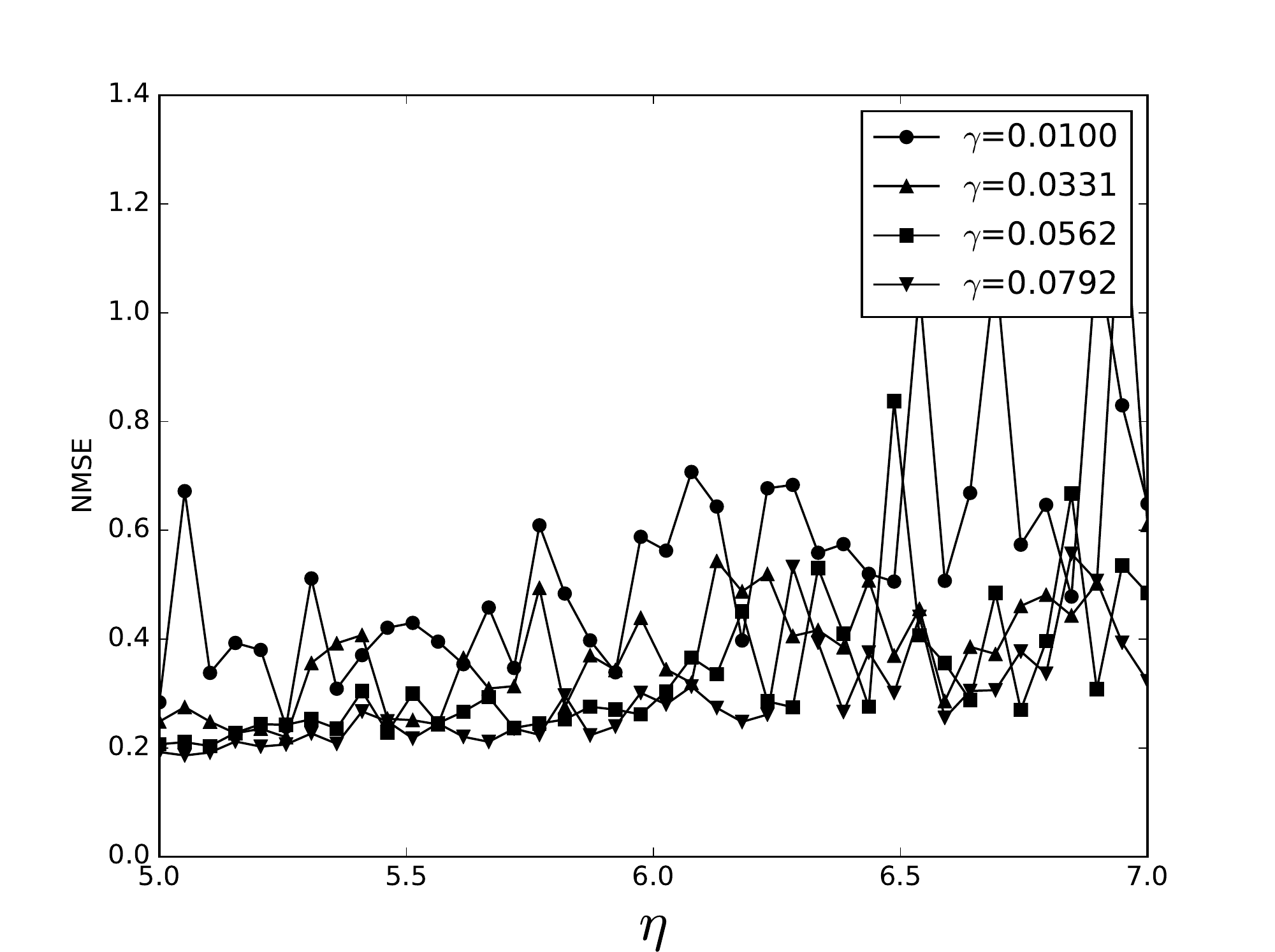}
%\caption{Caption of second subfigure}
\end{subfigure}
    \caption{NARMA 10 task with MG NDN and OSI, $\gamma$ and $\eta$ for a parameter region away from the optimum.} \label{fig:narma_gammaeta_bad}
\end{figure}

\section{Spearmint demos}\label{app:speardemo}

We plot heat maps and Spearmint runs for the nonlinear channel equalization task.

\begin{figure}
\begin{subfigure}{0.55\textwidth}
\includegraphics[width=\linewidth]{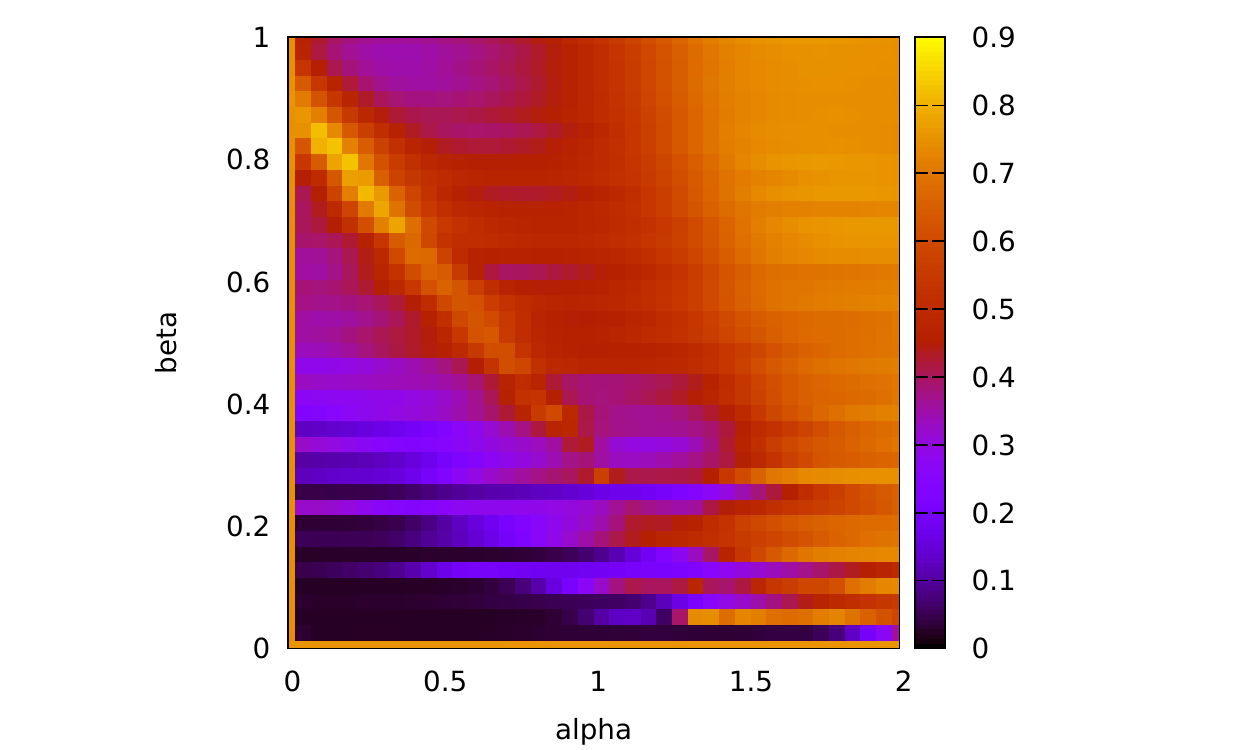}
%\caption{Caption of first subfigure}
\end{subfigure}
%\hspace*{\fill}
\begin{subfigure}{0.4\textwidth}
\includegraphics[width=\linewidth]{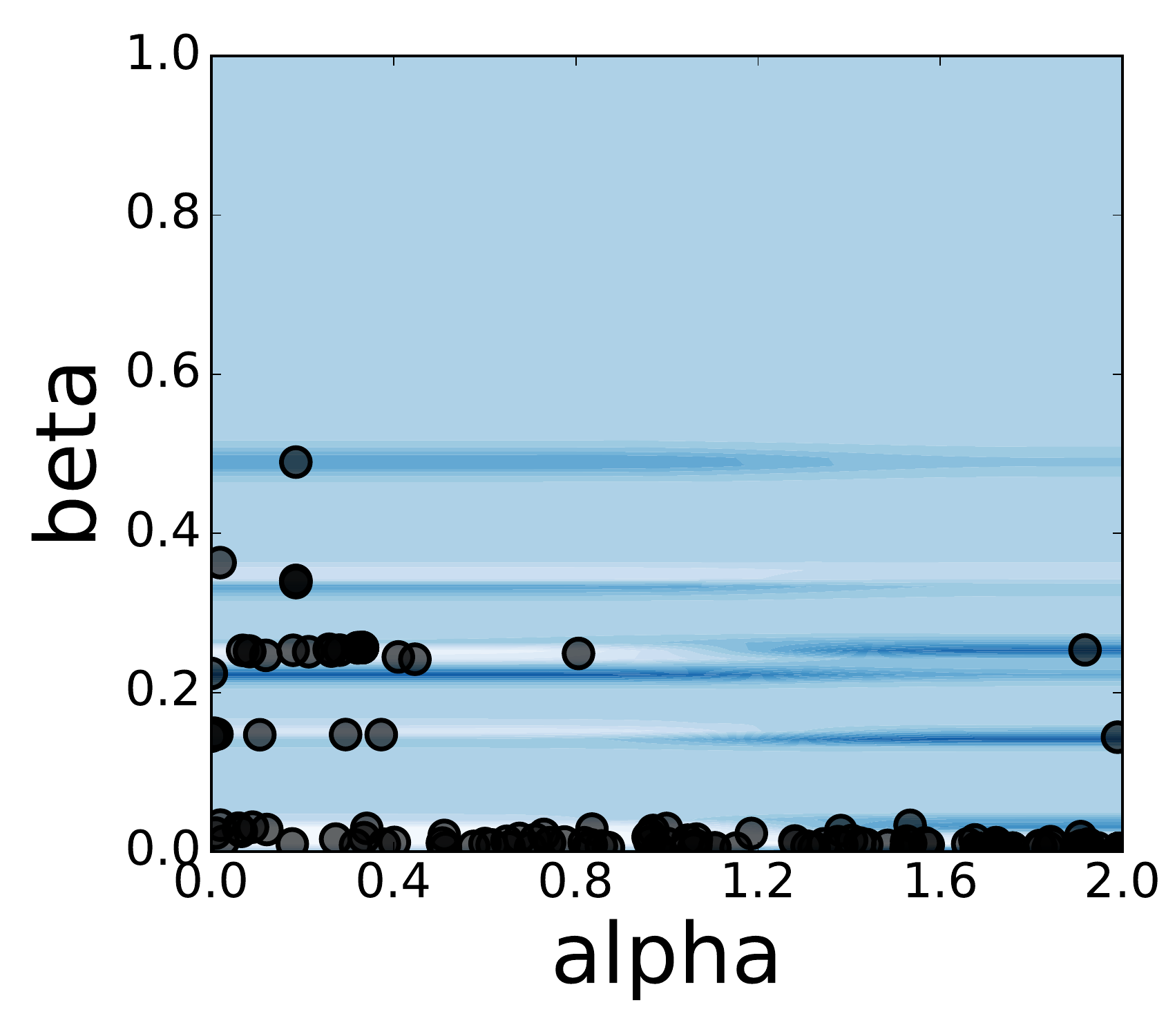}
%\caption{Caption of first subfigure}
\end{subfigure}
\caption{Sine method on nonlinear channel equalization task at SNR 16 dB. (left) Grid plot of simulated values. (right) Spearmint run with measurements (black points).} \label{fig:NLC_sine_16}
\end{figure}

\begin{figure}
\begin{subfigure}{0.55\textwidth}
\includegraphics[width=\linewidth]{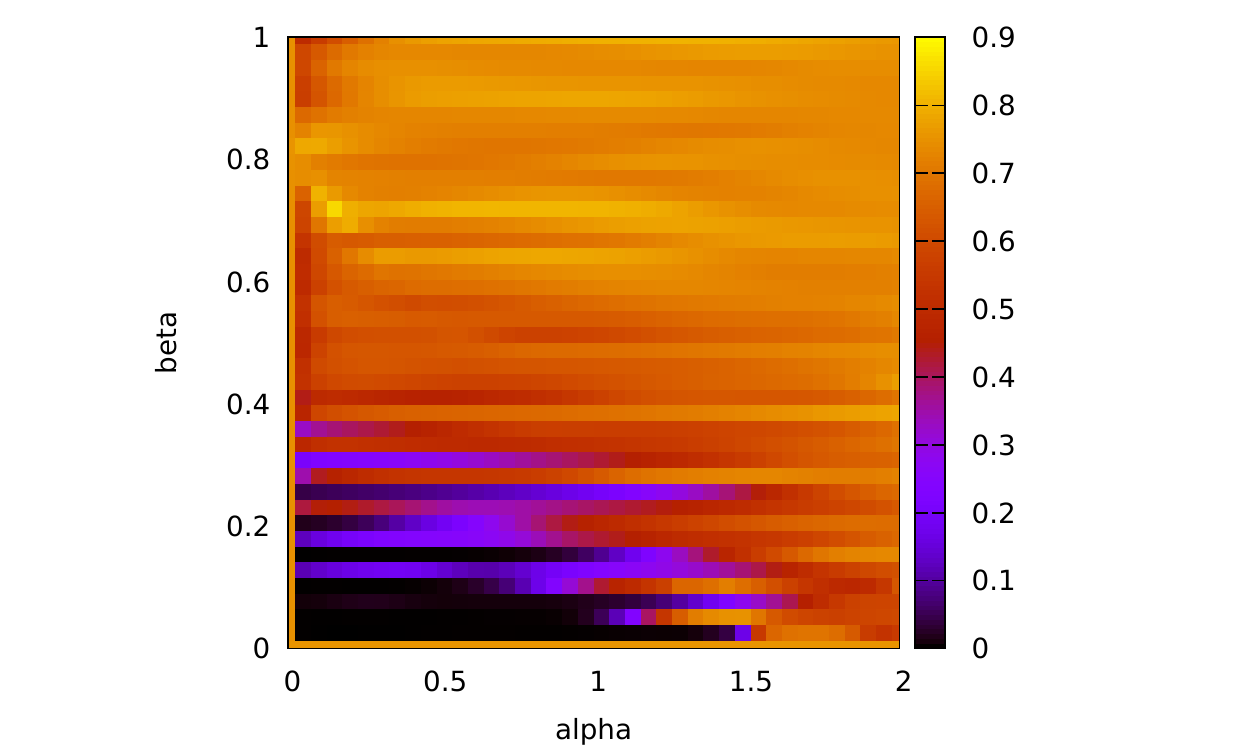}
%\caption{Caption of first subfigure}
\end{subfigure}
%\hspace*{\fill}
\begin{subfigure}{0.4\textwidth}
\includegraphics[width=\linewidth]{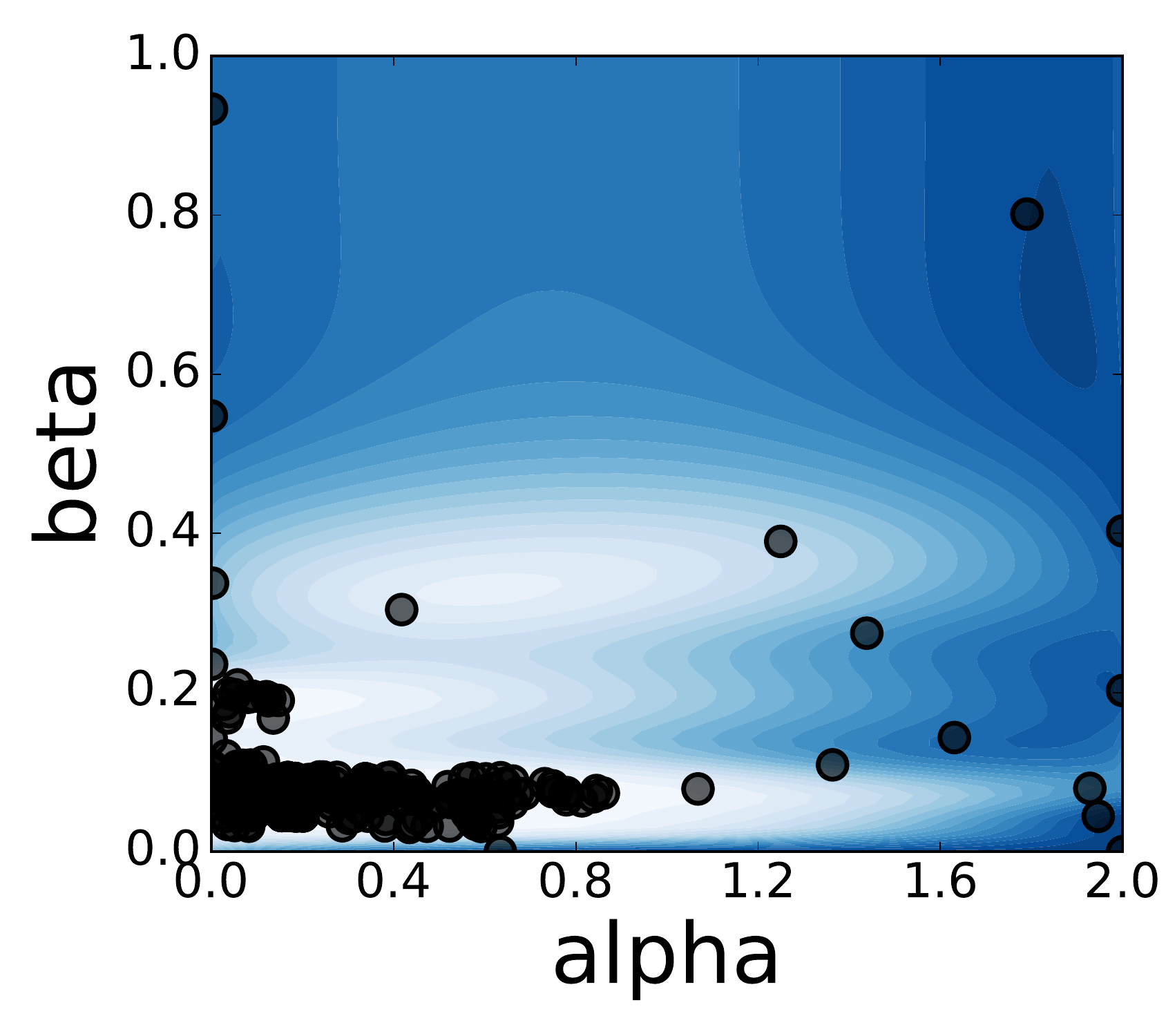}
%\caption{Caption of first subfigure}
\end{subfigure}
\caption{Sine method on nonlinear channel equalization task at SNR 32 dB. (left) Grid plot of simulated values. (right) Spearmint run with measurements (black points).} \label{fig:NLC_sine_32}
\end{figure}

\begin{figure}
\begin{subfigure}{0.55\textwidth}
\includegraphics[width=\linewidth]{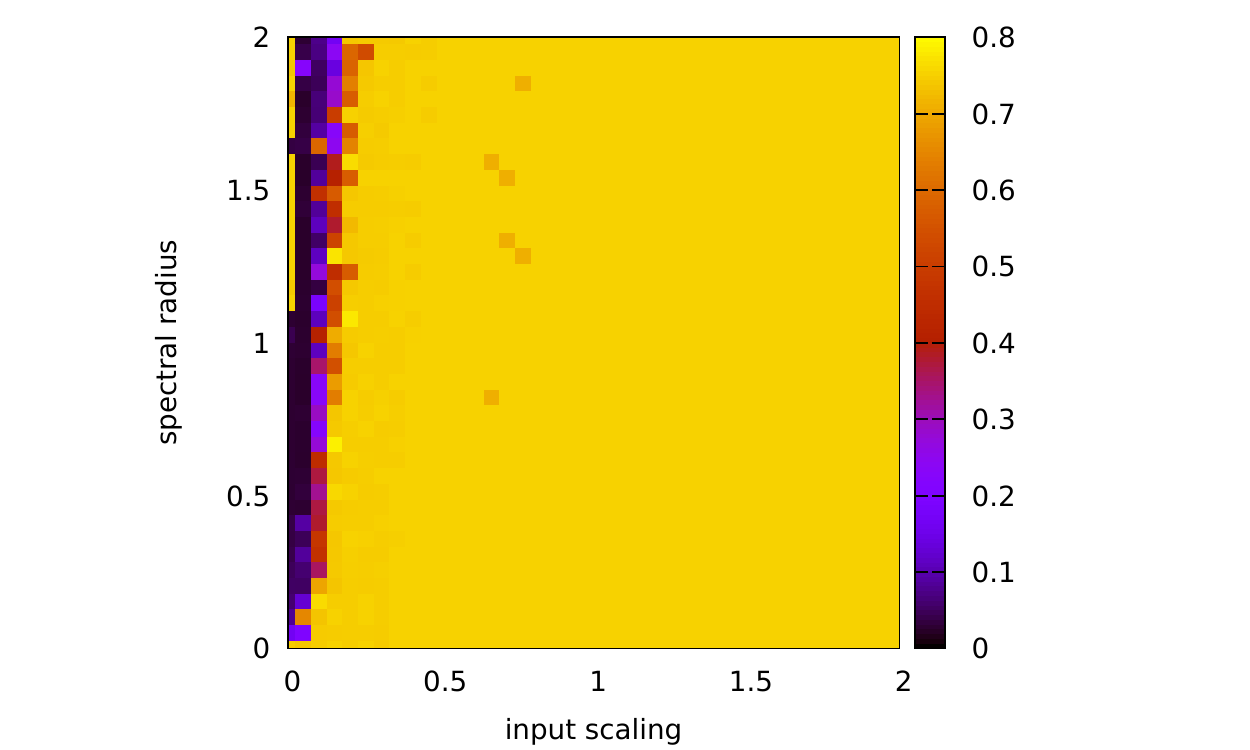}
%\caption{Caption of first subfigure}
\end{subfigure}
%\hspace*{\fill}
\begin{subfigure}{0.4\textwidth}
\includegraphics[width=\linewidth]{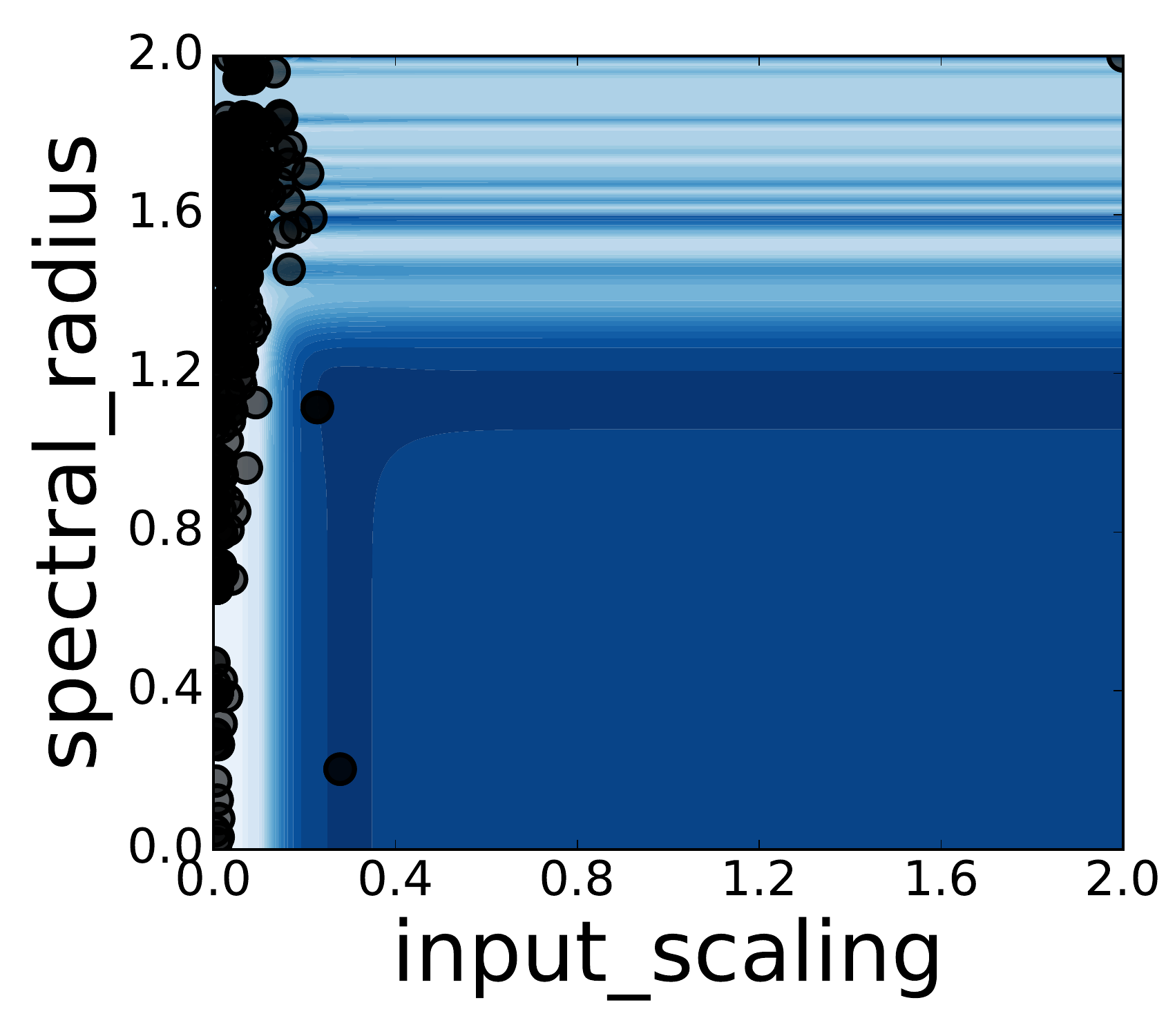}
%\caption{Caption of first subfigure}
\end{subfigure}
\caption{ESN on nonlinear channel equalization task at SNR 16 dB. (left) Grid plot of simulated values. (right) Spearmint run with measurements (black points).} \label{fig:NLC_esn_16}
\end{figure}

\begin{figure}
\begin{subfigure}{0.55\textwidth}
\includegraphics[width=\linewidth]{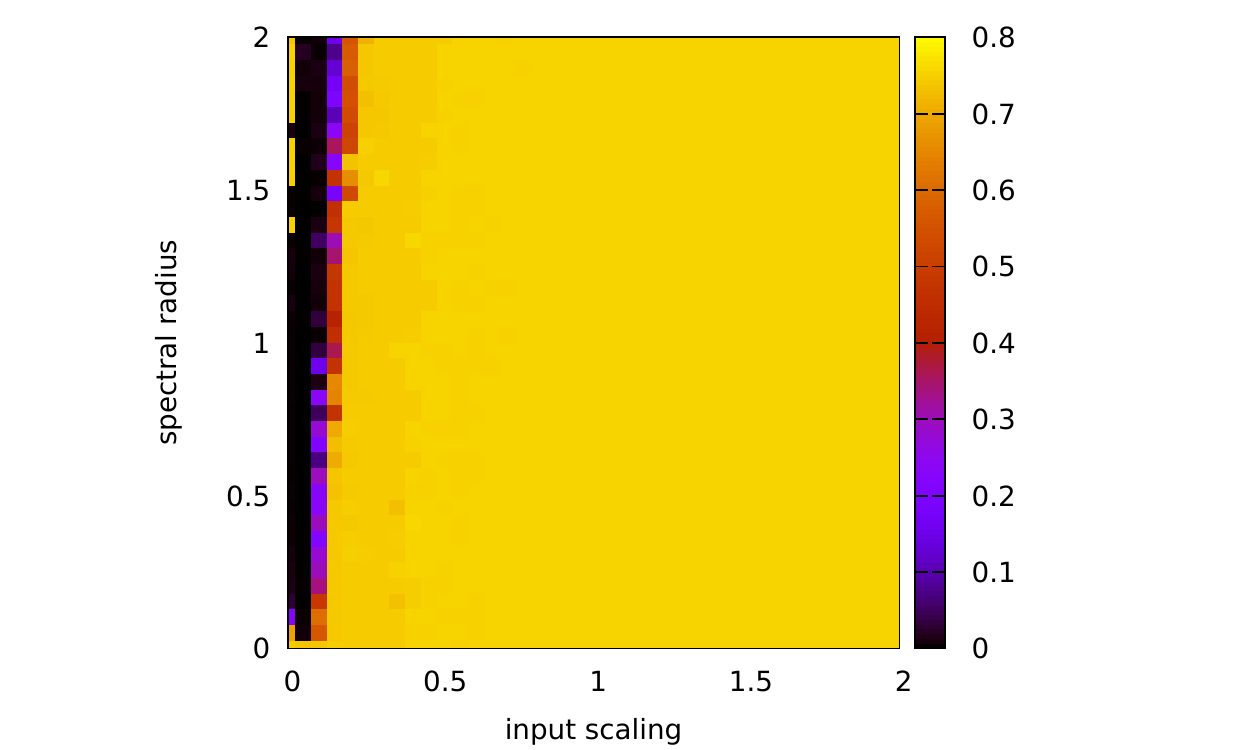}
%\caption{Caption of first subfigure}
\end{subfigure}
%\hspace*{\fill}
\begin{subfigure}{0.4\textwidth}
\includegraphics[width=\linewidth]{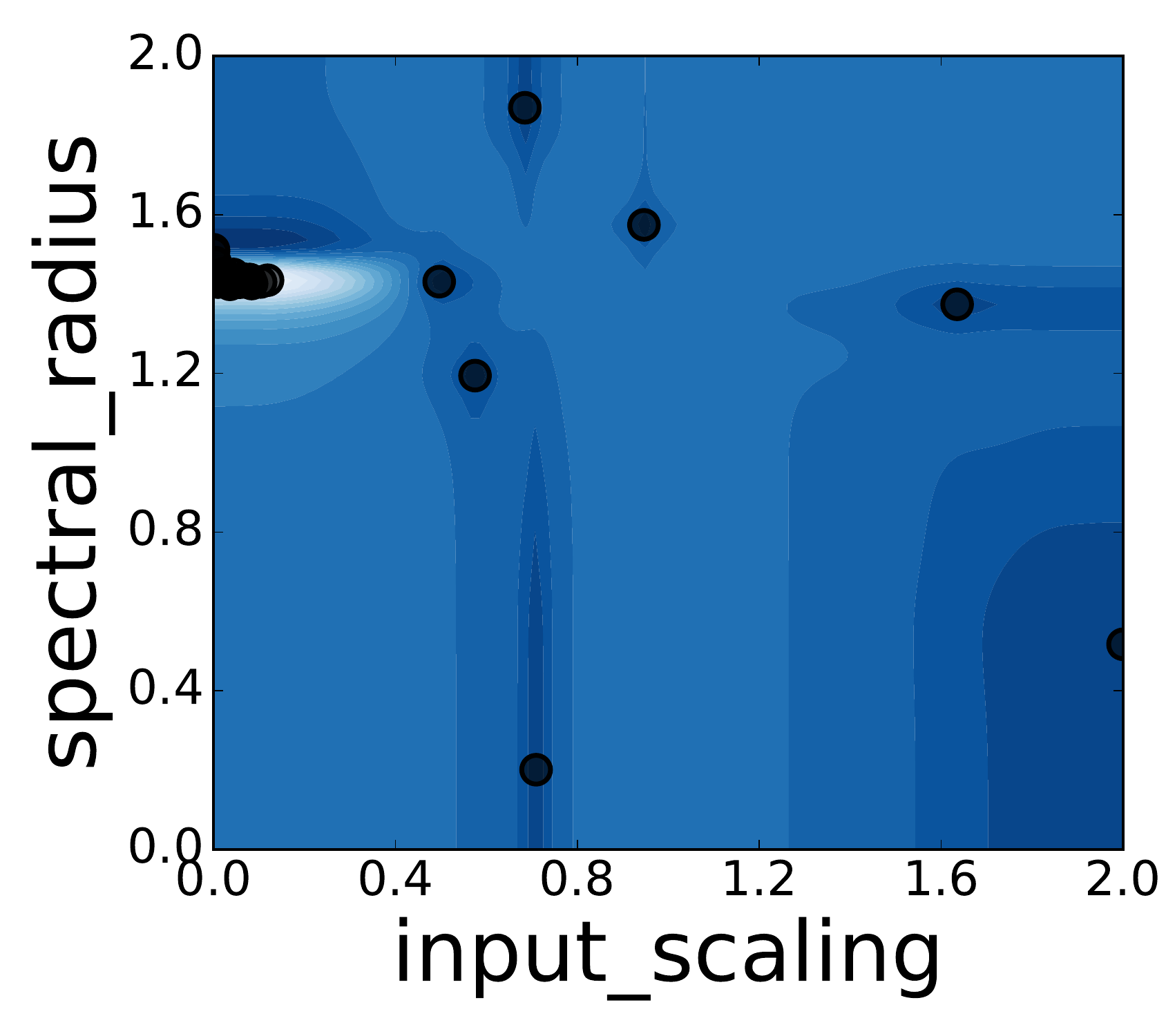}
%\caption{Caption of first subfigure}
\end{subfigure}
\caption{ESN on nonlinear channel equalization task at SNR 32 dB. (left) Grid plot of simulated values. (right) Spearmint run with measurements (black points).} \label{fig:NLC_esn_32}
\end{figure}

\end{appendix}

\section*{Acknowledgement}

Early work was done with Christian Van den Broeck, whom we thank for letting us work on this problem. 
The authors would like to thank 
Lars Keuninckx and Guy Van der Sande for discussions and reading the manuscript,
Bart Cleuren for reading the manuscript,
the reservoir computing group at IFISC for discussions,
and Peter Tino for answering our questions about echo state networks.
\\
TB is supported by the Fonds voor Wetenschappelijk Onderzoek (FWO), project R4859.
The computational resources and services used in this work were provided by the VSC (Flemish Supercomputer Center), funded by the Research Foundation - Flanders (FWO) and the Flemish Government - department EWI. In this context, the authors would like to thank Geert Jan Bex for his help with the deployment of our software on the VSC.


\begin{thebibliography}{10}
\providecommand{\url}[1]{#1}
\csname url@samestyle\endcsname
\providecommand{\newblock}{\relax}
\providecommand{\bibinfo}[2]{#2}
\providecommand{\BIBentrySTDinterwordspacing}{\spaceskip=0pt\relax}
\providecommand{\BIBentryALTinterwordstretchfactor}{4}
\providecommand{\BIBentryALTinterwordspacing}{\spaceskip=\fontdimen2\font plus
\BIBentryALTinterwordstretchfactor\fontdimen3\font minus
  \fontdimen4\font\relax}
\providecommand{\BIBforeignlanguage}[2]{{%
\expandafter\ifx\csname l@#1\endcsname\relax
\typeout{** WARNING: IEEEtran.bst: No hyphenation pattern has been}%
\typeout{** loaded for the language `#1'. Using the pattern for}%
\typeout{** the default language instead.}%
\else
\language=\csname l@#1\endcsname
\fi
#2}}
\providecommand{\BIBdecl}{\relax}
\BIBdecl

\bibitem{lecun2012efficient}
Y.~LeCun, L.~Bottou, G.~B. Orr, and K.-R. M{\"u}ller, ``Efficient backprop,''
  in \emph{Neural networks: Tricks of the trade}.\hskip 1em plus 0.5em minus
  0.4em\relax Springer, 2012, pp. 9--48.

\bibitem{lecun2015deep}
Y.~LeCun, Y.~Bengio, and G.~Hinton, ``Deep learning,'' \emph{Nature}, vol. 521,
  no. 7553, pp. 436--444, 2015.

\bibitem{hochreiter2001gradient}
S.~Hochreiter, Y.~Bengio, P.~Frasconi, and J.~Schmidhuber, ``Gradient flow in
  recurrent nets: the difficulty of learning long-term dependencies,'' in
  \emph{A field guide to dynamical recurrent neural networks}, S.~C. Kremer and
  J.~F. Kolen, Eds.\hskip 1em plus 0.5em minus 0.4em\relax IEEE Press, 2001.

\bibitem{arjovsky2016unitary}
M.~Arjovsky, A.~Shah, and Y.~Bengio, ``Unitary evolution recurrent neural
  networks,'' in \emph{International Conference on Machine Learning}, 2016, pp.
  1120--1128.

\bibitem{hochreiter1997long}
S.~Hochreiter and J.~Schmidhuber, ``Long short-term memory,'' \emph{Neural
  comput.}, vol.~9, no.~8, pp. 1735--1780, 1997.

\bibitem{schuster1997bidirectional}
M.~Schuster and K.~K. Paliwal, ``Bidirectional recurrent neural networks,''
  \emph{IEEE Trans. on Signal Proc.}, vol.~45, no.~11, pp. 2673--2681, 1997.

\bibitem{lukovsevivcius2012practical}
M.~Luko{\v{s}}evi{\v{c}}ius, ``A practical guide to applying echo state
  networks,'' in \emph{Neural networks: Tricks of the trade}.\hskip 1em plus
  0.5em minus 0.4em\relax Springer, 2012, pp. 659--686.

\bibitem{fischer2016photonic}
I.~Fischer, J.~Bueno, D.~Brunner, M.~C. Soriano, and C.~Mirasso, ``Photonic
  reservoir computing for ultra-fast information processing using semiconductor
  lasers,'' in \emph{ECOC 2016; 42nd European Conference on Optical
  Communication; Proceedings of}.\hskip 1em plus 0.5em minus 0.4em\relax VDE,
  2016, pp. 1--3.

\bibitem{duport2016fully}
F.~Duport, A.~Smerieri, A.~Akrout, M.~Haelterman, and S.~Massar, ``Fully
  analogue photonic reservoir computer,'' \emph{Scientific reports}, vol.~6,
  2016.

\bibitem{antonik2016towards}
P.~Antonik, M.~Hermans, F.~Duport, M.~Haelterman, and S.~Massar, ``Towards
  pattern generation and chaotic series prediction with photonic reservoir
  computers,'' in \emph{SPIE LASE}.\hskip 1em plus 0.5em minus 0.4em\relax
  International Society for Optics and Photonics, 2016, pp. 97\,320B--97\,320B.

\bibitem{katumba2017multiple}
A.~Katumba, M.~Freiberger, P.~Bienstman, and J.~Dambre, ``A multiple-input
  strategy to efficient integrated photonic reservoir computing,''
  \emph{Cognitive Computation}, pp. 1--8, 2017.

\bibitem{soriano2015delay}
M.~C. Soriano, S.~Ort{\'\i}n, L.~Keuninckx, L.~Appeltant, J.~Danckaert,
  L.~Pesquera, and G.~Van~der Sande, ``Delay-based reservoir computing: noise
  effects in a combined analog and digital implementation,'' \emph{IEEE Trans.
  Neural Netw. Learn. Syst.}, vol.~26, no.~2, pp. 388--393, 2015.

\bibitem{jaeger2007optimization}
H.~Jaeger, M.~Luko{\v{s}}evi{\v{c}}ius, D.~Popovici, and U.~Siewert,
  ``Optimization and applications of echo state networks with leaky-integrator
  neurons,'' \emph{Neural networks}, vol.~20, no.~3, pp. 335--352, 2007.

\bibitem{rabin2013sensitivity}
M.~J.~A. Rabin, M.~S. Hossain, M.~S. Ahsan, M.~A.~S. Mollah, and M.~T. Rahman,
  ``Sensitivity learning oriented nonmonotonic multi reservoir echo state
  network for short-term load forecasting,'' in \emph{Informatics, Electronics
  \& Vision (ICIEV), 2013 International Conference on}.\hskip 1em plus 0.5em
  minus 0.4em\relax IEEE, 2013, pp. 1--6.

\bibitem{sergio2012pso}
A.~Sergio and T.~Ludermir, ``Pso for reservoir computing optimization,''
  \emph{Artificial Neural Networks and Machine Learning--ICANN 2012}, pp.
  685--692, 2012.

\bibitem{basterrech2014experimental}
S.~Basterrech, E.~Alba, and V.~Sn{\'a}{\v{s}}el, ``An experimental analysis of
  the echo state network initialization using the particle swarm
  optimization,'' in \emph{Nature and Biologically Inspired Computing (NaBIC),
  2014 Sixth World Congress on}.\hskip 1em plus 0.5em minus 0.4em\relax IEEE,
  2014, pp. 214--219.

\bibitem{ferreira2009genetic}
A.~A. Ferreira and T.~B. Ludermir, ``Genetic algorithm for reservoir computing
  optimization,'' in \emph{Neural Networks, 2009. IJCNN 2009. International
  Joint Conference on}.\hskip 1em plus 0.5em minus 0.4em\relax IEEE, 2009, pp.
  811--815.

\bibitem{ferreira2013approach}
A.~A. Ferreira, T.~B. Ludermir, and R.~R. De~Aquino, ``An approach to reservoir
  computing design and training,'' \emph{Expert systems with applications},
  vol.~40, no.~10, pp. 4172--4182, 2013.

\bibitem{rigamonti2016echo}
M.~Rigamonti, P.~Baraldi, E.~Zio, I.~Roychoudhury, K.~Goebel, and S.~Poll,
  ``Echo state network for the remaining useful life preeiction of a turbofan
  engine,'' in \emph{Proceedings of the Third European Prognostic and Health
  Management Conference, PHME}, 2016.

\bibitem{NIPS2012_4522}
J.~Snoek, H.~Larochelle, and R.~P. Adams, ``Practical bayesian optimization of
  machine learning algorithms,'' in \emph{Advances in Neural Information
  Processing Systems 25}, F.~Pereira, C.~J.~C. Burges, L.~Bottou, and K.~Q.
  Weinberger, Eds.\hskip 1em plus 0.5em minus 0.4em\relax Curran Associates,
  Inc., 2012, pp. 2951--2959.

\bibitem{Goodfellow-et-al-2016-Book}
I.~Goodfellow, Y.~Bengio, and A.~Courville, \emph{Deep Learning}.\hskip 1em
  plus 0.5em minus 0.4em\relax MIT Press, 2016,
  \url{http://www.deeplearningbook.org}.

\bibitem{jaeger2001echo}
H.~Jaeger, ``The “echo state” approach to analysing and training recurrent
  neural networks-with an erratum note,'' \emph{Technical Report GMD Report
  148, German National Research Center for Information Technology}, 2001.

\bibitem{maass2002real}
W.~Maass, T.~Natschl{\"a}ger, and H.~Markram, ``Real-time computing without
  stable states: A new framework for neural computation based on
  perturbations,'' \emph{Neural comput.}, vol.~14, no.~11, pp. 2531--2560,
  2002.

\bibitem{scardapane2017randomness}
S.~Scardapane and D.~Wang, ``Randomness in neural networks: an overview,''
  \emph{Wiley Interdisciplinary Reviews: Data Mining and Knowledge Discovery},
  vol.~7, no.~2, 2017.

\bibitem{manjunath2013echo}
G.~Manjunath and H.~Jaeger, ``Echo state property linked to an input: Exploring
  a fundamental characteristic of recurrent neural networks,'' \emph{Neural
  computation}, vol.~25, no.~3, pp. 671--696, 2013.

\bibitem{wainrib2016local}
G.~Wainrib and M.~N. Galtier, ``A local echo state property through the largest
  lyapunov exponent,'' \emph{Neural Networks}, vol.~76, pp. 39--45, 2016.

\bibitem{appeltant2011information}
L.~Appeltant, M.~C. Soriano, G.~Van~der Sande, J.~Danckaert, S.~Massar,
  J.~Dambre, B.~Schrauwen, C.~R. Mirasso, and I.~Fischer, ``Information
  processing using a single dynamical node as complex system,'' \emph{Nat.
  Commun.}, vol.~2, p. 468, 2011.

\bibitem{larger2012photonic}
L.~Larger, M.~C. Soriano, D.~Brunner, L.~Appeltant, J.~M. Guti{\'e}rrez,
  L.~Pesquera, C.~R. Mirasso, and I.~Fischer, ``Photonic information processing
  beyond turing: an optoelectronic implementation of reservoir computing,''
  \emph{Optics express}, vol.~20, no.~3, pp. 3241--3249, 2012.

\bibitem{paquot2012optoelectronic}
Y.~Paquot, F.~Duport, A.~Smerieri, J.~Dambre, B.~Schrauwen, M.~Haelterman, and
  S.~Massar, ``Optoelectronic reservoir computing,'' \emph{Sci. Rep.}, vol.~2,
  2012.

\bibitem{brunner2013parallel}
D.~Brunner, M.~C. Soriano, C.~R. Mirasso, and I.~Fischer, ``Parallel photonic
  information processing at gigabyte per second data rates using transient
  states,'' \emph{Nat. Commun.}, vol.~4, p. 1364, 2013.

\bibitem{siliconRC}
K.~Vandoorne, P.~Mechet, T.~Van~Vaerenbergh, M.~Fiers, G.~Morthier,
  D.~Verstraeten, B.~Schrauwen, J.~Dambre, and P.~Bienstman, ``Experimental
  demonstration of reservoir computing on a silicon photonics chip,''
  \emph{Nat. Commun.}, vol.~5, 2014.

\bibitem{williams2006gaussian}
C.~K. Williams and C.~E. Rasmussen, ``Gaussian processes for machine
  learning,'' \emph{the MIT Press}, vol.~2, no.~3, p.~4, 2006.

\bibitem{brochu2010tutorial}
E.~Brochu, V.~M. Cora, and N.~De~Freitas, ``A tutorial on bayesian optimization
  of expensive cost functions, with application to active user modeling and
  hierarchical reinforcement learning,'' \emph{arXiv preprint arXiv:1012.2599},
  2010.

\bibitem{sivia2006data}
D.~Sivia and J.~Skilling, \emph{Data analysis: a Bayesian tutorial}.\hskip 1em
  plus 0.5em minus 0.4em\relax Oxford University Press, Oxford, 2006.

\bibitem{snoek2014input}
J.~Snoek, K.~Swersky, R.~S. Zemel, and R.~P. Adams, ``Input warping for
  bayesian optimization of non-stationary functions.'' in \emph{Proc. Int.
  Conf. Mach. Learn.}, 2014, pp. 1674--1682.

\bibitem{mockus1994application}
J.~Mockus, ``Application of bayesian approach to numerical methods of global
  and stochastic optimization,'' \emph{Journal of Global Optimization}, vol.~4,
  no.~4, pp. 347--365, 1994.

\bibitem{bull2011convergence}
A.~D. Bull, ``Convergence rates of efficient global optimization algorithms,''
  \emph{Journal of Machine Learning Research}, vol.~12, no. Oct, pp.
  2879--2904, 2011.

\bibitem{kawaguchi2015bayesian}
K.~Kawaguchi, L.~P. Kaelbling, and T.~Lozano-P{\'e}rez, ``Bayesian optimization
  with exponential convergence,'' in \emph{Advances in Neural Information
  Processing Systems}, 2015, pp. 2809--2817.

\bibitem{de2012exponential}
N.~De~Freitas, A.~Smola, and M.~Zoghi, ``Exponential regret bounds for gaussian
  process bandits with deterministic observations,'' in \emph{Proceedings of
  the 29th International Conference on Machine Learning}, 2012.

\bibitem{makridakis1994time}
N.~A. Gershenfeld and A.~S. Weigend, ``The future of time series: learning and
  understanding,'' in \emph{Time series prediction: Forecasting the future and
  understanding the past}, A.~S. Weigend and N.~A. Gershenfeld, Eds.\hskip 1em
  plus 0.5em minus 0.4em\relax Addison-Wesley Publishing Company, Reading, MA,
  USA, 1994.

\bibitem{sfonline}
``Santa fe laser data,''
  \url{http://www-psych.stanford.edu/~andreas/Time-Series/SantaFe.html},
  accessed: 2016-08-31.

\bibitem{atiya2000new}
A.~F. Atiya and A.~G. Parlos, ``New results on recurrent network training:
  unifying the algorithms and accelerating convergence,'' \emph{IEEE Trans.
  Neural Netw.}, vol.~11, no.~3, pp. 697--709, 2000.

\bibitem{mathews1994adaptive}
V.~J. Mathews and J.~Lee, ``Adaptive algorithms for bilinear filtering,'' in
  \emph{SPIE's 1994 International Symposium on Optics, Imaging, and
  Instrumentation}.\hskip 1em plus 0.5em minus 0.4em\relax International
  Society for Optics and Photonics, 1994, pp. 317--327.

\bibitem{jaeger2004harnessing}
H.~Jaeger and H.~Haas, ``Harnessing nonlinearity: Predicting chaotic systems
  and saving energy in wireless communication,'' \emph{Science}, vol. 304, no.
  5667, pp. 78--80, 2004.

\bibitem{rodan2010simple}
A.~Rodan and P.~Tino, ``Simple deterministically constructed recurrent neural
  networks,'' in \emph{International Conference on Intelligent Data Engineering
  and Automated Learning}.\hskip 1em plus 0.5em minus 0.4em\relax Springer,
  2010, pp. 267--274.

\bibitem{rodan2011minimum}
------, ``Minimum complexity echo state network,'' \emph{IEEE Trans. Neural
  Netw.}, vol.~22, no.~1, pp. 131--144, 2011.

\bibitem{scikit}
F.~Pedregosa \emph{et~al.}, ``Scikit-learn: Machine learning in python,''
  \emph{J. Mach. Learn. Res.}, vol.~12, pp. 2825--2830, 2011.

\bibitem{verstraeten2012oger}
D.~Verstraeten, B.~Schrauwen, S.~Dieleman, P.~Brakel, P.~Buteneers, and
  D.~Pecevski, ``Oger: modular learning architectures for large-scale
  sequential processing,'' \emph{J. Mach. Learn. Res.}, vol.~13, no. Oct, pp.
  2995--2998, 2012.

\bibitem{zito2009modular}
T.~Zito, N.~Wilbert, L.~Wiskott, and P.~Berkes, ``Modular toolkit for data
  processing (mdp): A python data processing framework,'' \emph{Front.
  Neuroinf.}, 2009.

\bibitem{appeltant2012reservoir}
L.~Appeltant, ``Reservoir computing based on delay-dynamical systems,''
  \emph{These de Doctorat, Vrije Universiteit Brussel/Universitat de les Illes
  Balears}, 2012.

\bibitem{appeltant2014constructing}
L.~Appeltant, G.~Van~der Sande, J.~Danckaert, and I.~Fischer, ``Constructing
  optimized binary masks for reservoir computing with delay systems,''
  \emph{Sci. Rep.}, vol.~4, p. 3629, 2014.

\bibitem{soriano2013optoelectronic}
M.~C. Soriano, S.~Ort{\'\i}n, D.~Brunner, L.~Larger, C.~R. Mirasso, I.~Fischer,
  and L.~Pesquera, ``Optoelectronic reservoir computing: tackling noise-induced
  performance degradation,'' \emph{Optics express}, vol.~21, no.~1, pp. 12--20,
  2013.

\bibitem{ortin2015unified}
S.~Ort{\'\i}n, M.~C. Soriano, L.~Pesquera, D.~Brunner, D.~San-Mart{\'\i}n,
  I.~Fischer, C.~R. Mirasso, and J.~M. Guti{\'e}rrez, ``A unified framework for
  reservoir computing and extreme learning machines based on a single
  time-delayed neuron,'' \emph{Sci. Rep.}, vol.~5, 2015.

\bibitem{hicke2013information}
K.~Hicke, M.~A. Escalona-Mor{\'a}n, D.~Brunner, M.~C. Soriano, I.~Fischer, and
  C.~R. Mirasso, ``Information processing using transient dynamics of
  semiconductor lasers subject to delayed feedback,'' \emph{IEEE J. Sel. Topics
  Quantum Electron.}, vol.~19, no.~4, pp. 1\,501\,610--1\,501\,610, 2013.

\bibitem{nguimdo2014fast}
R.~M. Nguimdo, G.~Verschaffelt, J.~Danckaert, and G.~Van~der Sande, ``Fast
  photonic information processing using semiconductor lasers with delayed
  optical feedback: Role of phase dynamics,'' \emph{Optics express}, vol.~22,
  no.~7, pp. 8672--8686, 2014.

\bibitem{nguimdo2015simultaneous}
------, ``Simultaneous computation of two independent tasks using reservoir
  computing based on a single photonic nonlinear node with optical feedback,''
  \emph{IEEE Trans. Neural Netw. Learn. Syst.}, vol.~26, no.~12, pp.
  3301--3307, 2015.

\bibitem{nguimdo2016reducing}
------, ``Reducing the phase sensitivity of laser-based optical reservoir
  computing systems,'' \emph{Optics express}, vol.~24, no.~2, pp. 1238--1252,
  2016.

\bibitem{wang2015optimizing}
H.~Wang and X.~Yan, ``Optimizing the echo state network with a binary particle
  swarm optimization algorithm,'' \emph{Knowledge-Based Systems}, vol.~86, pp.
  182--193, 2015.

\bibitem{dong2016scaling}
J.~Dong, S.~Gigan, F.~Krzakala, and G.~Wainrib, ``Scaling up echo-state
  networks with multiple light scattering,'' \emph{arXiv preprint
  arXiv:1609.05204}, 2016.

\end{thebibliography}
\end{document}